\documentclass{article} 
\usepackage[final]{colm2026_conference}

\usepackage[T1]{fontenc}   
\usepackage[utf8]{inputenc}  

\usepackage{pgfplots}
\pgfplotsset{width=10cm,compat=1.9}
\usepackage[]{algpseudocode}
\usepackage[]{algorithm}
\algtext*{EndFor}%
\algtext*{EndProcedure}%
\DeclareUnicodeCharacter{2212}{−}
\usepgfplotslibrary{groupplots,dateplot}
\usetikzlibrary{patterns,shapes.arrows}
\pgfplotsset{compat=newest}

\usepackage{tikzscale} 
\usepackage{xcolor}
\usepackage{bbm}
\newcommand{\probP}{\text{I\kern-0.15em P}}
\usepackage{color}
\usepackage{amssymb}
\usepackage{pifont}
 \usepackage{array}
\usepackage{tabularx,booktabs}
\usepackage{enumitem}
\usepackage{placeins}
\usepackage[normalem]{ulem}
\usepackage{ragged2e}
\usepackage{dsfont}
\usepackage{scalerel,xparse}
\usepackage{adjustbox}
\usepackage{subfigure}
\usepackage{bbm}

\newcolumntype{Y}{>{\raggedright\arraybackslash}X}

\definecolor{lightgreen}{RGB}{215, 242, 186}
\definecolor{darkgreen}{RGB}{156, 198, 155}
\definecolor{lightred}{RGB}{255, 214, 214}
\definecolor{darkred}{RGB}{255, 186, 186}

\usepackage{soul}
\sethlcolor{orange!50}

\newcommand{\hlpref}[1]{\sethlcolor{lightgreen}\hl{#1}\sethlcolor{orange!50}}
\newcommand{\hldispref}[1]{\sethlcolor{lightred}\hl{#1}\sethlcolor{orange!50}}

\usepackage{multirow}
\usepackage{fontawesome5}
\usepackage{xspace}
\usepackage{todonotes}
\usepackage[most]{tcolorbox}
\usepackage[table]{xcolor}
\usepackage{array}
\usepackage{colortbl}
\usepackage{makecell}
\usepackage{graphicx}

\usepackage{subcaption}
\usepackage{caption}
\usepackage{pifont}
\usepackage{amssymb}
\usepackage{tabularx, booktabs}

\newcommand{\cmark}{\ding{51}} 
\newcommand{\xmark}{\ding{55}} 

\definecolor{zoey green}{rgb}{0.684,0.836,0.227}

\newcommand{\ignore}[1]{}

\usepackage[table]{xcolor}
\usepackage{soul}
\definecolor{direct}{RGB}{218, 98, 125}
\definecolor{agentic}{RGB}{0, 166, 166}
\sethlcolor{direct!40}

\usepackage{newfloat}
\usepackage{listings}
\floatstyle{ruled}
\newfloat{listing}{tb}{lst}{}
\floatname{listing}{Listing}

\definecolor{bggray}{rgb}{0.95, 0.95, 0.95}
\usepackage[%
    framemethod=tikz,
    skipbelow=\topskip,
    skipabove=\topskip
]{mdframed}
\mdfsetup{%
    leftmargin=0pt,
    rightmargin=0pt,
    backgroundcolor=bggray,
    middlelinecolor=black,
    roundcorner=3
}
\newtcolorbox[list inside=prompt,auto counter,number within=section]{prompt}[1][]{
    colbacktitle=black!60,
    fonttitle=\small,
    coltitle=white,
    fontupper=\footnotesize,
    boxsep=4pt,
    left=0pt,
    top=0pt,
    bottom=0pt,
    boxrule=1pt,
    #1,
}

\usepackage{microtype}
\usepackage{hyperref}
\usepackage{url}
\usepackage{booktabs}
\usepackage{lineno}

\definecolor{darkblue}{rgb}{0, 0, 0.5}
\hypersetup{colorlinks=true, citecolor=darkblue, linkcolor=darkblue, urlcolor=darkblue}

\definecolor{multilingual}{RGB}{165,55,96}
\definecolor{intermediate}{RGB}{68,112,77}
\definecolor{cognitive}{RGB}{42,111,151}

\definecolor{feature}{RGB}{0,169,165}
\definecolor{sae}{RGB}{103,78,167}
\definecolor{testtime}{RGB}{245,133,73}

\usepackage{newfloat}
\usepackage{listings}
\floatstyle{ruled}
\newfloat{listing}{tb}{lst}{}
\floatname{listing}{Listing}

\usepackage{cleveref}

\usepackage{mdframed,xcolor}
\mdfdefinestyle{grayframe}{
  linecolor=lightgrayframe,
  linewidth=0.8pt,
  roundcorner=2pt,
  backgroundcolor=white
}

\title{What Makes \textit{Good} Multilingual Reasoning? \\ Disentangling Traces with Measurable Features}

\author{\textbf{Dayeon Ki} \textsuperscript{\raisebox{-0.2em}{\includegraphics[height=1em]{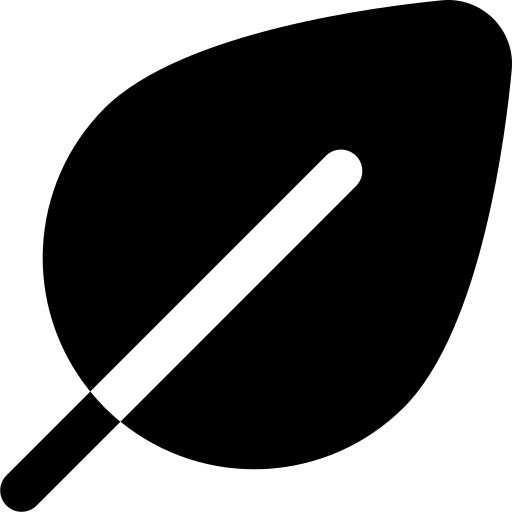}}}, 
\textbf{Kevin Duh} \textsuperscript{\raisebox{-0.2em}{\includegraphics[height=1em]{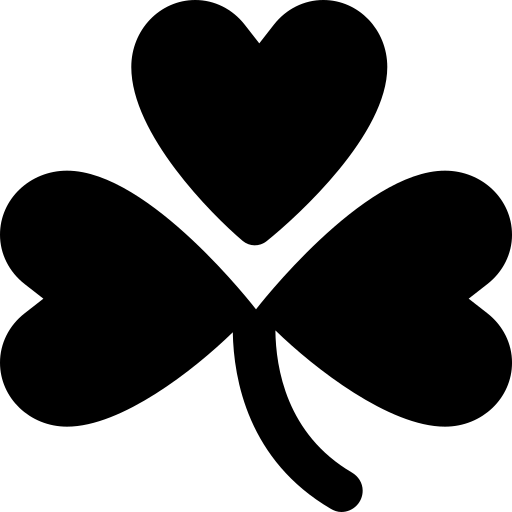}}}, 
\textbf{Marine Carpuat} \textsuperscript{\raisebox{-0.2em}{\includegraphics[height=1em]{figures/logo/leaf.png}}} \\
University of Maryland \textsuperscript{\raisebox{-0.2em}{\includegraphics[height=1em]{figures/logo/leaf.png}}}, Johns Hopkins University \textsuperscript{\raisebox{-0.2em}{\includegraphics[height=1em]{figures/logo/three-leaf-clover.png}}} \\
\texttt{dayeonki@umd.edu}
}

\begin{document}

\ifcolmsubmission
\linenumbers
\fi

\maketitle

\begin{abstract}

Large Reasoning Models (LRMs) still exhibit large performance gaps between English and other languages, yet much current work assumes these gaps can be closed simply by making reasoning in every language resemble English reasoning. 
This work challenges this assumption by asking instead: what actually characterizes successful reasoning traces in multilingual settings, and to what extent do English‑derived reasoning features genuinely help in other languages?
We first define a suite of measurable reasoning features spanning multilingual alignment, reasoning step, and reasoning flow aspects of reasoning traces, and use logistic regression to quantify how each feature associates with final answer accuracy. 
We further train sparse autoencoders over multilingual traces to automatically discover latent reasoning concepts that instantiate or extend these features.
Finally, we use the features to re-rank traces and measure their impact on accuracy at test time.
Across two mathematical reasoning benchmarks, four LRMs, and ten languages, we find that most features are positively associated with accuracy, but the strength of association varies considerably across languages and can even reverse in some.
Our findings challenge English-centric reward designs and point toward adaptive objectives that accommodate language-specific reasoning patterns, with concrete implications for multilingual benchmark and reward design.\footnote{Code and data available at \url{https://github.com/dayeonki/multilingual_reasoning}.}
\end{abstract}




\section{Introduction}

Advancing multilingual reasoning is critical for deploying Large Reasoning Models (LRMs) across diverse languages and improving user experiences worldwide \citep{shi2022languagemodelsmultilingualchainofthought, ghosh-etal-2025-survey}. 
Yet substantial performance and behavioral gaps remain when LRMs are queried in languages other than English \citep{huang-etal-2025-benchmax, tam2025languagemattersmultilingualinput}, leading to lower reasoning accuracy \citep{wang2025polymath, luo-etal-2025-mmath} and systematic mismatches between reasoning traces and final answers \citep{ovalle2026begdifferunderstandingreasoninganswer, reddy2026effectscriptsformatsllm}.

Recent approaches to closing these gaps often project reasoning in other languages into English space: they translate queries into English \citep{zhu-etal-2024-question, ko-etal-2025-understand, liu2026selfimprovingmultilinguallongreasoning}, or reward traces that are structurally \citep{lai-nissim-2024-mcot} or semantically similar to English reasoning \citep{she-etal-2024-mapo, faisal2025aligningmultilingualreasoningverifiable, zhang2026thinknativelyunlockingmultilingual, sutawika2026gainedtranslationprivilegedpairwise}.
While these methods can reduce accuracy gaps, they implicitly assume that the features signaling ``good'' reasoning in English transfer unchanged to other languages.
This assumption is fragile: emerging evidence suggests models can sometimes reason more effectively in the original language instead of translating to English \citep{gao2025thinkingmultilinguallyempowerllm}, and that traces when queried in other languages may follow distinct but equally valid reasoning trajectories, particularly for typologically distant languages \citep{tam2025languagemattersmultilingualinput}.
Blindly rewarding English-like reasoning therefore risks obscuring or even penalizing language-specific reasoning behaviors with correct answers.

This raises a central question: what actually characterizes successful reasoning traces in multilingual settings, and to what extent do English-derived reasoning features transfer across languages?
Answering this requires moving beyond final answer accuracy alone toward a more systematic analysis of reasoning traces themselves. 
In this work, we take a first step by analyzing multilingual traces through a suite of measurable, human-interpretable features, studying not only how these features relate to accuracy but also whether they can be used at test time to help re-rank traces for higher accuracy.

\begin{figure*}
    \centering
    \includegraphics[width=\linewidth]{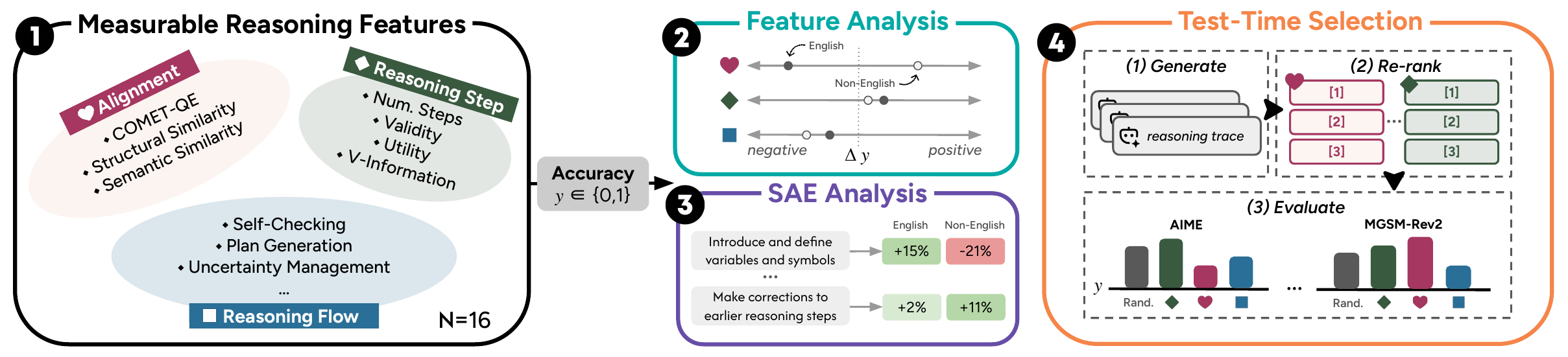}
    \caption{\textbf{Overview of our method.} We define 16 measurable reasoning features spanning multilingual alignment, reasoning step, and reasoning flow. We estimate each feature's effect on accuracy $y$ via regression (\textcolor{feature}{\textbf{Feature Analysis}}), validate and discover additional features using sparse autoencoders (\textcolor{sae}{\textbf{SAE Analysis}}), and use these features to select reasoning traces at inference (\textcolor{testtime}{\textbf{Test-Time Selection}}).}
    \label{fig:main_figure}
\end{figure*}

As illustrated in~\autoref{fig:main_figure}, we first define a set of measurable reasoning features spanning multilingual alignment, reasoning step, and reasoning flow aspects of reasoning traces (\S\ref{sec:define}).
We then quantify how each feature relates to per-language accuracy using univariate logistic regression (\S\ref{sec:feature}).
To move beyond this hand-designed set, we train sparse autoencoders (SAEs) over multilingual reasoning traces to automatically discover latent reasoning concepts, testing whether they recover and extend the same patterns (\S\ref{sec:sae}).
Finally, we use these features as test-time selection policies: for each language, we generate multiple candidate reasoning traces and re-rank them by each feature value, using the resulting accuracy as a probe of whether that feature can guide selection toward better multilingual reasoning performance (\S\ref{sec:testtime}).

Across two multilingual mathematical reasoning benchmarks, four LRMs, and 10 languages (\S\ref{sec:setup}), we find that most features are positively associated with accuracy, but the strength of this association varies considerably and can even reverse in some languages (\S\ref{sec:rq1}).
SAE-derived latent concepts qualitatively confirm these patterns and reveal additional behaviors not captured by our hand-designed features (\S\ref{sec:rq2}). 
In test-time selection, the conventional choice of semantic similarity to English traces is competitive but not universally best; for some models, alternative features such as utility yield higher accuracy (\S\ref{sec:rq3}).

Together, our findings challenge assumptions that uncritically favor English-like traces and instead point toward adaptive objectives that accommodate language-specific reasoning patterns, with concrete implications for multilingual benchmark and reward design (\S\ref{sec:implications}).

\section{Related Work}

\subsection{Multilingual Reasoning}
A growing body of work documents substantial performance and behavioral gaps across languages in LRMs \citep{tam2025languagemattersmultilingualinput, qi-etal-2025-models, wang2025polymath, luo-etal-2025-mmath}. 
To mitigate these gaps, training-time approaches typically operationalize successful multilingual reasoning traces through English-centric signals: they translate queries into English before reasoning \citep{zhu-etal-2024-question, ko-etal-2025-understand, liu2026selfimprovingmultilinguallongreasoning, huang2026tapotranslationaugmentedpolicy}, or design rewards that favor cross-lingual semantic similarity to English reference traces, via LLM-as-a-judge scores \citep{zhang2026thinknativelyunlockingmultilingual, sutawika2026gainedtranslationprivilegedpairwise} or cosine embedding similarity \citep{faisal2025aligningmultilingualreasoningverifiable, liu2026selfimprovingmultilinguallongreasoning}. 
We question this design choice by asking how other properties of multilingual reasoning traces\textemdash beyond semantic similarity to English\textemdash actually relate to reasoning performance, and aim to provide a more nuanced account of what constitutes a \textit{good} multilingual reasoning.



\subsection{Disentangling Reasoning Traces}
\label{sec:related_work1}

Recent work decomposes LRM traces into intermediate steps to characterize their structural and behavioral properties \citep{lanham2023measuring, paul2024making}. 
A diverse set of step-level evaluation criteria has been proposed, including factuality \citep{golovneva2022roscoe}, informativeness \citep{prasad-etal-2023-receval}, relevance \citep{jacovi2024chain}, utility and validity \citep{lee-hockenmaier-2025-evaluating}, and coherence \citep{do2025definesgoodreasoningllms}.
Structural analyses further link graph structure \citep{jiang-etal-2025-makes, gandhi2025cognitive, li2025llms} and self-revision patterns \citep{feng2025characterizeseffectivereasoningrevisiting} to performance through correlational methods.
Separately, sparse autoencoders (SAEs; \citet{cunningham2023sparse}) have been used to extract interpretable latent features that both explain reasoning behavior \citep{galichin2025icoveredbaseshere} and steer models toward higher performance \citep{ma2026falsifyingsparseautoencoderreasoning}.
However, this literature is almost entirely confined to English, operating on English datasets and feature spaces, leaving generalization to multilingual settings largely unexplored. 
Our work bridges this gap by examining how these features transfer across languages and relate to multilingual reasoning performance.


\section{Method}

Our goal is to characterize what constitutes a successful reasoning trace in multilingual settings. 
To this end, we first independently prompt each LRM with queries in each target language to produce reasoning traces and final answers, which serve as inputs to all subsequent analyses.\footnote{We ensure that all reasoning traces and final answers are compliant with the query language using the \texttt{lingua} language detection tool (\url{https://pypi.org/project/lingua-language-detector/}).}
We then \ding{202} define a set of measurable reasoning features (\S\ref{sec:define}), quantify how each feature relates to accuracy via \ding{203} regression-based feature analysis (\S\ref{sec:feature}) and \ding{204} SAE analysis (\S\ref{sec:sae}), and use \ding{205} test-time selection to probe whether features can guide models toward higher accuracy (\S\ref{sec:testtime}), as illustrated in~\autoref{fig:main_figure}.
All prompts are provided in~\autoref{appendix:prompts}.

\begin{table}
\centering
\resizebox{\linewidth}{!}{%
    \begin{tabular}{l p{1.1\linewidth} l}
    \toprule
    \textbf{Feature} & \textbf{Description} & \textbf{Range} \\
    \toprule

    \rowcolor{multilingual!20}
    \multicolumn{3}{c}{\raisebox{-0.2em}{\includegraphics[height=1.1em]{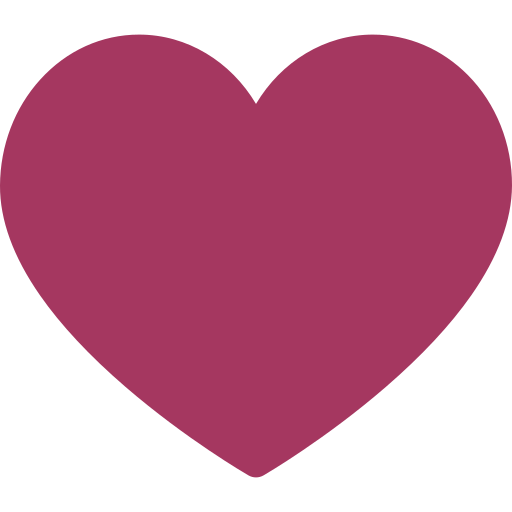}} \textcolor{multilingual}{\textbf{Multilingual Alignment}}} \\
    \midrule

    \textbf{COMET-QE} & Translation quality of non-English queries, measured with COMET-QE \citep{rei-etal-2020-comet} & [0,1] \\ 
    \textbf{Structural Similarity} & Structural alignment between English and non-English reasoning traces, measured via Smith-Waterman local sequence alignment algorithm \citep{smith1981identification} & [0,1] \\
    \textbf{Semantic Similarity} & Cosine similarity between English and non-English reasoning traces, using \textsc{LaBSE} as the embedding model \citep{feng-etal-2022-language} & [0,1] \\

    \midrule
    \rowcolor{intermediate!20}
    \multicolumn{3}{c}{\raisebox{-0.2em}{\includegraphics[height=1.1em]{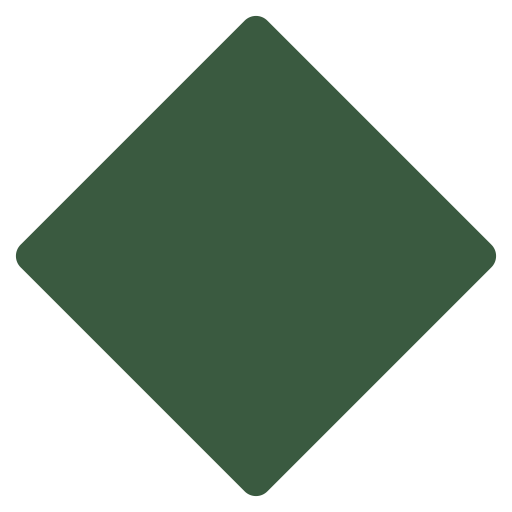}} \textcolor{intermediate}{\textbf{Reasoning Step}}} \\
    \midrule

    \textbf{Num. Steps} & Number of reasoning steps in the trace, segmented by \texttt{\textbackslash n\textbackslash n} \citep{xu2025mind} &  [0,$\infty)$ \\ 
    \textbf{Validity} & Logical consistency of a step with respect to its dependencies; proportion of dependency steps that entail the step (zeroed if any step is labeled as ``contradiction'') \citep{prasad-etal-2023-receval, you-etal-2025-probabilistic}, measured with an off-the-shelf NLI model \citep{manakul2023selfcheckgpt} & [0,1] \\
    \textbf{Direct Utility} & Degree to which reasoning steps directly contribute to getting the final answer \citep{lee-hockenmaier-2025-evaluating}; proportion of steps lying on a dependency path to the last step labeled ``Final Answer Emission'' (including the last step itself) & [0,1] \\
    \textbf{Indirect Utility} & Proportion of steps that lie on a dependency path to a direct utility step (i.e., they support steps that directly contribute to the final answer) & [0,1] \\
    \textbf{$\mathcal{V}$-Information} & Extent to which a reasoning trace $t$ increases the model ($\mathcal{V}$)'s confidence in the gold answer $a$; defined as $\mathcal{V}I(t \rightarrow a)=\mathrm{log}p_\mathcal{V}(a\mid q,t)-\mathrm{log}p_\mathcal{V}(a\mid q)$, where $q$ is the query & $(-\infty,\infty)$ \\

    \midrule
    \rowcolor{cognitive!20}
    \multicolumn{3}{c}{\raisebox{-0.2em}{\includegraphics[height=1.1em]{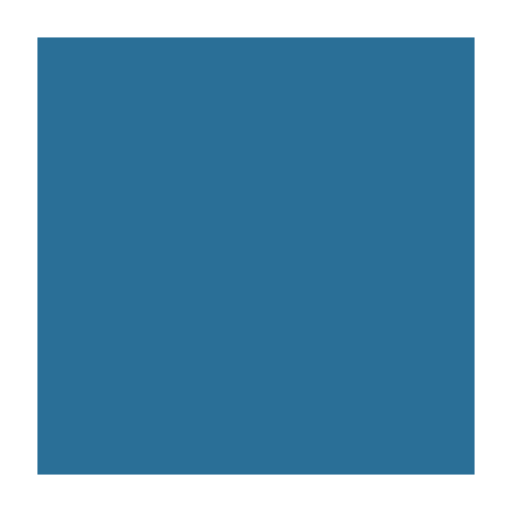}} \textcolor{cognitive}{\textbf{Reasoning Flow}}} \\
    \midrule

    \textbf{Self-Checking} & Steps that verify previous steps, check calculations, and re-confirm & [0,1] \\ 
    \textbf{Active Computation} & Steps that perform algebra, calculations, manipulations toward the answer & [0,1] \\
    \textbf{Problem Setup} & Steps that parse or rephrase the problem (initial reading or comprehension) & [0,1] \\
    \textbf{Plan Generation} & Steps that state or decide on a plan of action (often meta-reasoning) & [0,1] \\
    \textbf{Final Answer Emission} & Steps that explicitly state the final boxed answer or earlier sentences that contain the final answer & [0,1] \\
    \textbf{Fact Retrieval} & Steps that recall facts, formulas, problem details (without immediate computation) & [0,1] \\
    \textbf{Result Consolidation} & Steps that aggregate intermediate results, summarize, or prepare the final answer & [0,1] \\
    \textbf{Uncertainty Management} & Steps that express confusion, re-evaluate, and propose alternative plans (including backtracking) & [0,1] \\

    \bottomrule
    \end{tabular}
}
\caption{\textbf{Full list of measurable reasoning features.} We group features into \textcolor{multilingual}{\textbf{Multilingual Alignment}}, \textcolor{intermediate}{\textbf{Reasoning Step}}, and \textcolor{cognitive}{\textbf{Reasoning Flow}} dimensions. \textcolor{cognitive}{\textbf{Reasoning Flow}} features are adapted from \citet{bogdan2025thoughtanchorsllmreasoning}. All features are higher is better (↑). Implementation details are provided in~\autoref{appendix:features}.} 
\label{tab:features}
\end{table}


\subsection{Define Measurable Reasoning Features}
\label{sec:define}

We curate a set of 16 reasoning features spanning three dimensions of multilingual reasoning, applied to the traces generated for each query; the full list is provided in~\autoref{tab:features}:
\begin{itemize}[leftmargin=10pt, itemsep=1pt, parsep=-1pt]
    \item \raisebox{-0.2em}{\includegraphics[height=1.1em]{figures/logo/multilingual.png}} \textcolor{multilingual}{\textbf{Multilingual Alignment}} ($N$=3): Motivated by prior evidence that query understanding is a key bottleneck in multilingual reasoning \citep{zhu-etal-2024-question}, and by reward designs that privilege similarity to English traces \citep{zhang2026thinknativelyunlockingmultilingual, sutawika2026gainedtranslationprivilegedpairwise}, we define features that capture how faithfully queries and traces in other languages align with their English counterparts, both structurally and semantically.

    \item \raisebox{-0.2em}{\includegraphics[height=1.1em]{figures/logo/intermediate.png}} \textcolor{intermediate}{\textbf{Reasoning Step}} ($N$=5): To test whether step quality measures developed for English traces transfer to other languages (\S\ref{sec:related_work1}), we define step-level features that quantify trace length, logical consistency, informativeness, and usefulness. Dependencies are annotated with \textsc{GPT-4o} \citep{openai2024gpt4ocard} and per-step scores are aggregated by averaging.

    \item \raisebox{-0.2em}{\includegraphics[height=1.1em]{figures/logo/cognitive.png}} \textcolor{cognitive}{\textbf{Reasoning Flow}} ($N$=8): Following \citet{bogdan2025thoughtanchorsllmreasoning}, we annotate eight high-level cognitive-behavioral patterns characterizing the model's reasoning flow (e.g., planning, self-checking, fact retrieval) using \textsc{GPT-4o}.\footnote{Prior work shows that \textsc{GPT-4o} is strong at both (i) accurately annotating cognitive-behavioral tags for individual reasoning steps \citep{bogdan2025thoughtanchorsllmreasoning} and (ii) detecting dependencies in reasoning traces \citep{mukherjee2025premiseaugmented}. Our human verification of \textsc{GPT-4o} annotations on a subset of languages and traces confirms high agreement with human judgments (Appendix~\ref{appendix:human}).} Each feature is represented as the proportion of steps in the trace assigned that tag.
\end{itemize}

\subsection{\textcolor{feature}{Feature Analysis}}
\label{sec:feature}

We analyze how each measurable reasoning feature relates to final answer accuracy using univariate logistic regression \citep{movva2025whatshumanfeedbacklearning}, applied separately per language and feature over the generated traces. 
For each query, let $y \in \{0,1\}$ denote whether the model's final answer is correct and let $z_j \in \mathbb{R}$ denote the value of feature $j$. We first standardize each continuous feature to zero mean and unit variance within a language:
\begin{equation}
\tilde{z}_j = \frac{z_j-\mu_j}{\sigma_j}.
\end{equation}
For each model and language $\ell$, we fit a univariate logistic regression for each feature $j$:\footnote{We use the univariate setup for two reasons: (i) it isolates each feature's marginal relationship to accuracy without interference from other, potentially correlated features (e.g., direct and indirect utility) \citep{williams2012margins, long2014regression}, and (ii) it avoids multicollinearity issues that can obscure individual effect sizes \citep{Kim2019Multicollinearity}. We report consistent trends with smaller effect sizes under multivariate logistic regression in Appendix~\ref{appendix:multivariate}.}
\begin{equation}
P(y=1\mid\tilde{z}_{j,\ell})=\sigma(\alpha_\ell+\beta_{j,\ell} \tilde{z}_{j,\ell}),
\end{equation}
where $\sigma(\cdot)$ is the logistic sigmoid, $\alpha_\ell$ is a language-specific intercept, and $\beta_{j,\ell}$ captures the association between feature $j$ and accuracy in language $\ell$. 
We summarize the effect of feature $j$ by the discrete change in predicted accuracy when moving from one standard deviation below ($-1$) to one above ($+1$) the mean:
\begin{equation}
\Delta\mathrm{Acc}_{j,\ell}=\hat{y}_{j,\ell}\!\mid_{\tilde{z}_{j,\ell}=+1}-\hat{y}_{j,\ell}\!\mid_{\tilde{z}_{j,\ell}=-1},
\end{equation}
where $\hat{y}_{j,\ell}=P(y=1\mid\tilde{z}_{j,\ell})$. 
Intuitively, a positive $\Delta\mathrm{Acc}_{j,\ell}$ indicates that higher values of feature $j$ are associated with higher predicted accuracy in language $\ell$.

To assess whether a feature's effect differs significantly between English and other languages, we fit a pooled interaction logistic regression with an interaction term for each feature $j$ across all languages:
\begin{equation}
P(y=1 \mid \tilde{z}_j, \mathrm{en}) = \sigma\bigl(\alpha + \beta_1\,\mathrm{en} + \beta_2\,\tilde{z}_j + \beta_3\,(\mathrm{en} \cdot \tilde{z}_j)\bigr),
\end{equation}
where $\mathrm{en} \in \{0,1\}$ indicates whether the reasoning trace is from English queries. 
We report Wald-style $p$-values for the interaction coefficient $\beta_3$, which tests whether the association between feature $j$ and accuracy differs significantly between English and non-English traces.

\subsection{\textcolor{sae}{Sparse Autoencoder (SAE) Analysis}}
\label{sec:sae}

We complement the hand-designed feature analysis with a finer-grained, representation-driven approach by training sparse autoencoders (SAEs) over multilingual reasoning traces.
Following recent work using SAEs for hypothesis generation from text \citep{movva2025whatshumanfeedbacklearning, movva2025sparse}, we treat each reasoning trace as input and final answer accuracy as the target, and learn a set of sparse, interpretable latent concepts that explain variation in accuracy. 

For each model and language, we chunk each reasoning trace into segments of up to 400 words, encode each chunk with \textsc{LaBSE} embeddings \citep{feng-etal-2022-language}, and train a Batch TopK SAE \citep{bussmann2024batchtopk} to reconstruct them.
Each chunk inherits the accuracy label of its parent trace.
We then identify predictive SAE neurons via a correlation-based criterion: for each neuron, we compute the Pearson correlation between its activation vector and the binary accuracy label, and retain the top-20 neurons by absolute correlation.

To interpret each neuron, we prompt \textsc{GPT-4o} with the 10 chunks that most strongly activate it alongside 10 randomly sampled non-activating chunks, and ask for a short natural language description of the distinguishing concept.
For each concept, we report its separation score (the difference in accuracy when the concept is present vs.\ absent) and its prevalence (the fraction of examples in which it appears). 
Finally, we compare the learned concepts to our features to assess the extent to which SAEs recover the same underlying patterns.\footnote{Ablations used to select this specific SAE configuration are provided in Appendix~\ref{appendix:ablation}.}

\subsection{\textcolor{testtime}{Test-Time Selection}}
\label{sec:testtime}

We cast test-time selection as a best-of-$n$ problem \citep{charniak2005coarse, lightman2024lets, wang-etal-2024-math, rajaee-etal-2026-best}, using our reasoning features as selection policies to probe whether they can help re-rank traces for better multilingual reasoning at inference time.
Specifically, for a fixed model and language, we generate $n=32$ candidate reasoning traces per query under the same prompt and decoding setup, drawing 8 independent samples at each temperature $t\in\{0.3, 0.6, 0.8, 1.0\}$ \citep{feng2025characterizeseffectivereasoningrevisiting}.
Given these 32 candidates, we select a subset of 6 hand-designed features with the most salient negative, neutral, and positive $\Delta \mathrm{Acc}$ values as selection policies. For a given feature $j$, we use its value as a selection score, re-rank all candidates accordingly, and take the top-scoring trace as the final output.

We report pass@1, the fraction of queries for which the selected trace yields a correct final answer, and compare against a random-selection baseline to quantify how effectively each feature serves as a test-time selection policy.

\section{Experiment Setup}
\label{sec:setup}

\noindent\textbf{Dataset.}
We evaluate on two multilingual mathematical reasoning benchmarks of varying difficulty: MGSM-Rev2 \citep{peter2025mindgapnottranslation} and AIME 2024--25 \citep{qi-etal-2025-models}. 
MGSM-Rev2 is a revised version of MGSM \citep{shi2022languagemodelsmultilingualchainofthought} that corrects translation errors and ambiguities, updating 15.8\% of queries on average; it contains human-translated middle-school-level problems. AIME consists of challenging high-school-level competition problems originally written in English and machine-translated into other languages using \textsc{GPT-4o-mini}. Detailed dataset statistics are provided in Appendix~\autoref{tab:data_statistics}-\ref{tab:mgsmrev2_stats}.


\noindent\textbf{Languages.}
We study ten languages representing diverse resource levels, language families, writing scripts, and linguistic typologies: Bengali (bn), English (en), German (de), Spanish (es), French (fr), Russian (ru), Swahili (sw), Telugu (te), Thai (th), and Chinese (zh).
Per-language characteristics are detailed in Appendix~\autoref{tab:lang_statistics}.

\noindent\textbf{Models.}
We use four open-weight LRMs varying in size, degree of multilinguality, and training data: \textsc{Distill-Qwen 1.5b}, \textsc{7b} \citep{deepseekai2025deepseekr1incentivizingreasoningcapability}, \textsc{Qwen-3 4b}, and \textsc{8b} \citep{yang2025qwen3}.
Model details are provided in Appendix~\autoref{tab:model_statistics}.

\section{Results}
\label{sec:results}

\begin{figure*}
    \centering
    \includegraphics[width=\linewidth]{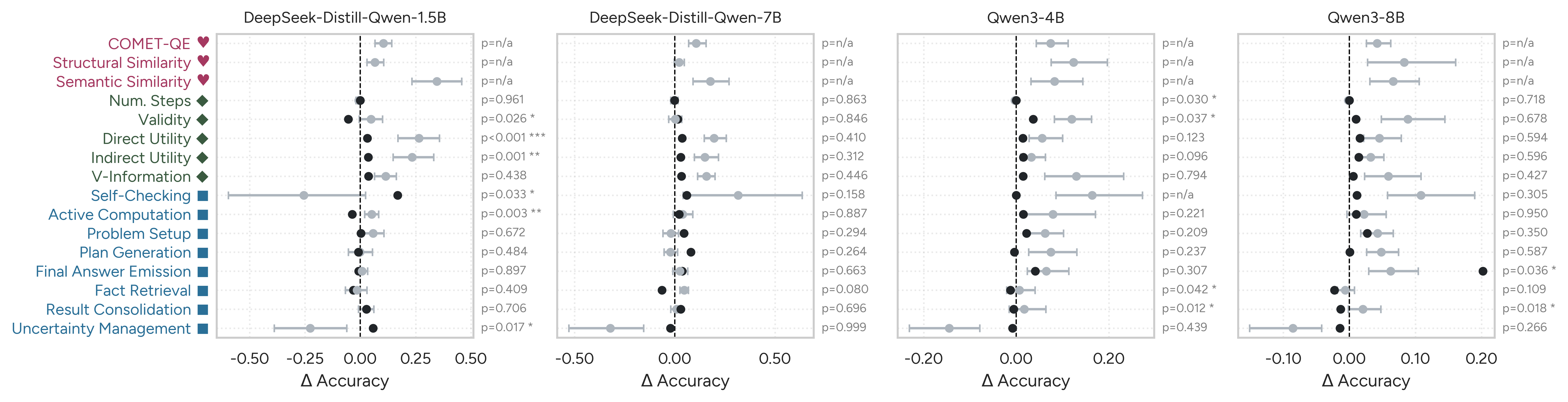}
    \includegraphics[width=\linewidth]{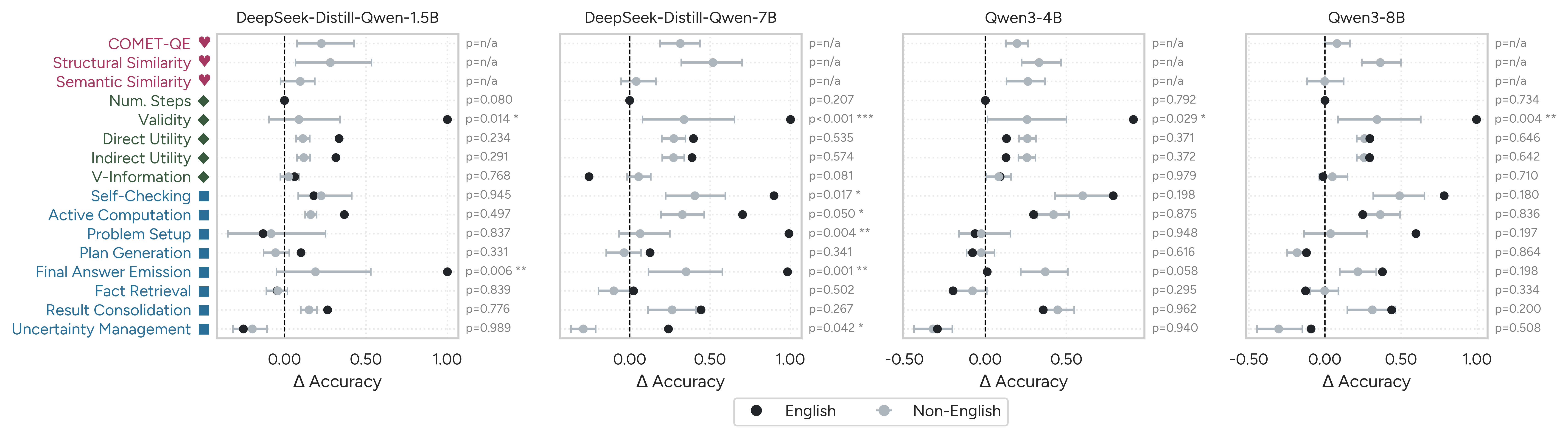}
    \caption{\textbf{English versus non-English feature analysis results.} Top: MGSM-Rev2; bottom: AIME. The $y$-axis lists each measurable reasoning feature; the $x$-axis shows its effect on predicted accuracy ($\Delta \mathrm{Acc}$), where positive values indicate that higher feature values are associated with higher accuracy. 
    We only report non-English values for \textcolor{multilingual}{\textbf{Multilingual Alignment}} features. 
    We report $p$-values for English to non-English difference on the right, using *: significant with $p$ $<$ 0.05; **: $p$ $<$ 0.01; ***: $p$ $<$ 0.001; non-marked: not statistically significant.}
    \label{fig:res1}
\end{figure*}

We begin by comparing feature analysis results for reasoning traces from English and non-English queries, then examine per-language effects (\S\ref{sec:rq1}). We test whether our SAE analysis recovers these findings (\S\ref{sec:rq2}) and align them with our test-time selection results (\S\ref{sec:rq3}).

\subsection{RQ1: What Characterizes \textit{Good} Multilingual Reasoning? [\textcolor{feature}{Feature Analysis}]}
\label{sec:rq1}

\autoref{fig:res1} shows the feature analysis results for English versus non-English languages on MGSM-Rev2 and AIME. We highlight several interesting findings below:

\noindent\textbf{Feature analysis recovers patterns from prior work.}
The number of reasoning steps feature (\textcolor{intermediate}{num. steps}) is associated with near-zero $\Delta \mathrm{Acc}$ for both English and non-English languages, suggesting that the previously reported weak relationship between trace length and accuracy in English \citep{vanhoyweghen2025lexicalhintsaccuracyllm} extends to other languages.
We also find that all \textcolor{multilingual}{\textbf{Multilingual Alignment}} features induce positive $\Delta \mathrm{Acc}$: positive \textcolor{multilingual}{COMET-QE} values suggest that accurate translation of non-English queries is crucial, consistent with prior findings that query understanding is a key bottleneck in multilingual reasoning \citep{peter2025mindgapnottranslation, kang2026multilingualreasoninggapsemerge}, and \textcolor{multilingual}{semantic similarity} helps explain the accuracy gains from English-similarity-based training objectives \citep{zhang2026thinknativelyunlockingmultilingual, sutawika2026gainedtranslationprivilegedpairwise}. 

We further find that \textcolor{multilingual}{structural similarity} to English traces is strongly associated with accuracy. 
On the more challenging AIME benchmark, \textcolor{multilingual}{structural similarity} yields higher $\Delta \mathrm{Acc}$ than \textcolor{multilingual}{semantic similarity} for all four LRMs, plausibly because AIME traces are much longer on average (275 steps) than MGSM-Rev2 traces (19 steps), making local sequence matching easier than aligning the overall semantics.


\begin{figure*}
    \centering
    \includegraphics[width=\linewidth]{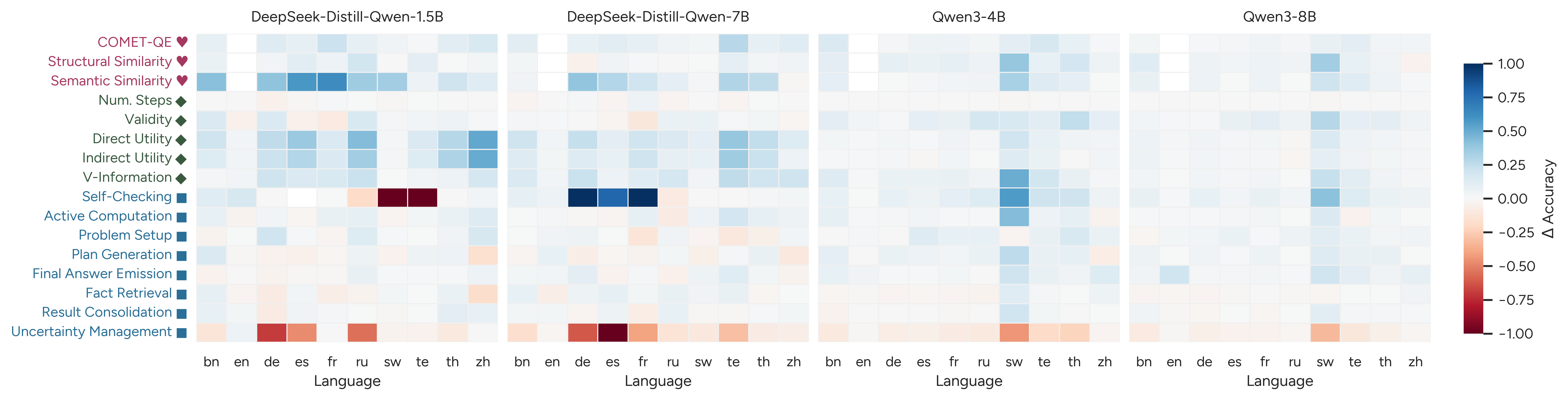}
    \includegraphics[width=\linewidth]{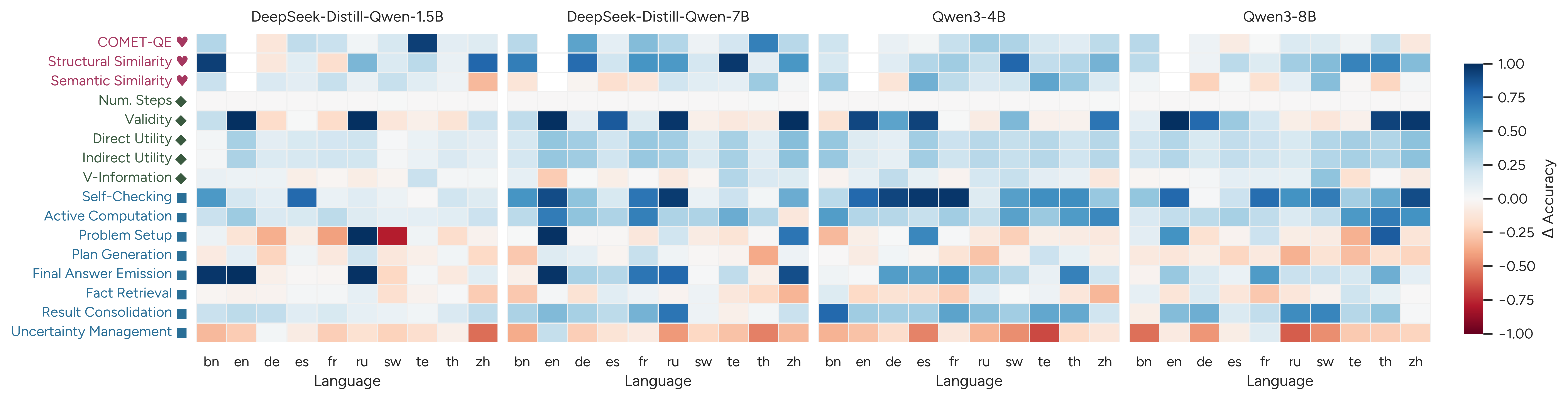}
    \caption{\textbf{Per-language feature analysis results.} Top: MGSM-Rev2; bottom: AIME. Raw feature values and accuracies for each language are provided in Appendix~\ref{appendix:raw_results}.}
    \label{fig:res2}
\end{figure*}

\noindent\textbf{Features show near-zero effects for English MGSM-Rev2 traces.}
As shown in the top panel of ~\autoref{fig:res1}, English traces on MGSM-Rev2 cluster tightly around near-zero $\Delta \mathrm{Acc}$, indicating that none of our features strongly predicts accuracy. This contrasts with non-English traces on the same dataset and with traces on AIME.
A plausible explanation is that MGSM-Rev2 queries are easy enough in English that LRMs frequently solve problems via latent reasoning with minimal reliance on explicit trace behaviors. 
Indeed, this is consistent with prior evidence that models can compute answers directly in their latent representations and then use traces primarily to \textit{surface} the answer, especially for English and simpler benchmarks like MGSM-Rev2 \citep{liu2026largereasoningmodelsnot}.

\noindent\textbf{Most features share directional effects across English and other languages.}
All \textcolor{intermediate}{\textbf{Reasoning Step}} features, including \textcolor{intermediate}{validity}, \textcolor{intermediate}{direct} and \textcolor{intermediate}{indirect utility}, and \textcolor{intermediate}{$\mathcal{V}$-Information}, generally induce positive $\Delta \mathrm{Acc}$ for both English and other languages across all models. 
On MGSM-Rev2, non-English languages typically show larger effect sizes, plausibly because English already benefits from strong latent reasoning. On AIME, the pattern reverses, with \textcolor{intermediate}{validity} consistently showing a significantly stronger effect for English.

\textcolor{cognitive}{\textbf{Reasoning Flow}} features follow a similar pattern: English and non-English traces largely share the same direction of effect, with variation in magnitude.
For instance, traces with more steps performing calculations (\textcolor{cognitive}{active computation}) or aggregating intermediate results (\textcolor{cognitive}{result consolidation}) are generally associated with positive $\Delta \mathrm{Acc}$, while steps expressing confusion or exploring alternatives (\textcolor{cognitive}{uncertainty management}) tend to yield negative $\Delta \mathrm{Acc}$.

\noindent\textbf{Language-level analysis reveals conflicts.}
Decomposing feature effects by language in~\autoref{fig:res2} reveals patterns that diverge from the aggregate English vs. non-English view.
On MGSM-Rev2 (top), while most features behave consistently, several show clear conflicts between English and other specific languages: for instance, more \textcolor{cognitive}{self-checking} steps improve $\Delta \mathrm{Acc}$ for English but are associated with negative $\Delta \mathrm{Acc}$ in Swahili and Telugu.
These conflicts are more pronounced on AIME (bottom), where language-level divergences are more frequent: \textcolor{intermediate}{validity} is near one for English but can be negatively associated with accuracy in other languages, and more steps explicitly stating the final answer (\textcolor{cognitive}{final answer emission}) are beneficial in English yet sometimes harmful in other languages.
We show several qualitative examples in Appendix~\ref{appendix:qualitative}.

Together, our feature analysis shows that most reasoning features have consistent directional effects, but their magnitudes vary substantially and can even conflict across languages.

\begin{table}
\centering
\resizebox{\linewidth}{!}{%
    \begin{tabular}{l p{1.1\linewidth} llll}
    \toprule
    \textbf{Lang.} & \textbf{Concept} (\hlpref{↑ preferred}, \hldispref{↓ dispreferred}) & \textbf{Sep.} & \textbf{$p$} & \textbf{Prev.} & \textbf{G/T} \\
    \toprule

    \rowcolor{gray!20}
    \multicolumn{6}{c}{\textbf{{MGSM-Rev2}}} \\
    \midrule

    \textbf{Bengali} & \cellcolor{lightred!80}Explicitly questions or re-interprets problem statement for contradiction or inconsistencies & \cellcolor{lightred!80}$-$39\% & \cellcolor{lightred!80}* & \cellcolor{lightred!80}13\% & \cellcolor{lightred!80}G \\
    \textbf{Bengali} & \cellcolor{lightred!80}Discuss potential translation errors in the problem statement & \cellcolor{lightred!80}$-$28\% & \cellcolor{lightred!80}* & \cellcolor{lightred!80}79\% &
    \cellcolor{lightred!80}G \\
    \textbf{English} & \cellcolor{lightgreen!70}Engages in self-questioning and reconsideration of problem interpretations & \cellcolor{lightgreen!70}$+$24\% & \cellcolor{lightgreen!70}** & \cellcolor{lightgreen!70}92\% &
    \cellcolor{lightgreen!70}G \\
    \textbf{German} & \cellcolor{darkred!80}Repeats the same phrase or reasoning step multiple times verbatim within the trace & \cellcolor{darkred!80}$-$82\% & \cellcolor{darkred!80}** & 
    \cellcolor{darkred!80}13\% &
    \cellcolor{darkred!80}G \\
    \textbf{Swahili} & \cellcolor{lightgreen!30}Include formatted final answer statement using ``**Final Answer**'' with a boxed numeric result & \cellcolor{lightgreen!30}$+$10\% & 
    \cellcolor{lightgreen!30}* & 
    \cellcolor{lightgreen!30}19\% &
    \cellcolor{lightgreen!30}G \\
    \textbf{Telugu} & \cellcolor{lightgreen!30}Translates problem statement from Telugu to English to clarify ambiguous terms & \cellcolor{lightgreen!30}$+$2\% & 
    \cellcolor{lightgreen!30}*** & \cellcolor{lightgreen!30}50\% &
    \cellcolor{lightgreen!30}G \\
    \textbf{Thai} & \cellcolor{lightred!80}Mixes multiple languages within the reasoning trace & \cellcolor{lightred!80}$-$36\% & 
    \cellcolor{lightred!80}* & \cellcolor{lightred!80}90\% & 
    \cellcolor{lightred!80}G \\
    \textbf{Chinese} & \cellcolor{lightgreen!70}Breaks down a multi-step into explicitly named sequential parts, with ordinal adverbs such as ``first'', ``second'', ``next'' and ``finally'' & \cellcolor{lightgreen!70}$+$31\% &
    \cellcolor{lightgreen!70}** &
    \cellcolor{lightgreen!70}54\% & 
    \cellcolor{lightgreen!70}G \\

    \midrule
    \rowcolor{gray!20}
    \multicolumn{6}{c}{\textbf{{AIME}}} \\
    \midrule

    \textbf{Bengali} & \cellcolor{lightred!60}Makes repeated corrections to earlier reasoning steps throughout the trace & \cellcolor{lightred!60}$-$12\% & 
    \cellcolor{lightred!60}** &
    \cellcolor{lightred!60}90\% & 
    \cellcolor{lightred!60}G \\
    \textbf{English} & \cellcolor{lightgreen!30}Uses iterative reasoning and re-evaluation of previous steps in the calculations & \cellcolor{lightgreen!30}$+$14\% & 
    \cellcolor{lightgreen!30}* &
    \cellcolor{lightgreen!30}99\% &
    \cellcolor{lightgreen!30}T \\
    \textbf{German} & \cellcolor{lightgreen!30}Explicitly uses logarithmic identities to manipulate equations and solve for variables & \cellcolor{lightgreen!30}$+$11\% & 
    \cellcolor{lightgreen!30}* &
    \cellcolor{lightgreen!30}18\% & 
    \cellcolor{lightgreen!30}T \\
    \textbf{Spanish} & \cellcolor{lightred!60}Frequently uses the phrase ``Wait'' to indicate reconsideration of previous steps & \cellcolor{lightred!60}$-$17\% & 
    \cellcolor{lightred!60}** &
    \cellcolor{lightred!60}83\% & 
    \cellcolor{lightred!60}G \\
    \textbf{Chinese} & \cellcolor{lightgreen!70}Uses a series of logical deductions and checks to verify previous calculations & \cellcolor{lightgreen!70}$+$38\% & 
    \cellcolor{lightgreen!70}*** &
    \cellcolor{lightgreen!70}62\% & 
    \cellcolor{lightgreen!70}T \\ 

    \bottomrule
    \end{tabular}
}
\caption{\textbf{Example of reasoning concepts discovered in SAE analysis.} \textbf{Sep.}: separation score; \textbf{$p$}: $p$-value for separation score (*: significant with $p$ $<$ 0.05; **: $p$ $<$ 0.01; ***: $p$ $<$ 0.001); \textbf{Prev.}: feature prevalence score (i.e., how often the concept occurs); \textbf{G/T}: General (G) or task-specific (T) concept.
We show consistency of SAE concepts across models, languages, and datasets in Appendix~\ref{appendix:consistency}.}
\label{tab:res3}
\end{table}

\subsection{RQ2: Can We Automatically Discover Reasoning Features? [\textcolor{sae}{SAE Analysis}]}
\label{sec:rq2}

For each dataset and language, we select the top-3 latent reasoning concepts with the highest absolute separation scores (i.e., how much accuracy differs when a concept is present vs. absent) and statistically significant separation $p$-values.
\autoref{tab:res3} presents a subset of concepts that are least overlapping across languages, several of which we discuss below: 

\noindent\textbf{SAE analysis confirms findings from feature analysis.}
SAEs trained over multilingual reasoning traces largely recover the patterns identified by our hand-designed features.
On MGSM-Rev2, concepts associated with \textcolor{cognitive}{uncertainty management}, such as ``questioning the problem statement'' ($-$39\%) or ``repeating the same phrase'' ($-$82\%), tend to show negative $\Delta \mathrm{Acc}$ for most non-English languages, while the corresponding feature is associated \textit{only} with positive $\Delta \mathrm{Acc}$ for English, aligning with the preferred concept ``engages in self-questioning and reconsideration of problem interpretations'' ($+$24\%).
Concepts related to high \textcolor{multilingual}{COMET-QE} are similarly consistent: ``discussing potential translation errors'' ($-$28\%) is dispreferred for non-English traces while ``translating the problem into English'' ($+$2\%) is preferred.
We observe analogous patterns in AIME: behaviors linked to \textcolor{cognitive}{uncertainty management} such as ``making repeated corrections'' ($-$12\%) or ``frequently using the wait phrase'' ($-$17\%) are likewise dispreferred, consistent with their negative $\Delta \mathrm{Acc}$ from feature analysis, while behaviors related to \textcolor{cognitive}{active computation} such as ``using logarithmic identities'' ($+$11\%), are preferred.

\noindent\textbf{SAE analysis discovers new reasoning patterns.}
Beyond corroborating our feature analysis, SAE also refines and extends it with finer-grained reasoning patterns.
For example, while our feature analysis captures the benefits of informative reasoning steps (via \textcolor{intermediate}{validity} and \textcolor{intermediate}{utility}) likely associated with \textcolor{cognitive}{active computation} or \textcolor{cognitive}{result consolidation}, SAE reveals \textit{how} these behaviors concretely manifest in traces, such as ``breaking down multi-step reasoning with ordinal adverbs'' ($+$31\%) or ``using a series of logical deductions'' ($+$38\%). 
It also surfaces behaviors not covered by our hand-designed feature set, including ``mixing multiple languages within a single reasoning trace'' ($-$36\%), associated with lower accuracy.

\subsection{RQ3: Can Features Re-rank Traces For Higher Accuracy? [\textcolor{testtime}{Test-Time Selection}]}
\label{sec:rq3}

\begin{figure*}
    \centering
    \includegraphics[width=\linewidth]{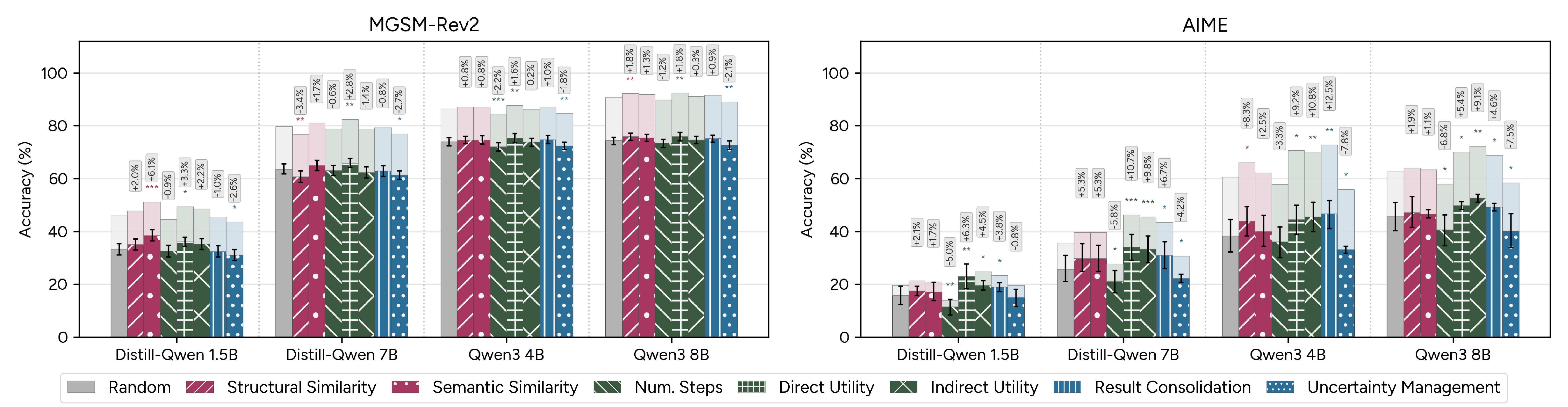}
    \caption{\textbf{Pass@1 per model using each feature as test-time selection policy.} Solid bars show results aggregated over languages other than English; light-shaded bars show English results as reference. For solid bars, error bars show 95\% bootstrap confidence intervals and we report $p$-values of the paired bootstrap test against the random selection baseline, using *: significant with $p$ $<$ 0.05; **: $p$ $<$ 0.01; ***: $p$ $<$ 0.001; non-marked: not statistically significant.}
    \label{fig:res4}
\end{figure*}

Having identified features associated with higher or lower accuracy through both feature (\S\ref{sec:rq1}) and SAE analysis (\S\ref{sec:rq2}), we now ask whether these features can be used to re-rank traces for better multilingual reasoning at inference time. Results are shown in~\autoref{fig:res4}.

On MGSM-Rev2 (left), \textcolor{multilingual}{semantic similarity} to English traces is a competitive selector over the random baseline, especially for \textsc{Distill-Qwen 1.5b}, but does not consistently improve accuracy across other models.
Aside from \textcolor{cognitive}{uncertainty management}, which consistently yields significantly lower accuracy than the baseline (at most -2.7\%), most features achieve similar pass@1 to random selection, consistent with our earlier finding that MGSM-Rev2 shows near-zero effects across features (\S\ref{sec:rq1}).
This suggests that on relatively easy benchmarks like MGSM-Rev2, feature-based selection offers limited gains over random choice, and that improving multilingual reasoning performance may instead require stronger changes to training objectives or model architectures.

The more challenging AIME benchmark (right) makes these differences more pronounced.
While \textcolor{multilingual}{structural} and \textcolor{multilingual}{semantic similarity} offer modest, non-significant gains, \textcolor{intermediate}{direct utility}, \textcolor{intermediate}{indirect utility}, and \textcolor{cognitive}{result consolidation} consistently yield significantly higher accuracy than the random baseline across all models.
Features with neutral or negative $\Delta \mathrm{Acc}$, such as \textcolor{intermediate}{num.steps} and \textcolor{cognitive}{uncertainty management}, also show correspondingly stronger effects in the direction of lower accuracy.\footnote{To assess the practicality of our test-time selection policies, we ablate $n$, the number of sampled generations, across the range of temperatures used. Detailed results are provided in Appendix~\ref{appendix:n_ablation}.}

Overall, these results confirm that our features provide useful signals for selecting higher-quality traces, particularly on harder benchmarks: selecting by \textcolor{intermediate}{direct utility} alone improves accuracy by up to 10\% with a simple inference-time strategy.
At the same time, the limited gains from \textcolor{multilingual}{semantic similarity} challenge reward designs that uncritically favor English-like traces, and point instead toward objectives that accommodate language-specific reasoning behaviors.


\section{Discussion \& Conclusion}
\label{sec:implications}

We discuss important implications of our findings for advancing multilingual reasoning.

\noindent\textbf{Reasoning Benchmark Design.}
(1) The positive $\Delta \mathrm{Acc}$ associated with \textcolor{multilingual}{COMET-QE} underscores that benchmark queries in languages other than English should be accurately rendered, ideally via human translation \citep{shi2022languagemodelsmultilingualchainofthought} or human verification on a representative subset \citep{chen-etal-2024-breaking, wang2025polymath, dobler2026multilingualreasoninggymmultilingual}.
(2) Our findings suggest that \textcolor{intermediate}{\textbf{Reasoning Step}} measures of step-level correctness \citep{zeng2025mrgsmk, li2025thinkbenchevaluatingthinkingefficiency}, originally developed for English meta-reasoning benchmarks \citep{mirzadeh2024gsm, xia2025evaluating, zheng2025processbench, song-etal-2025-prmbench}, transfer well across languages: these features, especially \textcolor{intermediate}{utility}-based signals, consistently induce positive $\Delta \mathrm{Acc}$, supporting their use as language-agnostic evaluation tools.

\noindent\textbf{Reward Model Design.} (1) Both our feature analysis (\S\ref{sec:rq1}) and test-time selection experiments (\S\ref{sec:rq3}) show that rewarding \textcolor{multilingual}{semantic similarity} to English traces, while competitive, is not a universally effective signal. 
Alternative signals, such as \textcolor{intermediate}{direct} and \textcolor{intermediate}{indirect utility} or \textcolor{cognitive}{result consolidation}, can yield stronger multilingual reasoning performance, especially on challenging benchmarks like AIME. 
This motivates training objectives that explicitly incorporate these signals rather than relying on \textcolor{multilingual}{semantic similarity} alone.
(2) As shown in~\autoref{fig:res2}, the same feature can induce conflicting effects across languages, highlighting the need to encourage language-specific reasoning patterns. Together with recent work on adaptive reasoning \citep{wu2025arm, gao2026explangimprovedexplorationexploitation}, we point toward adaptive reward designs that flexibly decide when to remain in the original language, when to translate, and which feature signals to prioritize.

Overall, we present a systematic study of what characterizes effective reasoning in multilingual settings by decomposing reasoning traces into measurable features and relating them to accuracy. 
We find mostly consistent directional effects but language-varying, and sometimes conflicting, effect sizes. 
Test-time selection shows that while rewarding \textcolor{multilingual}{semantic similarity} to English traces is often competitive, \textcolor{intermediate}{utility} and \textcolor{cognitive}{active computation} features can better re-rank traces for higher accuracy in some languages. 
Our findings challenge English-centric assumptions about what constitutes ``good'' reasoning, and argue for multilingual benchmarks and adaptive reward designs that explicitly accommodate language-specific reasoning patterns.

\section{Limitations}

\paragraph{Limited scope.}
Our analysis is constrained to the current experimental setup. First, our hand-designed feature set includes 16 measurable reasoning features, which is necessarily non-exhaustive and does not cover the full space of potentially informative signals for multilingual reasoning. 

Second, we focus exclusively on mathematical reasoning benchmarks, which offer several advantages: (i) parallel queries across a representative set of languages, (ii) a clear verifiable correctness signal (accuracy), and (iii) typically multi-step problems that make them a natural testbed for our \textcolor{cognitive}{\textbf{Reasoning Flow}} features. 
While we believe the features that capture general reasoning behaviors will generalize to other tasks (e.g., \textcolor{multilingual}{\textbf{Multilingual Alignment}} and \textcolor{intermediate}{\textbf{Reasoning Step}} features), others may not (e.g., \textcolor{cognitive}{active computation}). Also, since the most informative feature space may differ across domains, additional features may need to be introduced to accommodate the nature of reasoning in other settings.

Finally, our experiments are limited to four LRMs, so the observed patterns may not hold for other architectures or training regimes; we view our work as an initial step toward broader investigations.
Also, while our test-time selection experiments serve as a proxy for assessing how re-ranking reasoning traces based on features impacts accuracy, we leave further causal analyses such as intervening on model representations for future work.

\paragraph{Conflation of feature set.}
Due to the nature of multilingual reasoning benchmarks, which are typically constructed through machine translation of English queries, translation quality may have downstream effects: translation errors can impair prompt comprehension, which in turn may degrade reasoning quality. 
We aim to minimize such conflation through two measures. First, we verify the translation quality of AIME queries (a machine-translated dataset) in Appendix~\autoref{tab:mgsmrev2_stats}, reporting a comparably high average COMET score (0.810) relative to human-translated MGSM-Rev2 (0.851). Second, our main feature analysis uses univariate logistic regression to isolate each feature’s marginal relationship with accuracy, avoiding interference across features.
Yet, fully disentangling these effects is a fundamental challenge in the multilingual setting, which encourages work along this direction.

\paragraph{Reliance on \textsc{GPT-4o} annotation.}
Many of our core features, including \textcolor{intermediate}{\textbf{Reasoning Step}} features such as \textcolor{intermediate}{validity} and \textcolor{intermediate}{utility}, as well as \textcolor{cognitive}{\textbf{Reasoning Flow}} features, rely on \textsc{GPT-4o}-based annotation. 
This choice is motivated by prior evidence that \textsc{GPT-4o} can reliably perform fine-grained reasoning-trace tagging for mathematical problems in English using the same prompt template in Prompt C.1 \citep{bogdan2025thoughtanchorsllmreasoning}. 
To partially address its generalization to languages other than English, we conduct human verification on a small subset of traces from non-English queries and report high agreement with human judgments in Appendix~\ref{appendix:human}, yet a more comprehensive annotation remains future work.

\paragraph{Non-independent evaluation set.}
We use the same set of mathematical queries for both the feature analysis and the test-time selection experiments, as the limited number of test instances in each dataset (250 for MGSM-Rev2 and 60 for AIME) necessitates using the full set to observe statistically significant effects.

\section*{Acknowledgments}

We thank the members of the \textsc{CLIP} lab at the University of Maryland for their valuable feedback and support, with special thanks to Calvin Bao and Hieu Tran for their comments on an earlier draft.

\bibliography{colm2026_conference}

@misc{li2025thinkbenchevaluatingthinkingefficiency,
      title={THINK-Bench: Evaluating Thinking Efficiency and Chain-of-Thought Quality of Large Reasoning Models}, 
      author={Zhiyuan Li and Yi Chang and Yuan Wu},
      year={2025},
      eprint={2505.22113},
      archivePrefix={arXiv},
      primaryClass={cs.CL},
      url={https://arxiv.org/abs/2505.22113}, 
}

@inproceedings{
zeng2025mrgsmk,
title={{MR}-{GSM}8K: A Meta-Reasoning Benchmark for Large Language Model Evaluation},
author={Zhongshen Zeng and Pengguang Chen and Shu Liu and Haiyun Jiang and Jiaya Jia},
booktitle={The Thirteenth International Conference on Learning Representations},
year={2025},
url={https://openreview.net/forum?id=br4H61LOoI}
}

@inproceedings{
lightman2024lets,
title={Let's Verify Step by Step},
author={Hunter Lightman and Vineet Kosaraju and Yuri Burda and Harrison Edwards and Bowen Baker and Teddy Lee and Jan Leike and John Schulman and Ilya Sutskever and Karl Cobbe},
booktitle={The Twelfth International Conference on Learning Representations},
year={2024},
url={https://openreview.net/forum?id=v8L0pN6EOi}
}

@article{Wang2017Inconsistency,
  author  = {Wang, Hongyue and Peng, Jing and Wang, Bokai and Lu, Xiang and Zheng, Julia Z. and Wang, Kejia and Tu, Xin M. and Feng, Changyong},
  title   = {Inconsistency Between Univariate and Multiple Logistic Regressions},
  journal = {Shanghai Archives of Psychiatry},
  year    = {2017},
  volume  = {29},
  number  = {2},
  pages   = {124--128},
  doi     = {10.11919/j.issn.1002-0829.217031},
  issn    = {1002-0829}
}

@inproceedings{
wu2025arm,
title={{ARM}: Adaptive Reasoning Model},
author={Siye Wu and Jian Xie and Yikai Zhang and Aili Chen and Kai Zhang and Yu Su and Yanghua Xiao},
booktitle={The Thirty-ninth Annual Conference on Neural Information Processing Systems},
year={2025},
url={https://openreview.net/forum?id=z9oeQrcNh9}
}

@inproceedings{xia2025evaluating,
  title={Evaluating mathematical reasoning beyond accuracy},
  author={Xia, Shijie and Li, Xuefeng and Liu, Yixin and Wu, Tongshuang and Liu, Pengfei},
  booktitle={Proceedings of the AAAI Conference on Artificial Intelligence},
  volume={39},
  number={26},
  pages={27723--27730},
  year={2025}
}

@misc{gao2026explangimprovedexplorationexploitation,
      title={ExpLang: Improved Exploration and Exploitation in LLM Reasoning with On-Policy Thinking Language Selection}, 
      author={Changjiang Gao and Zixian Huang and Kaichen Yang and Jiajun Chen and Jixing Li and Shujian Huang},
      year={2026},
      eprint={2602.21887},
      archivePrefix={arXiv},
      primaryClass={cs.CL},
      url={https://arxiv.org/abs/2602.21887}, 
}

@inproceedings{
mukherjee2025premiseaugmented,
title={Premise-Augmented Reasoning Chains Improve Error Identification in Math reasoning with {LLM}s},
author={Sagnik Mukherjee and Abhinav Chinta and Takyoung Kim and Tarun Anoop Sharma and Dilek Hakkani Tur},
booktitle={Forty-second International Conference on Machine Learning},
year={2025},
url={https://openreview.net/forum?id=4tYckHNVXV}
}

@article{mirzadeh2024gsm,
  title={Gsm-symbolic: Understanding the limitations of mathematical reasoning in large language models},
  author={Mirzadeh, Iman and Alizadeh, Keivan and Shahrokhi, Hooman and Tuzel, Oncel and Bengio, Samy and Farajtabar, Mehrdad},
  journal={arXiv preprint arXiv:2410.05229},
  year={2024}
}

@inproceedings{song-etal-2025-prmbench,
    title = "{PRMB}ench: A Fine-grained and Challenging Benchmark for Process-Level Reward Models",
    author = "Song, Mingyang  and
      Su, Zhaochen  and
      Qu, Xiaoye  and
      Zhou, Jiawei  and
      Cheng, Yu",
    editor = "Che, Wanxiang  and
      Nabende, Joyce  and
      Shutova, Ekaterina  and
      Pilehvar, Mohammad Taher",
    booktitle = "Proceedings of the 63rd Annual Meeting of the Association for Computational Linguistics (Volume 1: Long Papers)",
    month = jul,
    year = "2025",
    address = "Vienna, Austria",
    publisher = "Association for Computational Linguistics",
    url = "https://aclanthology.org/2025.acl-long.1230/",
    doi = "10.18653/v1/2025.acl-long.1230",
    pages = "25299--25346",
    ISBN = "979-8-89176-251-0",
}

@misc{liu2026largereasoningmodelsnot,
      title={Large Reasoning Models Are (Not Yet) Multilingual Latent Reasoners}, 
      author={Yihong Liu and Raoyuan Zhao and Hinrich Schütze and Michael A. Hedderich},
      year={2026},
      eprint={2601.02996},
      archivePrefix={arXiv},
      primaryClass={cs.CL},
      url={https://arxiv.org/abs/2601.02996}, 
}

@misc{feng2025characterizeseffectivereasoningrevisiting,
      title={What Characterizes Effective Reasoning? Revisiting Length, Review, and Structure of CoT}, 
      author={Yunzhen Feng and Julia Kempe and Cheng Zhang and Parag Jain and Anthony Hartshorn},
      year={2025},
      eprint={2509.19284},
      archivePrefix={arXiv},
      primaryClass={cs.LG},
      url={https://arxiv.org/abs/2509.19284}, 
}

@inproceedings{lee-hockenmaier-2025-evaluating,
    title = "Evaluating Step-by-step Reasoning Traces: A Survey",
    author = "Lee, Jinu  and
      Hockenmaier, Julia",
    editor = "Christodoulopoulos, Christos  and
      Chakraborty, Tanmoy  and
      Rose, Carolyn  and
      Peng, Violet",
    booktitle = "Findings of the Association for Computational Linguistics: EMNLP 2025",
    month = nov,
    year = "2025",
    address = "Suzhou, China",
    publisher = "Association for Computational Linguistics",
    url = "https://aclanthology.org/2025.findings-emnlp.94/",
    doi = "10.18653/v1/2025.findings-emnlp.94",
    pages = "1789--1814",
    ISBN = "979-8-89176-335-7",
}

@article{yang2025qwen3,
  title={Qwen3 technical report},
  author={Yang, An and Li, Anfeng and Yang, Baosong and Zhang, Beichen and Hui, Binyuan and Zheng, Bo and Yu, Bowen and Gao, Chang and Huang, Chengen and Lv, Chenxu and others},
  journal={arXiv preprint arXiv:2505.09388},
  year={2025}
}

@article{gandhi2025cognitive,
  title={Cognitive behaviors that enable self-improving reasoners, or, four habits of highly effective stars},
  author={Gandhi, Kanishk and Chakravarthy, Ayush and Singh, Anikait and Lile, Nathan and Goodman, Noah D},
  journal={arXiv preprint arXiv:2503.01307},
  year={2025}
}

@article{li2025llms,
  title={LLMs Can Easily Learn to Reason from Demonstrations Structure, not content, is what matters!},
  author={Li, Dacheng and Cao, Shiyi and Griggs, Tyler and Liu, Shu and Mo, Xiangxi and Tang, Eric and Hegde, Sumanth and Hakhamaneshi, Kourosh and Patil, Shishir G and Zaharia, Matei and others},
  journal={arXiv preprint arXiv:2502.07374},
  year={2025}
}

@inproceedings{jiang-etal-2025-makes,
    title = "What Makes a Good Reasoning Chain? Uncovering Structural Patterns in Long Chain-of-Thought Reasoning",
    author = "Jiang, Gangwei  and
      Liu, Yahui  and
      Li, Zhaoyi  and
      Bi, Wei  and
      Zhang, Fuzheng  and
      Song, Linqi  and
      Wei, Ying  and
      Lian, Defu",
    editor = "Christodoulopoulos, Christos  and
      Chakraborty, Tanmoy  and
      Rose, Carolyn  and
      Peng, Violet",
    booktitle = "Proceedings of the 2025 Conference on Empirical Methods in Natural Language Processing",
    month = nov,
    year = "2025",
    address = "Suzhou, China",
    publisher = "Association for Computational Linguistics",
    url = "https://aclanthology.org/2025.emnlp-main.329/",
    doi = "10.18653/v1/2025.emnlp-main.329",
    pages = "6490--6514",
    ISBN = "979-8-89176-332-6",
}

@misc{do2025definesgoodreasoningllms,
      title={What Defines Good Reasoning in LLMs? Dissecting Reasoning Steps with Multi-Aspect Evaluation}, 
      author={Heejin Do and Jaehui Hwang and Dongyoon Han and Seong Joon Oh and Sangdoo Yun},
      year={2025},
      eprint={2510.20603},
      archivePrefix={arXiv},
      primaryClass={cs.AI},
      url={https://arxiv.org/abs/2510.20603}, 
}

@misc{ovalle2026begdifferunderstandingreasoninganswer,
      title={Beg to Differ: Understanding Reasoning-Answer Misalignment Across Languages}, 
      author={Anaelia Ovalle and Candace Ross and Sebastian Ruder and Adina Williams and Karen Ullrich and Mark Ibrahim and Levent Sagun},
      year={2026},
      eprint={2512.22712},
      archivePrefix={arXiv},
      primaryClass={cs.CL},
      url={https://arxiv.org/abs/2512.22712}, 
}

@misc{tam2025languagemattersmultilingualinput,
      title={Language Matters: How Do Multilingual Input and Reasoning Paths Affect Large Reasoning Models?}, 
      author={Zhi Rui Tam and Cheng-Kuang Wu and Yu Ying Chiu and Chieh-Yen Lin and Yun-Nung Chen and Hung-yi Lee},
      year={2025},
      eprint={2505.17407},
      archivePrefix={arXiv},
      primaryClass={cs.CL},
      url={https://arxiv.org/abs/2505.17407}, 
}

@misc{bogdan2025thoughtanchorsllmreasoning,
      title={Thought Anchors: Which LLM Reasoning Steps Matter?}, 
      author={Paul C. Bogdan and Uzay Macar and Neel Nanda and Arthur Conmy},
      year={2025},
      eprint={2506.19143},
      archivePrefix={arXiv},
      primaryClass={cs.LG},
      url={https://arxiv.org/abs/2506.19143}, 
}

@misc{zhang2026thinknativelyunlockingmultilingual,
      title={Think Natively: Unlocking Multilingual Reasoning with Consistency-Enhanced Reinforcement Learning}, 
      author={Xue Zhang and Yunlong Liang and Fandong Meng and Songming Zhang and Kaiyu Huang and Yufeng Chen and Jinan Xu and Jie Zhou},
      year={2026},
      eprint={2510.07300},
      archivePrefix={arXiv},
      primaryClass={cs.CL},
      url={https://arxiv.org/abs/2510.07300}, 
}

@misc{sutawika2026gainedtranslationprivilegedpairwise,
      title={Gained in Translation: Privileged Pairwise Judges Enhance Multilingual Reasoning}, 
      author={Lintang Sutawika and Gokul Swamy and Zhiwei Steven Wu and Graham Neubig},
      year={2026},
      eprint={2601.18722},
      archivePrefix={arXiv},
      primaryClass={cs.CL},
      url={https://arxiv.org/abs/2601.18722}, 
}

@misc{faisal2025aligningmultilingualreasoningverifiable,
      title={Aligning Multilingual Reasoning with Verifiable Semantics from a High-Resource Expert Model}, 
      author={Fahim Faisal and Kaiqiang Song and Song Wang and Simin Ma and Shujian Liu and Haoyun Deng and Sathish Reddy Indurthi},
      year={2025},
      eprint={2509.25543},
      archivePrefix={arXiv},
      primaryClass={cs.CL},
      url={https://arxiv.org/abs/2509.25543}, 
}

@misc{reddy2026effectscriptsformatsllm,
      title={The Effect of Scripts and Formats on LLM Numeracy}, 
      author={Varshini Reddy and Craig W. Schmidt and Seth Ebner and Adam Wiemerslage and Yuval Pinter and Chris Tanner},
      year={2026},
      eprint={2601.15251},
      archivePrefix={arXiv},
      primaryClass={cs.CL},
      url={https://arxiv.org/abs/2601.15251}, 
}

@inproceedings{wang-etal-2024-math,
    title = "Math-Shepherd: Verify and Reinforce {LLM}s Step-by-step without Human Annotations",
    author = "Wang, Peiyi  and
      Li, Lei  and
      Shao, Zhihong  and
      Xu, Runxin  and
      Dai, Damai  and
      Li, Yifei  and
      Chen, Deli  and
      Wu, Yu  and
      Sui, Zhifang",
    editor = "Ku, Lun-Wei  and
      Martins, Andre  and
      Srikumar, Vivek",
    booktitle = "Proceedings of the 62nd Annual Meeting of the Association for Computational Linguistics (Volume 1: Long Papers)",
    month = aug,
    year = "2024",
    address = "Bangkok, Thailand",
    publisher = "Association for Computational Linguistics",
    url = "https://aclanthology.org/2024.acl-long.510/",
    doi = "10.18653/v1/2024.acl-long.510",
    pages = "9426--9439",
}

@misc{vanhoyweghen2025lexicalhintsaccuracyllm,
      title={Lexical Hints of Accuracy in LLM Reasoning Chains}, 
      author={Arne Vanhoyweghen and Brecht Verbeken and Andres Algaba and Vincent Ginis},
      year={2025},
      eprint={2508.15842},
      archivePrefix={arXiv},
      primaryClass={cs.CL},
      url={https://arxiv.org/abs/2508.15842}, 
}

@inproceedings{qi-etal-2025-models,
    title = "When Models Reason in Your Language: Controlling Thinking Language Comes at the Cost of Accuracy",
    author = "Qi, Jirui  and
      Chen, Shan  and
      Xiong, Zidi  and
      Fern{\'a}ndez, Raquel  and
      Bitterman, Danielle  and
      Bisazza, Arianna",
    editor = "Christodoulopoulos, Christos  and
      Chakraborty, Tanmoy  and
      Rose, Carolyn  and
      Peng, Violet",
    booktitle = "Findings of the Association for Computational Linguistics: EMNLP 2025",
    month = nov,
    year = "2025",
    address = "Suzhou, China",
    publisher = "Association for Computational Linguistics",
    url = "https://aclanthology.org/2025.findings-emnlp.1103/",
    doi = "10.18653/v1/2025.findings-emnlp.1103",
    pages = "20279--20296",
    ISBN = "979-8-89176-335-7",
}

@misc{peter2025mindgapnottranslation,
      title={Mind the Gap... or Not? How Translation Errors and Evaluation Details Skew Multilingual Results}, 
      author={Jan-Thorsten Peter and David Vilar and Tobias Domhan and Dan Malkin and Markus Freitag},
      year={2025},
      eprint={2511.05162},
      archivePrefix={arXiv},
      primaryClass={cs.CL},
      url={https://arxiv.org/abs/2511.05162}, 
}

@misc{shi2022languagemodelsmultilingualchainofthought,
      title={Language Models are Multilingual Chain-of-Thought Reasoners}, 
      author={Freda Shi and Mirac Suzgun and Markus Freitag and Xuezhi Wang and Suraj Srivats and Soroush Vosoughi and Hyung Won Chung and Yi Tay and Sebastian Ruder and Denny Zhou and Dipanjan Das and Jason Wei},
      year={2022},
      eprint={2210.03057},
      archivePrefix={arXiv},
      primaryClass={cs.CL},
      url={https://arxiv.org/abs/2210.03057}, 
}

@misc{deepseekai2025deepseekr1incentivizingreasoningcapability,
      title={DeepSeek-R1: Incentivizing Reasoning Capability in LLMs via Reinforcement Learning}, 
      author={DeepSeek-AI},
      year={2025},
      eprint={2501.12948},
      archivePrefix={arXiv},
      primaryClass={cs.CL},
      url={https://arxiv.org/abs/2501.12948}, 
}

@article{zhang-etal-2023-miracl,
    title = "{MIRACL}: A Multilingual Retrieval Dataset Covering 18 Diverse Languages",
    author = "Zhang, Xinyu  and
      Thakur, Nandan  and
      Ogundepo, Odunayo  and
      Kamalloo, Ehsan  and
      Alfonso-Hermelo, David  and
      Li, Xiaoguang  and
      Liu, Qun  and
      Rezagholizadeh, Mehdi  and
      Lin, Jimmy",
    journal = "Transactions of the Association for Computational Linguistics",
    volume = "11",
    year = "2023",
    address = "Cambridge, MA",
    publisher = "MIT Press",
    url = "https://aclanthology.org/2023.tacl-1.63/",
    doi = "10.1162/tacl_a_00595",
    pages = "1114--1131",
}

@misc{huang2026tapotranslationaugmentedpolicy,
      title={TAPO: Translation Augmented Policy Optimization for Multilingual Mathematical Reasoning}, 
      author={Xu Huang and Zhejian Lai and Zixian Huang and Jiajun Chen and Shujian Huang},
      year={2026},
      eprint={2603.25419},
      archivePrefix={arXiv},
      primaryClass={cs.CL},
      url={https://arxiv.org/abs/2603.25419}, 
}

@dataset{aime_1983_2024,
  author = {Hemish Veeraboina},
  title = {AIME Problem Set 1983-2024},
  year = {2023},
  publisher = {Kaggle},
  url = {https://www.kaggle.com/datasets/hemishveeraboina/aime-problem-set-1983-2024}
}

@misc{balunovic_srimatharena_2025,
  title = {MathArena: Evaluating LLMs on Uncontaminated Math Competitions},
  author = {Mislav Balunović and Jasper Dekoninck and Ivo Petrov and Nikola Jovanović and Martin Vechev},
  copyright = {MIT},
  url = {https://matharena.ai/},
  publisher = {SRI Lab, ETH Zurich},
  month = feb,
  year = {2025},
}

@inproceedings{rei-etal-2020-comet,
    title = "{COMET}: A Neural Framework for {MT} Evaluation",
    author = "Rei, Ricardo  and
      Stewart, Craig  and
      Farinha, Ana C  and
      Lavie, Alon",
    editor = "Webber, Bonnie  and
      Cohn, Trevor  and
      He, Yulan  and
      Liu, Yang",
    booktitle = "Proceedings of the 2020 Conference on Empirical Methods in Natural Language Processing (EMNLP)",
    month = nov,
    year = "2020",
    address = "Online",
    publisher = "Association for Computational Linguistics",
    url = "https://aclanthology.org/2020.emnlp-main.213/",
    doi = "10.18653/v1/2020.emnlp-main.213",
    pages = "2685--2702",
}

@inproceedings{kocmi-federmann-2023-large,
    title = "Large Language Models Are State-of-the-Art Evaluators of Translation Quality",
    author = "Kocmi, Tom  and
      Federmann, Christian",
    editor = "Nurminen, Mary  and
      Brenner, Judith  and
      Koponen, Maarit  and
      Latomaa, Sirkku  and
      Mikhailov, Mikhail  and
      Schierl, Frederike  and
      Ranasinghe, Tharindu  and
      Vanmassenhove, Eva  and
      Vidal, Sergi Alvarez  and
      Aranberri, Nora  and
      Nunziatini, Mara  and
      Escart{\'i}n, Carla Parra  and
      Forcada, Mikel  and
      Popovic, Maja  and
      Scarton, Carolina  and
      Moniz, Helena",
    booktitle = "Proceedings of the 24th Annual Conference of the European Association for Machine Translation",
    month = jun,
    year = "2023",
    address = "Tampere, Finland",
    publisher = "European Association for Machine Translation",
    url = "https://aclanthology.org/2023.eamt-1.19/",
    pages = "193--203",
}

@article{lanham2023measuring,
  title={Measuring faithfulness in chain-of-thought reasoning},
  author={Lanham, Tamera and Chen, Anna and Radhakrishnan, Ansh and Steiner, Benoit and Denison, Carson and Hernandez, Danny and Li, Dustin and Durmus, Esin and Hubinger, Evan and Kernion, Jackson and others},
  journal={arXiv preprint arXiv:2307.13702},
  year={2023}
}

@inproceedings{paul2024making,
  title={Making reasoning matter: Measuring and improving faithfulness of chain-of-thought reasoning},
  author={Paul, Debjit and West, Robert and Bosselut, Antoine and Faltings, Boi},
  booktitle={Findings of the Association for Computational Linguistics: EMNLP 2024},
  pages={15012--15032},
  year={2024}
}

@inproceedings{jacovi2024chain,
  title={A chain-of-thought is as strong as its weakest link: A benchmark for verifiers of reasoning chains},
  author={Jacovi, Alon and Bitton, Yonatan and Bohnet, Bernd and Herzig, Jonathan and Honovich, Or and Tseng, Michael and Collins, Michael and Aharoni, Roee and Geva, Mor},
  booktitle={Proceedings of the 62nd Annual Meeting of the Association for Computational Linguistics (Volume 1: Long Papers)},
  pages={4615--4634},
  year={2024}
}

@article{golovneva2022roscoe,
  title={Roscoe: A suite of metrics for scoring step-by-step reasoning},
  author={Golovneva, Olga and Chen, Moya and Poff, Spencer and Corredor, Martin and Zettlemoyer, Luke and Fazel-Zarandi, Maryam and Celikyilmaz, Asli},
  journal={arXiv preprint arXiv:2212.07919},
  year={2022}
}

@article{williams2012margins,
  title={Using the margins command to estimate and interpret adjusted predictions and marginal effects},
  author={Williams, Richard},
  journal={The Stata Journal: Promoting Communications on Statistics and Stata},
  volume={12},
  number={2},
  pages={308--331},
  year={2012}
}

@book{long2014regression,
  title        = {Regression Models for Categorical Dependent Variables Using Stata},
  author       = {Long, J. Scott and Freese, Jeremy},
  edition      = {3rd},
  year         = {2014},
  publisher    = {Stata Press},
  isbn         = {978-1-59718-111-2},
}

@article{Kim2019Multicollinearity,
  author    = {Kim, Jong Hae},
  title     = {Multicollinearity and misleading statistical results},
  journal   = {Korean Journal of Anesthesiology},
  volume    = {72},
  number    = {6},
  pages     = {558--569},
  year      = {2019},
  doi       = {10.4097/kja.19087},
  url       = {https://doi.org/10.4097/kja.19087}
}

@article{manakul2023selfcheckgpt,
  title={Selfcheckgpt: Zero-resource black-box hallucination detection for generative large language models},
  author={Manakul, Potsawee and Liusie, Adian and Gales, Mark JF},
  journal={arXiv preprint arXiv:2303.08896},
  year={2023}
}

@inproceedings{you-etal-2025-probabilistic,
    title = "Probabilistic Soundness Guarantees in {LLM} Reasoning Chains",
    author = "You, Weiqiu  and
      Xue, Anton  and
      Havaldar, Shreya  and
      Rao, Delip  and
      Jin, Helen  and
      Callison-Burch, Chris  and
      Wong, Eric",
    editor = "Christodoulopoulos, Christos  and
      Chakraborty, Tanmoy  and
      Rose, Carolyn  and
      Peng, Violet",
    booktitle = "Proceedings of the 2025 Conference on Empirical Methods in Natural Language Processing",
    month = nov,
    year = "2025",
    address = "Suzhou, China",
    publisher = "Association for Computational Linguistics",
    url = "https://aclanthology.org/2025.emnlp-main.382/",
    doi = "10.18653/v1/2025.emnlp-main.382",
    pages = "7506--7525",
    ISBN = "979-8-89176-332-6",
}

@inproceedings{prasad-etal-2023-receval,
    title = "{R}e{CE}val: Evaluating Reasoning Chains via Correctness and Informativeness",
    author = "Prasad, Archiki  and
      Saha, Swarnadeep  and
      Zhou, Xiang  and
      Bansal, Mohit",
    editor = "Bouamor, Houda  and
      Pino, Juan  and
      Bali, Kalika",
    booktitle = "Proceedings of the 2023 Conference on Empirical Methods in Natural Language Processing",
    month = dec,
    year = "2023",
    address = "Singapore",
    publisher = "Association for Computational Linguistics",
    url = "https://aclanthology.org/2023.emnlp-main.622/",
    doi = "10.18653/v1/2023.emnlp-main.622",
    pages = "10066--10086",
}

@inproceedings{
bussmann2024batchtopk,
title={BatchTopK Sparse Autoencoders},
author={Bart Bussmann and Patrick Leask and Neel Nanda},
booktitle={NeurIPS 2024 Workshop on Scientific Methods for Understanding Deep Learning},
year={2024},
url={https://openreview.net/forum?id=d4dpOCqybL}
}

@misc{kang2026multilingualreasoninggapsemerge,
      title={Why Do Multilingual Reasoning Gaps Emerge in Reasoning Language Models?}, 
      author={Deokhyung Kang and Seonjeong Hwang and Daehui Kim and Hyounghun Kim and Gary Geunbae Lee},
      year={2026},
      eprint={2510.27269},
      archivePrefix={arXiv},
      primaryClass={cs.CL},
      url={https://arxiv.org/abs/2510.27269}, 
}

@misc{movva2025whatshumanfeedbacklearning,
      title={What's In My Human Feedback? Learning Interpretable Descriptions of Preference Data}, 
      author={Rajiv Movva and Smitha Milli and Sewon Min and Emma Pierson},
      year={2025},
      eprint={2510.26202},
      archivePrefix={arXiv},
      primaryClass={cs.CL},
      url={https://arxiv.org/abs/2510.26202}, 
}

@inproceedings{rajaee-etal-2026-best,
    title = "Best-of-{L}: Cross-Lingual Reward Modeling for Mathematical Reasoning",
    author = "Rajaee, Sara  and
      Choenni, Rochelle  and
      Shutova, Ekaterina  and
      Monz, Christof",
    editor = "Demberg, Vera  and
      Inui, Kentaro  and
      Marquez, Llu{\'i}s",
    booktitle = "Findings of the {A}ssociation for {C}omputational {L}inguistics: {EACL} 2026",
    month = mar,
    year = "2026",
    address = "Rabat, Morocco",
    publisher = "Association for Computational Linguistics",
    url = "https://aclanthology.org/2026.findings-eacl.99/",
    doi = "10.18653/v1/2026.findings-eacl.99",
    pages = "1930--1939",
    ISBN = "979-8-89176-386-9",
}

@article{cunningham2023sparse,
  title={Sparse autoencoders find highly interpretable features in language models},
  author={Cunningham, Hoagy and Ewart, Aidan and Riggs, Logan and Huben, Robert and Sharkey, Lee},
  journal={arXiv preprint arXiv:2309.08600},
  year={2023}
}

@inproceedings{
movva2025sparse,
title={Sparse Autoencoders for Hypothesis Generation},
author={Rajiv Movva and Kenny Peng and Nikhil Garg and Jon Kleinberg and Emma Pierson},
booktitle={Forty-second International Conference on Machine Learning},
year={2025},
url={https://openreview.net/forum?id=4R0pugRyN5}
}

@misc{galichin2025icoveredbaseshere,
      title={I Have Covered All the Bases Here: Interpreting Reasoning Features in Large Language Models via Sparse Autoencoders}, 
      author={Andrey Galichin and Alexey Dontsov and Polina Druzhinina and Anton Razzhigaev and Oleg Y. Rogov and Elena Tutubalina and Ivan Oseledets},
      year={2025},
      eprint={2503.18878},
      archivePrefix={arXiv},
      primaryClass={cs.CL},
      url={https://arxiv.org/abs/2503.18878}, 
}

@misc{ma2026falsifyingsparseautoencoderreasoning,
      title={Falsifying Sparse Autoencoder Reasoning Features in Language Models}, 
      author={George Ma and Zhongyuan Liang and Irene Y. Chen and Somayeh Sojoudi},
      year={2026},
      eprint={2601.05679},
      archivePrefix={arXiv},
      primaryClass={cs.LG},
      url={https://arxiv.org/abs/2601.05679}, 
}

@misc{liu2026selfimprovingmultilinguallongreasoning,
      title={Self-Improving Multilingual Long Reasoning via Translation-Reasoning Integrated Training}, 
      author={Junxiao Liu and Zhijun Wang and Yixiao Li and Zhejian Lai and Liqian Huang and Xin Huang and Xue Han and Junlan Feng and Shujian Huang},
      year={2026},
      eprint={2602.05940},
      archivePrefix={arXiv},
      primaryClass={cs.CL},
      url={https://arxiv.org/abs/2602.05940}, 
}

@inproceedings{
wang2025polymath,
title={PolyMath: Evaluating Mathematical Reasoning in Multilingual Contexts},
author={Yiming Wang and Pei Zhang and Jialong Tang and Hao-Ran Wei and Baosong Yang and Rui Wang and Chenshu Sun and Feitong Sun and Jiran Zhang and Junxuan Wu and Qiqian Cang and Yichang Zhang and Fei Huang and Junyang Lin and Fei Huang and Jingren Zhou},
booktitle={The Thirty-ninth Annual Conference on Neural Information Processing Systems Datasets and Benchmarks Track},
year={2025},
url={https://openreview.net/forum?id=B1vCImy6yI}
}

@inproceedings{luo-etal-2025-mmath,
    title = "{MMATH}: A Multilingual Benchmark for Mathematical Reasoning",
    author = "Luo, Wenyang  and
      Zhao, Wayne Xin  and
      Sha, Jing  and
      Wang, Shijin  and
      Wen, Ji-Rong",
    editor = "Christodoulopoulos, Christos  and
      Chakraborty, Tanmoy  and
      Rose, Carolyn  and
      Peng, Violet",
    booktitle = "Findings of the Association for Computational Linguistics: EMNLP 2025",
    month = nov,
    year = "2025",
    address = "Suzhou, China",
    publisher = "Association for Computational Linguistics",
    url = "https://aclanthology.org/2025.findings-emnlp.598/",
    doi = "10.18653/v1/2025.findings-emnlp.598",
    pages = "11187--11202",
    ISBN = "979-8-89176-335-7"
}

@misc{gao2025thinkingmultilinguallyempowerllm,
      title={Could Thinking Multilingually Empower LLM Reasoning?}, 
      author={Changjiang Gao and Xu Huang and Wenhao Zhu and Shujian Huang and Lei Li and Fei Yuan},
      year={2025},
      eprint={2504.11833},
      archivePrefix={arXiv},
      primaryClass={cs.CL},
      url={https://arxiv.org/abs/2504.11833}, 
}

@inproceedings{lai-nissim-2024-mcot,
    title = "m{C}o{T}: Multilingual Instruction Tuning for Reasoning Consistency in Language Models",
    author = "Lai, Huiyuan  and
      Nissim, Malvina",
    editor = "Ku, Lun-Wei  and
      Martins, Andre  and
      Srikumar, Vivek",
    booktitle = "Proceedings of the 62nd Annual Meeting of the Association for Computational Linguistics (Volume 1: Long Papers)",
    month = aug,
    year = "2024",
    address = "Bangkok, Thailand",
    publisher = "Association for Computational Linguistics",
    url = "https://aclanthology.org/2024.acl-long.649/",
    doi = "10.18653/v1/2024.acl-long.649",
    pages = "12012--12026",
}

@inproceedings{zhu-etal-2024-question,
    title = "Question Translation Training for Better Multilingual Reasoning",
    author = "Zhu, Wenhao  and
      Huang, Shujian  and
      Yuan, Fei  and
      She, Shuaijie  and
      Chen, Jiajun  and
      Birch, Alexandra",
    editor = "Ku, Lun-Wei  and
      Martins, Andre  and
      Srikumar, Vivek",
    booktitle = "Findings of the Association for Computational Linguistics: ACL 2024",
    month = aug,
    year = "2024",
    address = "Bangkok, Thailand",
    publisher = "Association for Computational Linguistics",
    url = "https://aclanthology.org/2024.findings-acl.498/",
    doi = "10.18653/v1/2024.findings-acl.498",
    pages = "8411--8423",
}

@inproceedings{she-etal-2024-mapo,
    title = "{MAPO}: Advancing Multilingual Reasoning through Multilingual-Alignment-as-Preference Optimization",
    author = "She, Shuaijie  and
      Zou, Wei  and
      Huang, Shujian  and
      Zhu, Wenhao  and
      Liu, Xiang  and
      Geng, Xiang  and
      Chen, Jiajun",
    editor = "Ku, Lun-Wei  and
      Martins, Andre  and
      Srikumar, Vivek",
    booktitle = "Proceedings of the 62nd Annual Meeting of the Association for Computational Linguistics (Volume 1: Long Papers)",
    month = aug,
    year = "2024",
    address = "Bangkok, Thailand",
    publisher = "Association for Computational Linguistics",
    url = "https://aclanthology.org/2024.acl-long.539/",
    doi = "10.18653/v1/2024.acl-long.539",
    pages = "10015--10027",
}

@inproceedings{huang-etal-2025-benchmax,
    title = "{B}ench{MAX}: A Comprehensive Multilingual Evaluation Suite for Large Language Models",
    author = "Huang, Xu  and
      Zhu, Wenhao  and
      Hu, Hanxu  and
      He, Conghui  and
      Li, Lei  and
      Huang, Shujian  and
      Yuan, Fei",
    editor = "Christodoulopoulos, Christos  and
      Chakraborty, Tanmoy  and
      Rose, Carolyn  and
      Peng, Violet",
    booktitle = "Findings of the Association for Computational Linguistics: EMNLP 2025",
    month = nov,
    year = "2025",
    address = "Suzhou, China",
    publisher = "Association for Computational Linguistics",
    url = "https://aclanthology.org/2025.findings-emnlp.909/",
    doi = "10.18653/v1/2025.findings-emnlp.909",
    pages = "16751--16774",
    ISBN = "979-8-89176-335-7",
}

@inproceedings{ko-etal-2025-understand,
    title = "Understand, Solve and Translate: Bridging the Multilingual Mathematical Reasoning Gap",
    author = "Ko, Hyunwoo  and
      Son, Guijin  and
      Choi, Dasol",
    editor = "Adelani, David Ifeoluwa  and
      Arnett, Catherine  and
      Ataman, Duygu  and
      Chang, Tyler A.  and
      Gonen, Hila  and
      Raja, Rahul  and
      Schmidt, Fabian  and
      Stap, David  and
      Wang, Jiayi",
    booktitle = "Proceedings of the 5th Workshop on Multilingual Representation Learning (MRL 2025)",
    month = nov,
    year = "2025",
    address = "Suzhuo, China",
    publisher = "Association for Computational Linguistics",
    url = "https://aclanthology.org/2025.mrl-main.6/",
    doi = "10.18653/v1/2025.mrl-main.6",
    pages = "78--95",
    ISBN = "979-8-89176-345-6",
}

@inproceedings{kwon2023efficient,
  title={Efficient Memory Management for Large Language Model Serving with PagedAttention},
  author={Woosuk Kwon and Zhuohan Li and Siyuan Zhuang and Ying Sheng and Lianmin Zheng and Cody Hao Yu and Joseph E. Gonzalez and Hao Zhang and Ion Stoica},
  booktitle={Proceedings of the ACM SIGOPS 29th Symposium on Operating Systems Principles},
  year={2023}
}

@article{smith1981identification,
  title={Identification of common molecular subsequences},
  author={Smith, Temple F and Waterman, Michael S and others},
  journal={Journal of molecular biology},
  volume={147},
  number={1},
  pages={195--197},
  year={1981},
  publisher={Elsevier Science}
}

@inproceedings{feng-etal-2022-language,
    title = "Language-agnostic {BERT} Sentence Embedding",
    author = "Feng, Fangxiaoyu  and
      Yang, Yinfei  and
      Cer, Daniel  and
      Arivazhagan, Naveen  and
      Wang, Wei",
    editor = "Muresan, Smaranda  and
      Nakov, Preslav  and
      Villavicencio, Aline",
    booktitle = "Proceedings of the 60th Annual Meeting of the Association for Computational Linguistics (Volume 1: Long Papers)",
    month = may,
    year = "2022",
    address = "Dublin, Ireland",
    publisher = "Association for Computational Linguistics",
    url = "https://aclanthology.org/2022.acl-long.62/",
    doi = "10.18653/v1/2022.acl-long.62",
    pages = "878--891",
}

@inproceedings{
xu2025mind,
title={Mind the Gap: Bridging Thought Leap for Improved Chain-of-Thought Tuning},
author={Haolei Xu and Yuchen Yan and Yongliang Shen and Wenqi Zhang and Guiyang Hou and Shengpei Jiang and Kaitao Song and Weiming Lu and Jun Xiao and Yueting Zhuang},
booktitle={The Thirty-ninth Annual Conference on Neural Information Processing Systems},
year={2025},
url={https://openreview.net/forum?id=2ogTw5ue7v}
}

@inproceedings{ghosh-etal-2025-survey,
    title = "A Survey of Multilingual Reasoning in Language Models",
    author = "Ghosh, Akash  and
      Datta, Debayan  and
      Saha, Sriparna  and
      Agarwal, Chirag",
    editor = "Christodoulopoulos, Christos  and
      Chakraborty, Tanmoy  and
      Rose, Carolyn  and
      Peng, Violet",
    booktitle = "Findings of the Association for Computational Linguistics: EMNLP 2025",
    month = nov,
    year = "2025",
    address = "Suzhou, China",
    publisher = "Association for Computational Linguistics",
    url = "https://aclanthology.org/2025.findings-emnlp.474/",
    doi = "10.18653/v1/2025.findings-emnlp.474",
    pages = "8920--8936",
    ISBN = "979-8-89176-335-7",
}

@misc{openai2024gpt4ocard,
      title={GPT-4o System Card}, 
      author={OpenAI and : and Aaron Hurst and Adam Lerer and Adam P. Goucher and Adam Perelman and Aditya Ramesh and Aidan Clark and AJ Ostrow and Akila Welihinda and Alan Hayes and Alec Radford and Aleksander Mądry and Alex Baker-Whitcomb and Alex Beutel and Alex Borzunov and Alex Carney and Alex Chow and Alex Kirillov and Alex Nichol and Alex Paino and Alex Renzin and Alex Tachard Passos and Alexander Kirillov and Alexi Christakis and Alexis Conneau and Ali Kamali and Allan Jabri and Allison Moyer and Allison Tam and Amadou Crookes and Amin Tootoochian and Amin Tootoonchian and Ananya Kumar and Andrea Vallone and Andrej Karpathy and Andrew Braunstein and Andrew Cann and Andrew Codispoti and Andrew Galu and Andrew Kondrich and Andrew Tulloch and Andrey Mishchenko and Angela Baek and Angela Jiang and Antoine Pelisse and Antonia Woodford and Anuj Gosalia and Arka Dhar and Ashley Pantuliano and Avi Nayak and Avital Oliver and Barret Zoph and Behrooz Ghorbani and Ben Leimberger and Ben Rossen and Ben Sokolowsky and Ben Wang and Benjamin Zweig and Beth Hoover and Blake Samic and Bob McGrew and Bobby Spero and Bogo Giertler and Bowen Cheng and Brad Lightcap and Brandon Walkin and Brendan Quinn and Brian Guarraci and Brian Hsu and Bright Kellogg and Brydon Eastman and Camillo Lugaresi and Carroll Wainwright and Cary Bassin and Cary Hudson and Casey Chu and Chad Nelson and Chak Li and Chan Jun Shern and Channing Conger and Charlotte Barette and Chelsea Voss and Chen Ding and Cheng Lu and Chong Zhang and Chris Beaumont and Chris Hallacy and Chris Koch and Christian Gibson and Christina Kim and Christine Choi and Christine McLeavey and Christopher Hesse and Claudia Fischer and Clemens Winter and Coley Czarnecki and Colin Jarvis and Colin Wei and Constantin Koumouzelis and Dane Sherburn and Daniel Kappler and Daniel Levin and Daniel Levy and David Carr and David Farhi and David Mely and David Robinson and David Sasaki and Denny Jin and Dev Valladares and Dimitris Tsipras and Doug Li and Duc Phong Nguyen and Duncan Findlay and Edede Oiwoh and Edmund Wong and Ehsan Asdar and Elizabeth Proehl and Elizabeth Yang and Eric Antonow and Eric Kramer and Eric Peterson and Eric Sigler and Eric Wallace and Eugene Brevdo and Evan Mays and Farzad Khorasani and Felipe Petroski Such and Filippo Raso and Francis Zhang and Fred von Lohmann and Freddie Sulit and Gabriel Goh and Gene Oden and Geoff Salmon and Giulio Starace and Greg Brockman and Hadi Salman and Haiming Bao and Haitang Hu and Hannah Wong and Haoyu Wang and Heather Schmidt and Heather Whitney and Heewoo Jun and Hendrik Kirchner and Henrique Ponde de Oliveira Pinto and Hongyu Ren and Huiwen Chang and Hyung Won Chung and Ian Kivlichan and Ian O'Connell and Ian O'Connell and Ian Osband and Ian Silber and Ian Sohl and Ibrahim Okuyucu and Ikai Lan and Ilya Kostrikov and Ilya Sutskever and Ingmar Kanitscheider and Ishaan Gulrajani and Jacob Coxon and Jacob Menick and Jakub Pachocki and James Aung and James Betker and James Crooks and James Lennon and Jamie Kiros and Jan Leike and Jane Park and Jason Kwon and Jason Phang and Jason Teplitz and Jason Wei and Jason Wolfe and Jay Chen and Jeff Harris and Jenia Varavva and Jessica Gan Lee and Jessica Shieh and Ji Lin and Jiahui Yu and Jiayi Weng and Jie Tang and Jieqi Yu and Joanne Jang and Joaquin Quinonero Candela and Joe Beutler and Joe Landers and Joel Parish and Johannes Heidecke and John Schulman and Jonathan Lachman and Jonathan McKay and Jonathan Uesato and Jonathan Ward and Jong Wook Kim and Joost Huizinga and Jordan Sitkin and Jos Kraaijeveld and Josh Gross and Josh Kaplan and Josh Snyder and Joshua Achiam and Joy Jiao and Joyce Lee and Juntang Zhuang and Justyn Harriman and Kai Fricke and Kai Hayashi and Karan Singhal and Katy Shi and Kavin Karthik and Kayla Wood and Kendra Rimbach and Kenny Hsu and Kenny Nguyen and Keren Gu-Lemberg and Kevin Button and Kevin Liu and Kiel Howe and Krithika Muthukumar and Kyle Luther and Lama Ahmad and Larry Kai and Lauren Itow and Lauren Workman and Leher Pathak and Leo Chen and Li Jing and Lia Guy and Liam Fedus and Liang Zhou and Lien Mamitsuka and Lilian Weng and Lindsay McCallum and Lindsey Held and Long Ouyang and Louis Feuvrier and Lu Zhang and Lukas Kondraciuk and Lukasz Kaiser and Luke Hewitt and Luke Metz and Lyric Doshi and Mada Aflak and Maddie Simens and Madelaine Boyd and Madeleine Thompson and Marat Dukhan and Mark Chen and Mark Gray and Mark Hudnall and Marvin Zhang and Marwan Aljubeh and Mateusz Litwin and Matthew Zeng and Max Johnson and Maya Shetty and Mayank Gupta and Meghan Shah and Mehmet Yatbaz and Meng Jia Yang and Mengchao Zhong and Mia Glaese and Mianna Chen and Michael Janner and Michael Lampe and Michael Petrov and Michael Wu and Michele Wang and Michelle Fradin and Michelle Pokrass and Miguel Castro and Miguel Oom Temudo de Castro and Mikhail Pavlov and Miles Brundage and Miles Wang and Minal Khan and Mira Murati and Mo Bavarian and Molly Lin and Murat Yesildal and Nacho Soto and Natalia Gimelshein and Natalie Cone and Natalie Staudacher and Natalie Summers and Natan LaFontaine and Neil Chowdhury and Nick Ryder and Nick Stathas and Nick Turley and Nik Tezak and Niko Felix and Nithanth Kudige and Nitish Keskar and Noah Deutsch and Noel Bundick and Nora Puckett and Ofir Nachum and Ola Okelola and Oleg Boiko and Oleg Murk and Oliver Jaffe and Olivia Watkins and Olivier Godement and Owen Campbell-Moore and Patrick Chao and Paul McMillan and Pavel Belov and Peng Su and Peter Bak and Peter Bakkum and Peter Deng and Peter Dolan and Peter Hoeschele and Peter Welinder and Phil Tillet and Philip Pronin and Philippe Tillet and Prafulla Dhariwal and Qiming Yuan and Rachel Dias and Rachel Lim and Rahul Arora and Rajan Troll and Randall Lin and Rapha Gontijo Lopes and Raul Puri and Reah Miyara and Reimar Leike and Renaud Gaubert and Reza Zamani and Ricky Wang and Rob Donnelly and Rob Honsby and Rocky Smith and Rohan Sahai and Rohit Ramchandani and Romain Huet and Rory Carmichael and Rowan Zellers and Roy Chen and Ruby Chen and Ruslan Nigmatullin and Ryan Cheu and Saachi Jain and Sam Altman and Sam Schoenholz and Sam Toizer and Samuel Miserendino and Sandhini Agarwal and Sara Culver and Scott Ethersmith and Scott Gray and Sean Grove and Sean Metzger and Shamez Hermani and Shantanu Jain and Shengjia Zhao and Sherwin Wu and Shino Jomoto and Shirong Wu and Shuaiqi and Xia and Sonia Phene and Spencer Papay and Srinivas Narayanan and Steve Coffey and Steve Lee and Stewart Hall and Suchir Balaji and Tal Broda and Tal Stramer and Tao Xu and Tarun Gogineni and Taya Christianson and Ted Sanders and Tejal Patwardhan and Thomas Cunninghman and Thomas Degry and Thomas Dimson and Thomas Raoux and Thomas Shadwell and Tianhao Zheng and Todd Underwood and Todor Markov and Toki Sherbakov and Tom Rubin and Tom Stasi and Tomer Kaftan and Tristan Heywood and Troy Peterson and Tyce Walters and Tyna Eloundou and Valerie Qi and Veit Moeller and Vinnie Monaco and Vishal Kuo and Vlad Fomenko and Wayne Chang and Weiyi Zheng and Wenda Zhou and Wesam Manassra and Will Sheu and Wojciech Zaremba and Yash Patil and Yilei Qian and Yongjik Kim and Youlong Cheng and Yu Zhang and Yuchen He and Yuchen Zhang and Yujia Jin and Yunxing Dai and Yury Malkov},
      year={2024},
      eprint={2410.21276},
      archivePrefix={arXiv},
      primaryClass={cs.CL},
      url={https://arxiv.org/abs/2410.21276}, 
}

@inproceedings{charniak2005coarse,
  title={Coarse-to-fine n-best parsing and maxent discriminative reranking},
  author={Charniak, Eugene and Johnson, Mark},
  booktitle={Proceedings of the 43rd Annual Meeting of the Association for Computational Linguistics (ACL’05)},
  pages={173--180},
  year={2005}
}

@inproceedings{zheng2025processbench,
  title={Processbench: Identifying process errors in mathematical reasoning},
  author={Zheng, Chujie and Zhang, Zhenru and Zhang, Beichen and Lin, Runji and Lu, Keming and Yu, Bowen and Liu, Dayiheng and Zhou, Jingren and Lin, Junyang},
  booktitle={Proceedings of the 63rd Annual Meeting of the Association for Computational Linguistics (Volume 1: Long Papers)},
  pages={1009--1024},
  year={2025}
}

@inproceedings{chen-etal-2024-breaking,
    title = "Breaking Language Barriers in Multilingual Mathematical Reasoning: Insights and Observations",
    author = "Chen, Nuo  and
      Zheng, Zinan  and
      Wu, Ning  and
      Gong, Ming  and
      Zhang, Dongmei  and
      Li, Jia",
    editor = "Al-Onaizan, Yaser  and
      Bansal, Mohit  and
      Chen, Yun-Nung",
    booktitle = "Findings of the Association for Computational Linguistics: EMNLP 2024",
    month = nov,
    year = "2024",
    address = "Miami, Florida, USA",
    publisher = "Association for Computational Linguistics",
    url = "https://aclanthology.org/2024.findings-emnlp.411/",
    doi = "10.18653/v1/2024.findings-emnlp.411",
    pages = "7001--7016",
}

@misc{dobler2026multilingualreasoninggymmultilingual,
      title={Multilingual Reasoning Gym: Multilingual Scaling of Procedural Reasoning Environments}, 
      author={Konstantin Dobler and Simon Lehnerer and Federico Scozzafava and Jonathan Janke and Mohamed Ali},
      year={2026},
      eprint={2603.10793},
      archivePrefix={arXiv},
      primaryClass={cs.CL},
      url={https://arxiv.org/abs/2603.10793}, 
}
\bibliographystyle{colm2026_conference}

\clearpage
\appendix
\section{Measurable Reasoning Feature Details}
\label{appendix:features}

\subsection{\raisebox{-0.2em}{\includegraphics[height=1.1em]{figures/logo/multilingual.png}} \textcolor{multilingual}{\textbf{Multilingual Alignment}}}
\paragraph{Structural Similarity.}
We define structural similarity as how closely the sequence of \textcolor{cognitive}{\textbf{Reasoning Flow}} tags in a non-English trace matches that of its English counterpart. 
For each query, we annotate every step in both the English and non-English reasoning traces with \textcolor{cognitive}{\textbf{Reasoning Flow}} tags using \textsc{GPT-4o}, yielding two tag sequences \(\mathcal{S}_{\mathrm{en}}\) and \(\mathcal{S}_{\mathrm{non\text{-}en}}\). 
We then compute the Smith–Waterman \citep{smith1981identification} local alignment score between these sequences using a match score of \(+2\) and penalties of \(-1\) for both mismatches and gaps. Finally, we normalize this local alignment score by the maximum possible score \(2 \cdot \min(|\mathcal{S}_{\mathrm{en}}|, |\mathcal{S}_{\mathrm{non\text{-}en}}|)\), yielding a ratio in \([0, 1]\), where values near 1 indicate nearly identical reasoning structure and values near 0 indicate no meaningful shared subsequence.

\subsection{\raisebox{-0.2em}{\includegraphics[height=1.1em]{figures/logo/intermediate.png}} \textcolor{intermediate}{\textbf{Reasoning Step}}}

For both \textcolor{intermediate}{\textbf{Reasoning Step}} and \textcolor{cognitive}{\textbf{Reasoning Flow}} features, we use \textsc{GPT-4o} to segment and annotate each step in a reasoning trace with its corresponding dependency relationships and \textcolor{cognitive}{\textbf{Reasoning Flow}} tags, following the prompt in Appendix \ref{appendix:prompts} \citep{bogdan2025thoughtanchorsllmreasoning}.

\paragraph{Validity.}
We quantify step-level validity using an off-the-shelf Natural Language Inference (NLI) model \citep{manakul2023selfcheckgpt}. For each step, we treat all prior steps it depends on as premises and the step itself as the hypothesis, run a pre-trained \textsc{DeBERTa-v3 MNLI} classifier\footnote{\url{https://huggingface.co/potsawee/deberta-v3-large-mnli}}, and count how many dependencies are labeled as entailment, neutral, or contradiction.
We then compute the entailment, neutral, and contradiction rates over its dependencies, and define the validity score as 0 if any dependency is a contradiction, and otherwise as the entailment rate (i.e., the proportion of dependencies that entail the step) \citep{prasad-etal-2023-receval}.

\paragraph{Direct/Indirect Utility.}
We compute both direct and indirect utility from the dependency graph over steps. First, we locate the step tagged as ``Final Answer Emission'' and collect all its ancestors by recursively following the links labeled as ``depends on''. These steps (including the final step itself) are assigned direct utility 1, and all others 0, yielding the direct utility score as the fraction of steps with direct utility \citep{lee-hockenmaier-2025-evaluating}.
Next, any step that is a dependency of a direct utility step is assigned indirect utility 1 (supporting steps), and the indirect utility score is the fraction of such steps in the trace.

\paragraph{$\mathcal{V}$-Information.}
We measure how much a reasoning trace \(t\) increases a model \(\mathcal{V}\)'s confidence in the final answer \(a\). Concretely, for each query $q$, we construct two prompts: one that includes the full reasoning trace inside a \texttt{<think><\textbackslash think>} block before the final answer, and one that omits the trace and presents only the query. For each LRM, using \textsc{vLLM} with sampling temperature as 0.0, we compute the log-probability of the gold answer tokens under both prompts and define
\[
\mathcal{V}I(t \rightarrow a) = \log p_{\mathcal{V}}(a \mid q,t) - \log p_{\mathcal{V}}(a \mid q).
\]
Positive values indicate that providing the trace makes the model assign higher probability to the correct answer, while negative values indicate that the trace reduces its confidence.

\subsection{\raisebox{-0.2em}{\includegraphics[height=1.1em]{figures/logo/cognitive.png}} \textcolor{cognitive}{\textbf{Reasoning Flow}}}

Example for each tag is shown in~\autoref{tab:tag_examples}.

\begin{table}
\centering
\resizebox{\linewidth}{!}{%
    \begin{tabular}{llllllllll}
    \toprule
    \textbf{Tag} & \textbf{Examples} \\
    \toprule

    \textbf{Self-Checking} & \textit{Let me verify: $\pi r^2=\pi \times 5^2=25\pi$. Correct.} \\
    \textbf{Active Computation} & \textit{Substituting $r=5:A=\pi \times 5^2=25\pi$.} \\
    \textbf{Problem Setup} & \textit{I need to find the area of a circle with radium $5$cm.} \\
    \textbf{Plan Generation} & \textit{I'll solve this by applying the area formula.} \\
    \textbf{Final Answer Emission} & \textit{Therefore, the answer is ...} \\
    \textbf{Fact Retrieval} & \textit{The formula for the area of a circle is $A=\pi r^2$.} \\
    \textbf{Result Consolidation} & \textit{So the area is $25\pi$ square cm which is approximately ...} \\
    \textbf{Uncertainty Management} & \textit{Wait, I think I made a mistake earlier when substituting $r=5$. Let me reconsider ...} \\

    \bottomrule
    \end{tabular}
}
\caption{\textbf{Examples of each cognitive-behavioral tag from \citet{bogdan2025thoughtanchorsllmreasoning}.}}  
\label{tab:tag_examples}
\end{table}

\section{Experiment Setup Details}

For all models, we adopt the sampling configuration recommended in their respective technical reports.
Using the \textsc{vLLM} setup \citep{kwon2023efficient}, we set the maximum generation length to 32,768 tokens, temperature of 0.6, top-p value of 0.95, top-k value of 20, min-p value of 0.0, presence penalty of 0.0, and generate 3 responses per query to compute accuracy. We use \texttt{math-verify}\footnote{\url{https://github.com/huggingface/Math-Verify}} for extracting the \textbackslash \textbackslash \texttt{boxed\{\}} answer from the final response. 
We show prompt templates used for each language in~\autoref{appendix:prompts}.

\begin{table}[H]
\centering
\resizebox{\linewidth}{!}{%
    \begin{tabular}{llllllllll}
    \toprule
    \textbf{Dataset} & \textbf{\# Queries} & \textbf{Translation} \\
    \toprule

    \textbf{MGSM-Rev2} & 250 & Human-translated by professional translators \citep{shi2022languagemodelsmultilingualchainofthought} \\
    \textbf{AIME} & 60 (30+30) & Machine-translated with \textsc{GPT-4o-mini} \citep{qi-etal-2025-models} \\

    \bottomrule
    \end{tabular}
}
\caption{\textbf{Detailed statistics of evaluation datasets.} We report statistics for MGSM-Rev2 and AIME 2024--25 \citep{aime_1983_2024, balunovic_srimatharena_2025}.}  
\label{tab:data_statistics}
\end{table}
\begin{table}[H]
\centering
\resizebox{\linewidth}{!}{%
    \begin{tabular}{llllllllll}
    \toprule
    \textbf{Language} & \textbf{COMET (AIME)} & \textbf{COMET (MGSM)} & \textbf{COMET (Rev2)} & \textbf{GEMBA (MGSM)} & \textbf{GEMBA (Rev2)} & \textbf{\# Updated (\%)} \\
    \toprule

    \textbf{Bengali} & 0.814 & 0.872 & 0.874 & 99.2 & 99.4 & 46 (18.4\%) \\
    \textbf{English} & - & - & - & - & - & 22 (8.80\%) \\
    \textbf{German} & 0.815 & 0.846 & 0.847 & 98.6 & 99.2 & 35 (14.0\%) \\
    \textbf{Spanish} & 0.820 & 0.865 & 0.865 & 99.3 & 99.5 & 38 (15.2\%) \\
    \textbf{French} & 0.834 & 0.864 & 0.864 & 98.8 & 99.4 & 45 (18.0\%) \\
    \textbf{Russian} & 0.822 & 0.855 & 0.855 & 99.2 & 99.5 & 32 (12.8\%) \\
    \textbf{Swahili} & 0.790 & 0.823 & 0.825 & 97.2 & 97.8 & 43 (17.2\%)\\
    \textbf{Telugu} & 0.788 & 0.848 & 0.850 & 99.1 & 99.4 & 52 (20.8\%) \\
    \textbf{Thai} & 0.795 & 0.832 & 0.833 & 98.8 & 99.1 & 40 (16.0\%) \\
    \textbf{Chinese} & 0.809 & 0.848 & 0.848 & 98.7 & 99.3 & 43 (17.2\%) \\

    \bottomrule
    \end{tabular}
}
\caption{\textbf{COMET-QE, GEMBA-DA, and \# of updated queries per language.} COMET-QE scores \citep{rei-etal-2020-comet} are higher for MGSM-Rev2 than for machine-translated AIME, and both COMET-QE and GEMBA-DA \citep{kocmi-federmann-2023-large} (with \textsc{GPT-4o}) increase from MGSM to MGSM-Rev2.}
\label{tab:mgsmrev2_stats}
\end{table}
\definecolor{midpink}{RGB}{225, 149, 171}
\definecolor{darkpink}{RGB}{186, 67, 101}
\definecolor{darkerpink}{RGB}{99, 36, 54}

\definecolor{midgreen}{RGB}{96, 171, 117}
\definecolor{darkgreen}{RGB}{32, 84, 47}
\definecolor{darkergreen}{RGB}{0, 156, 43}

\definecolor{midblue}{RGB}{66, 135, 245}
\definecolor{darkblue}{RGB}{35, 77, 145}
\definecolor{darkerblue}{RGB}{96, 117, 150}

\begin{table}[H]
\centering
\resizebox{\linewidth}{!}{%
    \begin{tabular}{llllllllll}
    \toprule
    \textbf{Language Family} & \textbf{Language} & \textbf{Script} & \textbf{Synthesis} & \textbf{Word Order} & \textbf{Resource Level} & \textbf{\# Speakers} & \textbf{\# Wikipedia Size} \\
    \toprule

    \multirow{6}{*}{Indo-European} & English & Latin & \textcolor{midblue}{analytic} & \textcolor{midpink}{SVO} & \cellcolor{midgreen!40} high & \cellcolor{midgreen!80} 1,130M & \cellcolor{darkgreen!50} 5,758,285 \\
    
    & French & Latin & \textcolor{darkblue}{fusional} & \textcolor{midpink}{SVO} & \cellcolor{midgreen!40} high & \cellcolor{midgreen!30} 398M & \cellcolor{darkgreen!30} 2,325,608 \\
    
    & Spanish & Latin & \textcolor{darkblue}{fusional} & \textcolor{midpink}{SVO} & \cellcolor{midgreen!40} high & \cellcolor{midgreen!60} 592M & \cellcolor{darkgreen!20} 1,669,181 \\

    & German & Latin & \textcolor{darkblue}{fusional} & \textcolor{midpink}{SVO}, \textcolor{darkpink}{SOV} & \cellcolor{orange!30} mid & \cellcolor{orange!30} 178M & \cellcolor{darkgreen!30} 2,651,352 \\
    
    & Russian & Cyrillic & \textcolor{darkblue}{fusional} & \textcolor{midpink}{SVO} & \cellcolor{orange!30} mid & \cellcolor{midgreen!30} 260M & \cellcolor{darkgreen!20} 1,476,045 \\
    
    & Bengali & Bengali & \textcolor{darkblue}{fusional} & \textcolor{darkpink}{SOV} & \cellcolor{red!20} low & \cellcolor{midgreen!30} 337M & \cellcolor{red!15} 63,762 \\
    \midrule
    
    Sino-Tibetan & Chinese & Chinese & \textcolor{midblue}{analytic} & \textcolor{midpink}{SVO} & \cellcolor{midgreen!40} high & \cellcolor{midgreen!80} 1,350M & \cellcolor{darkgreen!20} 1,246,389 \\
    \midrule
    
    
    
    Niger-Congo & Swahili & Latin & \textcolor{darkerblue}{agglutinative} & \textcolor{midpink}{SVO} & \cellcolor{red!20} low & \cellcolor{red!30} 83M & \cellcolor{red!25} 47,793 \\
    \midrule

    Dravidian & Telugu & Telugu & \textcolor{darkerblue}{agglutinative} & \textcolor{darkpink}{SOV} & \cellcolor{red!20} low & \cellcolor{red!30} 96M & \cellcolor{red!15} 66,353 \\
    \midrule

    Kra-Dai & Thai & Thai & \textcolor{midblue}{analytic} & \textcolor{midpink}{SVO} & \cellcolor{red!20} low & \cellcolor{red!30} 72M & \cellcolor{orange!15} 128,179 \\
    
    \bottomrule
    \end{tabular}
}
\caption{\textbf{Characteristics of tested languages.} For each language, we show language family, script, linguistic typologies (synthesis and word order), and resource level measured by the number of speakers and Wikipedia articles \citep{zhang-etal-2023-miracl}.}  
\label{tab:lang_statistics}
\end{table}

\begin{table}[H]
\centering
\resizebox{\linewidth}{!}{%
    \begin{tabular}{llllllllll}
    \toprule
    \textbf{Model} & \textbf{Context Length} & \textbf{Vocab. Size} & \textbf{HuggingFace Model Identifier} \\
    \toprule

    \raisebox{-0.2em}{\includegraphics[height=1.1em]{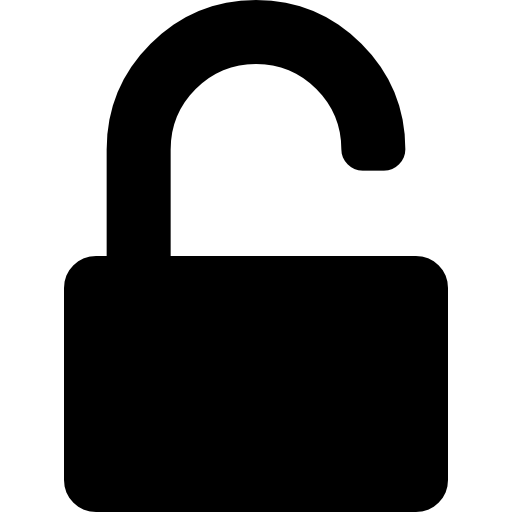}} \textbf{\textsc{Distill-Qwen 1.5b}} & 128K & 152K & \texttt{deepseek-ai/DeepSeek-R1-Distill-Qwen-1.5B} \\
    
    \raisebox{-0.2em}{\includegraphics[height=1.1em]{figures/logo/padlock-unlock.png}} \textbf{\textsc{Distill-Qwen 7b}} & 128K & 152K & \texttt{deepseek-ai/DeepSeek-R1-Distill-Qwen-7B} \\

    \raisebox{-0.2em}{\includegraphics[height=1.1em]{figures/logo/padlock-unlock.png}} \textbf{\textsc{Qwen-3} 4B} & 33K & 152K & \texttt{Qwen/Qwen3-4B} \\

    \raisebox{-0.2em}{\includegraphics[height=1.1em]{figures/logo/padlock-unlock.png}} \textbf{\textsc{Qwen-3} 8B} & 33K & 152K & \texttt{Qwen/Qwen3-8B} \\

    \bottomrule
    \end{tabular}
}
\caption{\textbf{List of evaluated models.} We report the context length, vocabulary size, and HuggingFace model identifiers. We use \textsc{Qwen-3} series models with \colorbox{gray!20}{\texttt{enable\_think=True}} mode.}  
\label{tab:model_statistics}
\end{table}

\section{Prompts}
\label{appendix:prompts}

\begin{figure}[H]
    \centering
    \begin{mdframed}[
        linecolor=lightgrayframe,
        linewidth=0.8pt,
        roundcorner=2pt,
        backgroundcolor=white
    ]
        \includegraphics[width=\linewidth]{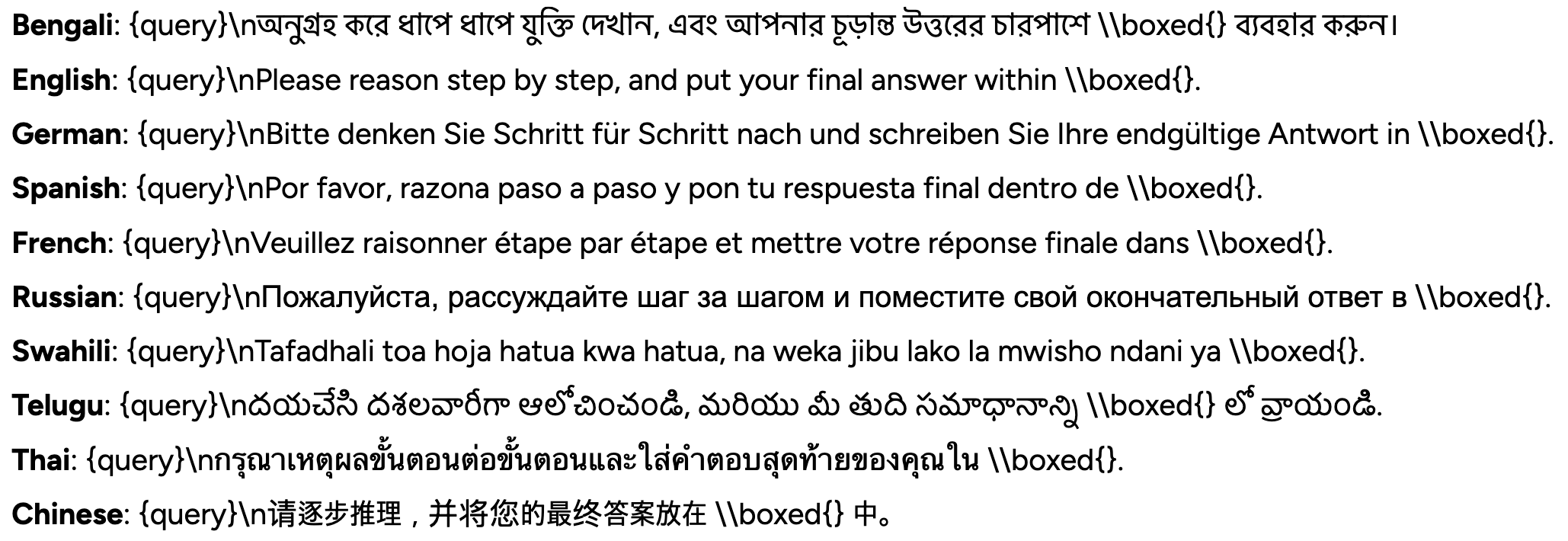}
    \end{mdframed}
    \caption{\textbf{Prompt templates used for sampling generations for each language.}}
    \label{fig:sampling}
\end{figure}

We provide the prompt templates used to sample generations for each language in our main experiments in~\autoref{fig:sampling}. We follow provider-recommended prompting practices to standardize output format \citep{deepseekai2025deepseekr1incentivizingreasoningcapability, yang2025qwen3}.

We also show the prompt template used to annotate \textcolor{cognitive}{\textbf{Reasoning Flow}} features for each reasoning step with \textsc{GPT-4o} in Prompt C.1. We adapt the prompt from \citet{bogdan2025thoughtanchorsllmreasoning}. We use a sampling temperature of 0.0.

\section{Detailed Results}
\subsection{Human Verification of \textsc{GPT-4o} Annotation}
\label{appendix:human}

We conduct a human verification study of \textsc{GPT-4o}-based annotations on a small subset of reasoning traces in languages other than English. To ensure that annotators can comfortably inspect an entire trace in a single sitting, we sample traces from MGSM-Rev2, whose traces are substantially shorter on average than those from AIME (19 vs.\ 275). We focus on French, Russian, and Chinese, for which we could reliably recruit native speaker participants online.

\begin{figure}[t]
    \centering
      \begin{mdframed}[
        linecolor=lightgrayframe,
        linewidth=0.8pt,
        roundcorner=2pt,
        backgroundcolor=white
      ]
        \includegraphics[width=0.8\textwidth]{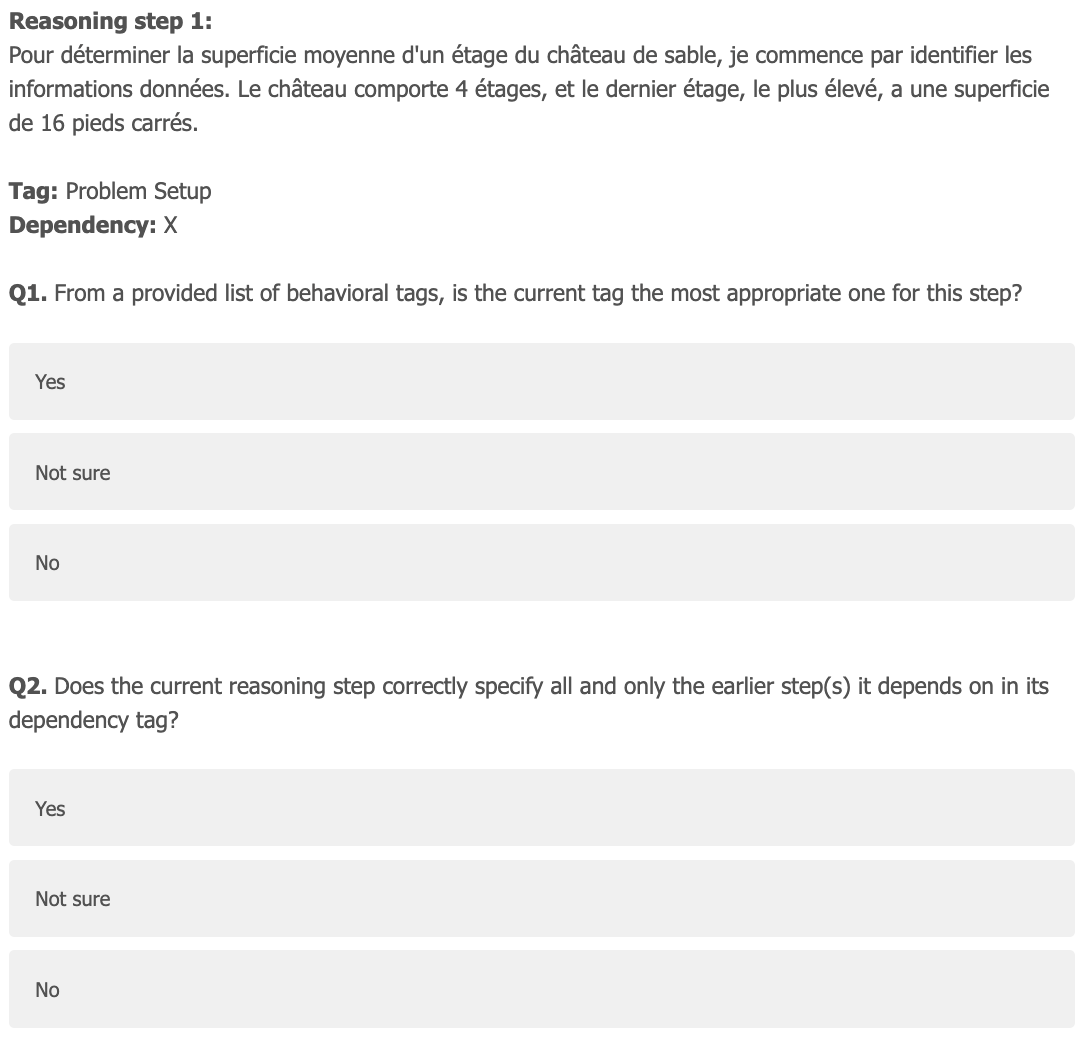}
      \end{mdframed}

    \caption{\textbf{Screenshots of task instructions provided to human annotators.} For each example, we show the question, the full reasoning trace, and an explanation of the available behavioral tags. For every reasoning step, annotators answer two questions (\textbf{Q1} and \textbf{Q2}) about the assigned behavioral tag and the annotated premise relationship.}
    \label{fig:annotation}
\end{figure}

We design the survey in Qualtrics\footnote{\url{https://www.qualtrics.com}} to recruit participants via Prolific\footnote{\url{https://www.prolific.com}} who self-report the target language as their first and primary language and also report fluency in English. We restrict to participants with at least 20 prior submission and a $\geq$ 95\% approval rate for quality control.
Since annotating even a single reasoning trace is time-consuming, we limit the evaluation to 2 traces per language, selecting one trace with more reasoning steps than the language-specific average and one with fewer.
As illustrated in~\autoref{fig:annotation}, each annotator is asked to judge (1) whether the annotated \textcolor{cognitive}{\textbf{Reasoning Flow}} tag is appropriate (Yes/Not sure/No) and (2) whether the dependency path annotation is appropriate (Yes/Not sure/No).
We recruit 3 annotators per language and compensate each with 3 USD (equivalent to 18 USD/hour), with a median completion time of 10 minutes.

As shown in~\autoref{tab:agreement}, we observe high inter-annotator agreement within each language (measured using Fleiss's $\kappa$\footnote{\href{https://www.statsmodels.org/dev/generated/statsmodels.stats.inter_rater.fleiss_kappa.html}{\texttt{https://www.statsmodels.org/statsmodels.stats.inter-rater.fleiss-kappa.html}}}), as well as strong agreement between human judgments and \textsc{GPT-4o} annotations (measured using majority voting).

\begin{table}
\centering
\resizebox{0.6\linewidth}{!}{%
    \begin{tabular}{llllllllll}
    \toprule
    \multirow{2}{*}{\textbf{Language}} & \multicolumn{2}{c}{\textbf{Fleiss's $\kappa$}} & \multicolumn{3}{c}{\textbf{Majority Voting}} \\
    \cmidrule(lr){2-3}
    \cmidrule(lr){4-6}
    & \textbf{Q1} & \textbf{Q2} & \textbf{Q1} & \textbf{Q2} & \textbf{Avg.} \\
    
    \toprule

    \textbf{French} & 0.455 & 0.618 & 0.875 & 0.875 & 0.875 \\
    \textbf{Russian} & 0.671 & 0.705 & 0.938 & 0.875 & 0.907 \\
    \textbf{Chinese} & 0.752 & 0.733 & 1.000 & 0.875 & 0.938 \\

    \bottomrule
    \end{tabular}
}
\caption{\textbf{Inter-annotator agreement of human verification of \textsc{GPT-4o} annotation.} We calculate Fleiss's $\kappa$ for agreement of 3 annotators and use majority voting for agreement with \textsc{GPT-4o} annotation. \textbf{Q1}: Question on \textcolor{cognitive}{\textbf{Reasoning Flow}} tag; \textbf{Q2}: Question on premise relationship.}  
\label{tab:agreement}
\end{table}

\subsection{Multivariate Logistic Regression}
\label{appendix:multivariate}

For our feature analysis (\S\ref{sec:feature}), we fit separate univariate logistic regressions to isolate each feature's marginal relationship to accuracy and avoid multicollinearity between potentially correlated features. For completeness, we also fit a single multivariate logistic regression per model and language using all features jointly.

Let $y \in \{0,1\}$ denote the final answer correctness and let $\mathbf{z}=(z_1,...,z_J)^\top \in \mathbb{R}^J$ denote the vector of feature values. As before, we standardize each continuous feature to zero mean and unit variance within each language:
\begin{equation}
\tilde{z}_j = \frac{z_j-\mu_j}{\sigma_j}.
\end{equation}
For each model and language $\ell$, we fit a multivariate logistic regression over all features jointly:
\begin{equation}
P(y=1\mid\tilde{\mathbf{z}}_{\ell})=\sigma(\alpha_\ell+\sum^J_{j=1}\beta_{j,\ell} \tilde{z}_{j,\ell}),
\end{equation}
where $\sigma(\cdot)$ is the logistic sigmoid, $\alpha_\ell$ is a language-specific intercept, and $\beta_{j,\ell}$ captures the association between feature $j$ and accuracy in language $\ell$ after controlling for all other features. We use $L_2$-regularization to stabilize estimation given the number of features relative to per-language sample sizes.

We summarize the effect of feature $j$ by discrete change in predicted accuracy when varying $\tilde{z}_{j,\ell}$ between $-1$ and $+1$ while holding all other features at their observed values:
\begin{equation}
\Delta\mathrm{Acc}^{\mathrm{multi}}_{j,\ell}=\hat{y}_{j,\ell}\!\mid_{\tilde{z}_{j,\ell}=+1,\,\tilde{\mathbf{z}}_{\neg j,\ell}}-\hat{y}_{j,\ell}\!\mid_{\tilde{z}_{j,\ell}=-1,\,\tilde{\mathbf{z}}_{\neg j,\ell}}
\end{equation}
where $\tilde{\mathbf{z}}_{\neg j,\ell}$ denotes all features except $j$ and
\begin{equation}
\hat{y}_\ell(\tilde{z}_{j,\ell},\tilde{\mathbf{z}}_{\neg j,\ell}) = \sigma(\alpha_\ell+\beta_{j,\ell} \tilde{z}_{j,\ell}+\sum_{k\neq j}\beta_{k,\ell} \tilde{z}_{k,\ell}).
\end{equation}

\begin{figure*}
    \centering
    \includegraphics[width=\linewidth]{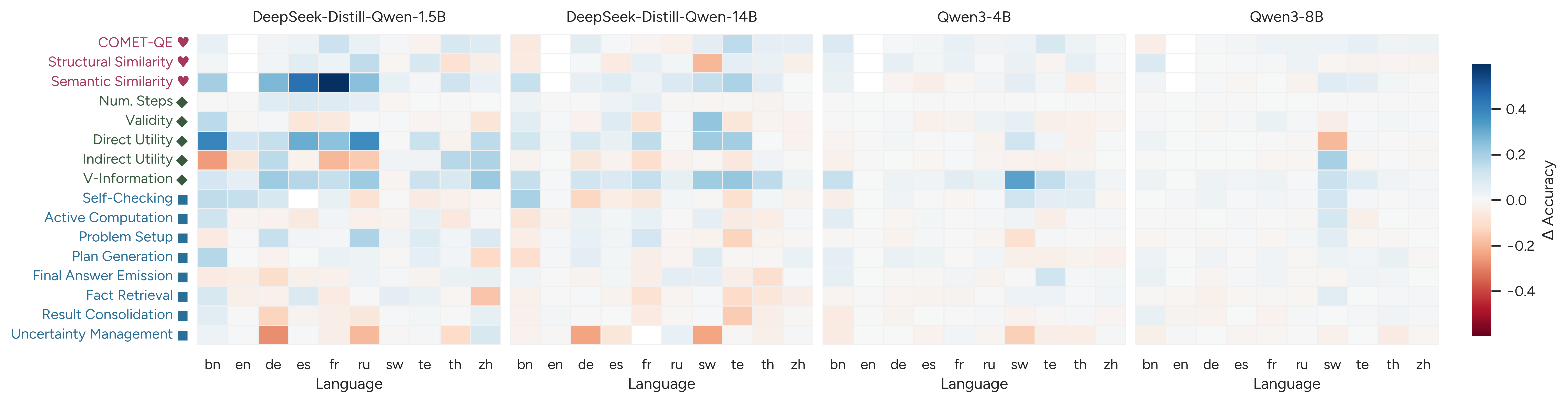}
    \includegraphics[width=\linewidth]{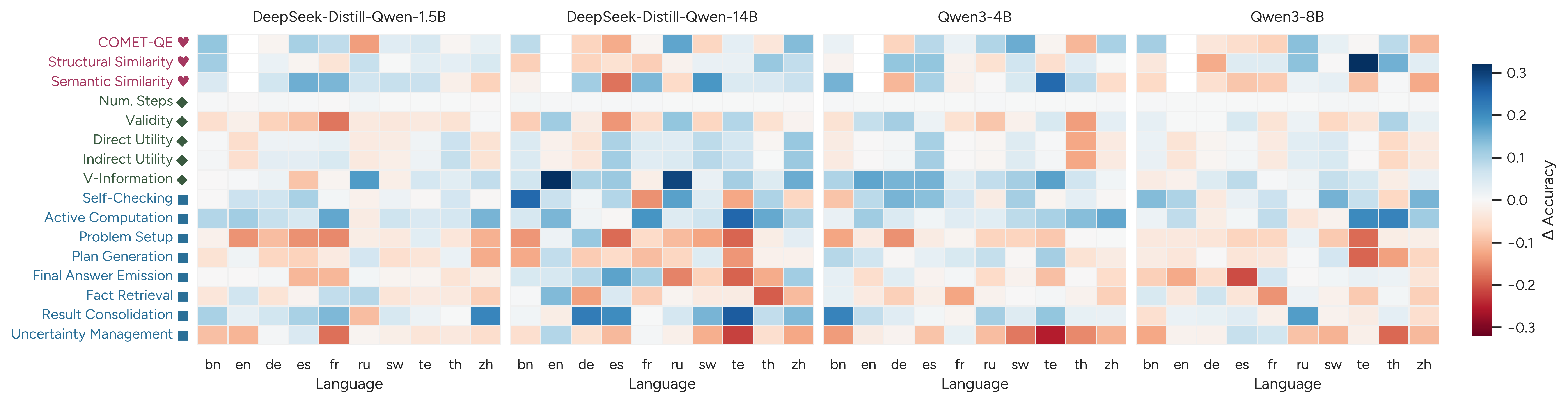}
    \caption{\textbf{Per-language feature analysis results with multivariate logistic regression.} Top: MGSM-Rev2; bottom: AIME. We show similar trends as in the univariate logistic regression results.}
    \label{fig:multivariate}
\end{figure*}

As shown in ~\autoref{fig:multivariate}, comparing to the per-language univariate results in~\autoref{fig:res2} we observe the following trends:
\begin{itemize}[leftmargin=10pt, itemsep=1pt, parsep=-1pt]    
    \item \textbf{Core patterns are preserved.} Features with the strongest positive $\Delta\mathrm{Acc}$ in the univariate regression, such as \textcolor{multilingual}{semantic similarity}, \textcolor{intermediate}{direct utility}, and \textcolor{intermediate}{indirect utility}, remain strongly positive in the multivariate setting, while features with negative $\Delta\mathrm{Acc}$, including \textcolor{cognitive}{problem setup} and \textcolor{cognitive}{uncertainty management}, generally remain negative.

    \item \textbf{Effect sizes are smaller in magnitude.} Per-language effect sizes are typically attenuated under multivariate regression, as coefficients now reflect \textit{partial} effects conditioned on all other features \citep{Wang2017Inconsistency}.

    \item \textbf{Reasoning flow features show more mixed behavior.} Since many \textcolor{cognitive}{\textbf{Reasoning Flow}} features co-occur within the same traces, their effects can be partially redistributed across correlated features in the multivariate setting, leading to trends that differ from the univariate analysis.
\end{itemize}

\subsection{SAE Analysis Ablations}
\label{appendix:ablation}

We perform ablation studies to identify the SAE configuration used in our main analysis (\S\ref{sec:rq2}). 
We summarize the configuration space in the HypotheSAEs implementation in~\autoref{tab:sae_config}; all options are drawn from \citet{movva2025sparse}.\footnote{\url{https://github.com/rmovva/HypotheSAEs}} We also report our qualitative observations and the rationales behind our final configuration choices.

Batch TopK SAE works as follows: given a batch of size $B$ and sparsity target $K$, Batch TopK keeps only the largest $B\cdot K$ activations across the batch and sets all others to zero, and at inference time applies a learned global threshold so that each chunk activates on average $K\ll M$ concepts.

\subsection{Feature Values \& Accuracy}
\label{appendix:raw_results}

We provide raw feature values and associated accuracies for each measurable reasoning feature across the four LRMs, with MGSM-Rev2 detailed in~\autoref{tab:raw_mgsm} AIME in \autoref{tab:raw_aime}.

\subsection{Qualitative Analysis of Traces}
\label{appendix:qualitative}

We show side-by-side qualitative examples of reasoning traces from queries in English versus in languages other than English in~\autoref{tab:qualitative}. We include (i) pairs where both traces reach the correct final answer but exhibit different reasoning behaviors, and (ii) pairs where one trace is correct and the other incorrect despite displaying similar reasoning behaviors.

For the first example (both \cmark), the English trace begins with explicit \textcolor{cognitive}{plan generation} followed by \textcolor{cognitive}{active computation}, whereas the Bengali trace immediately performs \textcolor{cognitive}{active computation} and follows a different logarithmic route to the correct answer. 
Similarly for the second example, the English trace contains substantial \textcolor{cognitive}{uncertainty management} (14 steps), which ultimately leads the model to re-check its interpretation before emitting the final answer, whereas the French trace reaches the same correct answer without any steps labeled as \textcolor{cognitive}{uncertainty management}.
In the third example (English \xmark, non-English \cmark), both traces start with \textcolor{cognitive}{problem setup}, proceed through \textcolor{cognitive}{active computation} and \textcolor{cognitive}{fact retrieval}, and solve for the same variable $x$ (the number of 30-minute episodes), but the English trace fails to incorporate the Thursday and Friday information and thus outputs an incorrect answer.

In~\autoref{tab:qualitative_2}, we further show an example where the same frequent use of \textcolor{cognitive}{self-checking} feature to opposite $\Delta\mathrm{Acc}$ effects for English and Swahili. In Swahili trace, the model produces many contradictory self-checks that repeatedly reopen already-settled intermediate interpretations. In English traces, however, the model checks a step, arrives at a definitive answer, and moves on, rather than looping back indefinitely.

\subsection{SAE Consistency}
\label{appendix:consistency}

We additionally analyze SAE consistency by taking the top-10 concepts per language and dataset and computing the average pairwise sentence embedding similarity. Tables \ref{tab:con_model}, \ref{tab:con_lang}, and \ref{tab:con_data} show results across models, languages, and datasets, respectively.

\begin{table}
\centering
\resizebox{0.8\linewidth}{!}{%
    \begin{tabular}{lllll}
    \toprule
    \textbf{Model} & \textbf{\textsc{Distill-Qwen 1.5b}} & \textbf{\textsc{Distill-Qwen 7b}} & \textbf{\textsc{Qwen3 4b}} & \textbf{\textsc{Qwen3 8b}} \\
    \toprule
    \textbf{\textsc{Distill-Qwen 1.5b}} & 1.000 & & & \\
    \textbf{\textsc{Distill-Qwen 7b}} & 0.846 & 1.000 & & \\
    \textbf{\textsc{Qwen3 4b}} & 0.513 & 0.602 & 1.000 & \\
    \textbf{\textsc{Qwen3 8b}} & 0.547 & 0.586 & 0.819 & 1.000 \\
    \bottomrule
    \end{tabular}
}
\caption{\textbf{SAE consistency across models for MGSM-Rev2.} Models within the same family (\textsc{Distill-Qwen} and \textsc{Qwen3}) show higher similarity.}
\label{tab:con_model}
\end{table}
\begin{table}
\centering
\resizebox{\linewidth}{!}{%
    \begin{tabular}{lllllllllll}
    \toprule
    \textbf{Language} & \textbf{Bengali} & \textbf{English} & \textbf{German} & \textbf{Spanish} & \textbf{French} & \textbf{Russian} & \textbf{Swahili} & \textbf{Telugu} & \textbf{Thai} & \textbf{Chinese} \\
    \toprule
    \textbf{Bengali} & 1.000 & & & & & & & & & \\
    \textbf{English} & 0.405 & 1.000 & & & & & & & & \\
    \textbf{German} & 0.319 & 0.571 & 1.000 & & & & & & & \\
    \textbf{Spanish} & 0.373 & 0.501 & 0.447 & 1.000 & & & & & & \\
    \textbf{French} & 0.383 & 0.523 & 0.463 & 0.648 & 1.000 & & & & & \\
    \textbf{Russian} & 0.391 & 0.421 & 0.389 & 0.392 & 0.401 & 1.000 & & & & \\
    \textbf{Swahili} & 0.203 & 0.224 & 0.211 & 0.219 & 0.217 & 0.208 & 1.000 & & & \\
    \textbf{Telugu} & 0.192 & 0.198 & 0.187 & 0.194 & 0.191 & 0.185 & 0.143 & 1.000 & & \\
    \textbf{Thai} & 0.178 & 0.183 & 0.171 & 0.177 & 0.175 & 0.169 & 0.134 & 0.148 & 1.000 & \\
    \textbf{Chinese} & 0.165 & 0.176 & 0.163 & 0.168 & 0.170 & 0.162 & 0.128 & 0.141 & 0.247 & 1.000 \\
    \bottomrule
    \end{tabular}
}
\caption{\textbf{SAE consistency across languages for \textsc{Distill-Qwen} series models with MGSM-Rev2.} Typologically related languages (e.g., Spanish and French) show relatively higher similarity.}
\label{tab:con_lang}
\end{table}
\begin{table}
\centering
\resizebox{0.5\linewidth}{!}{%
    \begin{tabular}{ll}
    \toprule
    \textbf{Model} & \textbf{AIME vs. MGSM-Rev2} \\
    \toprule
    \textsc{Distill-Qwen 1.5b} & 0.626 \\
    \textsc{Distill-Qwen 7b} & 0.614 \\
    \textsc{Qwen3 4b} & 0.587 \\
    \textsc{Qwen3 8b} & 0.562 \\
    \bottomrule
    \end{tabular}
}
\caption{\textbf{SAE consistency between AIME and MGSM-Rev2 datasets.} \textsc{Distill-Qwen} series models have higher similarity.}
\label{tab:con_data}
\end{table}

\subsection{Test-time Selection Policy Ablations}
\label{appendix:n_ablation}

We set $n=32$ in our main test-time selection experiments, sampling 8 independent candidate reasoning traces at each temperature $t\in\{0.3, 0.6, 0.8, 1.0\}$, following prior work on multi-temperature test-time scaling \citep{feng2025characterizeseffectivereasoningrevisiting}.
However, in practical deployment scenarios, sampling and re-ranking 32 candidates per query can be computationally expensive. To study the trade-off between compute and performance, we additionally evaluate smaller budgets $n \in \{4,8,16,32\}$, sampling $\frac{n}{4}$ traces at each temperature $t$ and reporting compute time in~\autoref{tab:n_ablation} and pass@1 in~\autoref{tab:n_ablation_2}. As $n$ increases, we observe higher computational overhead but also larger pass@1 gains over the random selection baseline, illustrating a compute-performance trade-off.

\begin{table}
\centering
\resizebox{0.7\linewidth}{!}{%
    \begin{tabular}{llllllllll}
    \toprule
    \textbf{Dataset} & \textbf{Model} & \textbf{$n=4$} & \textbf{$n=8$} & \textbf{$n=16$} & \textbf{$n=32$} \\
    \toprule

    \textbf{MGSM-Rev2} & \textsc{Distill-Qwen 1.5b} & 03:32 & 06:15 & 08:01 & 09:27 \\
    & \textsc{Distill-Qwen 7b} & 04:38 & 05:48 & 07:39 & 09:25 \\
    & \textsc{Qwen3 4b} & 01:35 & 04:19 & 05:50 & 07:11 \\
    & \textsc{Qwen3 8b} & 04:20 & 07:57 & 11:21 & 14:46 \\
    \midrule
    
    \textbf{AIME} & \textsc{Distill-Qwen 1.5b} & 06:32 & 14:09 & 19:26 & 23:58 \\
    & \textsc{Distill-Qwen 7b} & 19:40 & 41:27 & 60:03 & 77:32 \\
    & \textsc{Qwen3 4b} & 11:46 & 24:59 & 36:16 & 47:27 \\
    & \textsc{Qwen3 8b} & 26:43 & 52:10 & 79:07 & 106:49 \\

    \bottomrule
    \end{tabular}
}
\caption{\textbf{Computation time (MM:SS) for varying $n$ in test-time selection.} We show longer compute time with increasing $n$ for all dataset and models.}
\label{tab:n_ablation}
\end{table}

\begin{table}
\centering
\resizebox{\linewidth}{!}{%
    \begin{tabular}{llllllllll}
    \toprule
    \textbf{Model} & \textbf{$n$} & \textbf{Random} & \textbf{Struc.} & \textbf{Sem.} & \textbf{Num.} & \textbf{Direct} & \textbf{Indirect} & \textbf{Result} & \textbf{Uncertainty} \\
    \toprule

    \rowcolor{gray!20}
    \multicolumn{10}{c}{\textbf{MGSM-Rev2}} \\
    \midrule

    \textbf{\textsc{Distill-Qwen 1.5b}} & 4 & 0.366 & 0.367 & 0.378 & \textbf{0.367} & 0.381 & 0.381 & 0.360 & \textbf{0.360} \\
     & 8 & 0.368 & 0.367 & 0.395 & 0.367 & 0.387 & 0.383 & 0.360 & 0.349 \\
     & 16 & 0.365 & 0.365 & 0.402 & 0.361 & 0.396 & 0.387 & 0.361 & 0.351 \\
     & 32 & 0.369 & \textbf{0.371} & \textbf{0.430} & 0.360 & \textbf{0.402} & \textbf{0.391} & \textbf{0.368} & 0.343 \\
      \midrule
      
    \textbf{\textsc{Distill-Qwen 8b}} & 4 & 0.627 & \textbf{0.623} & 0.630 & 0.622 & 0.633 & 0.604 & 0.623 & 0.633 \\
     & 8 & 0.623 & 0.622 & 0.626 & 0.629 & 0.641 & 0.612 & \textbf{0.612} & \textbf{0.636} \\
     & 16 & 0.625 & 0.615 & 0.632 & \textbf{0.632} & 0.639 & 0.600 & \textbf{0.626} & 0.607 \\
     & 32 & 0.623 & 0.589 & \textbf{0.640} & 0.617 & \textbf{0.651} & 0.609 & 0.615 & 0.596 \\
      \midrule
      
    \textbf{\textsc{Qwen3 4b}} & 4 & 0.747 & 0.755 & 0.755 & 0.734 & 0.756 & \textbf{0.759} & \textbf{0.755} & \textbf{0.737} \\
     & 8 & 0.748 & \textbf{0.756} & 0.756 & \textbf{0.735} & 0.746 & 0.746 & 0.754 & 0.732 \\
     & 16 & 0.740 & 0.755 & \textbf{0.758} & 0.732 & 0.748 & 0.747 & 0.754 & 0.730 \\
     & 32 & 0.745 & 0.753 & 0.753 & 0.723 & \textbf{0.761} & 0.743 & 0.746 & 0.727 \\
     \midrule
     
    \textbf{\textsc{Qwen3 8b}} & 4 & 0.733 & 0.739 & 0.746 & 0.735 & 0.741 & 0.736 & 0.739 & 0.721 \\
     & 8 & 0.732 & 0.743 & 0.740 & \textbf{0.738} & 0.740 & 0.738 & 0.741 & \textbf{0.728} \\
     & 16 & 0.738 & 0.741 & 0.743 & 0.733 & 0.749 & 0.739 & 0.744 & 0.721 \\
     & 32 & 0.738 & \textbf{0.756} & \textbf{0.751} & 0.726 & \textbf{0.756} & \textbf{0.741} & \textbf{0.747} & 0.717 \\

    \midrule
    \rowcolor{gray!20}
    \multicolumn{10}{c}{\textbf{AIME}} \\
    \midrule

    \textbf{\textsc{Distill-Qwen 1.5b}} & 4 & 0.181 & 0.184 & 0.192 & \textbf{0.136} & 0.210 & 0.218 & 0.202 & 0.181 \\
     & 8 & 0.184 & 0.181 & 0.196 & 0.124 & 0.221 & 0.222 & 0.208 & \textbf{0.183} \\
     & 16 & 0.181 & 0.188 & 0.199 & 0.122 & 0.223 & 0.224 & 0.211 & 0.176 \\
     & 32 & 0.183 & \textbf{0.204} & \textbf{0.200} & 0.133 & \textbf{0.246} & \textbf{0.228} & \textbf{0.221} & 0.175 \\
     \midrule
     
    \textbf{\textsc{Distill-Qwen 7b}} & 4 & 0.247 & 0.272 & 0.254 & \textbf{0.212} & 0.324 & 0.319 & 0.267 & \textbf{0.245} \\
     & 8 & 0.243 & 0.277 & 0.257 & 0.204 & 0.327 & 0.325 & 0.292 & 0.240 \\
     & 16 & 0.241 & 0.281 & 0.266 & 0.201 & 0.331 & 0.331 & 0.295 & 0.227 \\
     & 32 & 0.244 & \textbf{0.297} & \textbf{0.297} & 0.186 & \textbf{0.351} & \textbf{0.342} & \textbf{0.311} & 0.202 \\
     \midrule
     
    \textbf{\textsc{Qwen3 4b}} & 4 & 0.402 & 0.411 & 0.404 & \textbf{0.411} & 0.407 & 0.431 & 0.410 & \textbf{0.400} \\
     & 8 & 0.390 & 0.418 & 0.406 & 0.391 & 0.425 & 0.455 & 0.403 & 0.367 \\
     & 16 & 0.404 & 0.424 & 0.411 & 0.378 & 0.423 & 0.483 & 0.457 & 0.333 \\
     & 32 & 0.397 & \textbf{0.480} & \textbf{0.422} & 0.364 & \textbf{0.489} & \textbf{0.505} & \textbf{0.523} & 0.319 \\
     \midrule
     
    \textbf{\textsc{Qwen3 8b}} & 4 & 0.457 & 0.466 & \textbf{0.459} & 0.459 & 0.488 & 0.488 & 0.482 & \textbf{0.444} \\
     & 8 & 0.460 & 0.469 & 0.461 & 0.455 & 0.495 & 0.503 & 0.489 & 0.437 \\
     & 16 & 0.453 & 0.471 & 0.461 & 0.417 & 0.504 & 0.516 & 0.492 & 0.441 \\
     & 32 & 0.458 & \textbf{0.477} & \textbf{0.469} & 0.390 & \textbf{0.512} & \textbf{0.549} & \textbf{0.504} & 0.383 \\

    \bottomrule
    \end{tabular}
}
\caption{\textbf{Pass@1 per model with varying $n$ in test-time selection.} \textbf{Struc.}: Structural similarity; \textbf{Sem.}: Semantic similarity; \textbf{Num.}: Number of reasoning steps. We show smaller gains for smaller $n$ for our features over the random baseline. We \textbf{bold} the highest pass@1 for each model and feature.}
\label{tab:n_ablation_2}
\end{table}



\begin{table}
\centering
\resizebox{\linewidth}{!}{%
    \begin{tabular}{lp{0.33\linewidth}p{0.9\linewidth}}
    \toprule
    \textbf{Configuration} & \textbf{Options} & \textbf{Rationale} \\
    \toprule

    \textbf{Chunking Method} & $\{0, 250-500\}$ & Following recommendations in \citet{movva2025sparse}, we split long inputs (reasoning traces) into chunks of roughly 250-500 words for better feature interpretation. We do not use sentence-level tokenization or splitting on our reasoning trace separator (\textbackslash n\textbackslash n), as these create many short segments and substantially increase the number of embedding calls. We also avoid using a separate LLM to summarize the input texts, both to preserve as much of the original information as possible and to limit the number of additional LLM calls. In practice, a chunk size of about 400 words (allowing the threshold to fall within a word) was computationally efficient.\\
    \midrule

    \textbf{Embedding Model} & \{\textsc{RoBERTa}, \textsc{ModernBERT}, \textbf{\textsc{LaBSE}}, \textsc{OpenAI text-embedding-3}\} & We use \textsc{LaBSE} embeddings as our embedding model for three reasons: (i) it is explicitly trained for multilingual sentence representations and covers all languages in our tested datasets, (ii) it provides off-the-shelf embeddings without requiring additional API calls or model fine-tuning, and (iii) in our qualitative observations, it produced more coherent reasoning concepts than the monolingual (\textsc{RoBERTa}/\textsc{ModernBERT}) alternatives. \\
    \midrule

    \textbf{SAE Parameters} & \{Vanilla TopK, Matryoshka TopK, \textbf{Batch TopK}\} & We follow the rules of thumb from \citet{movva2025sparse}, where $M$ (the total number of concepts) and $K$ (the number of active concepts per example) control concept granularity and sparsity. Larger $M$ allows the SAE to represent more fine-grained reasoning patterns across the inputs, while larger $K$ enables each trace to be decomposed into a richer combination of concepts. For our dataset size and chunked trace length, we select $M=256$ and $K=8$ as a middle ground: this is large enough to separate broad phenomena into distinct concepts without oversplitting into nearly-identical concepts. \\
    \midrule

    \textbf{Selection Method} & \{\textbf{Correlation}, LASSO, Separation score\} & We use the correlation-based selector, which ranks neurons by the Pearson correlation between their activations and the target label. We find that this method is fast, simplest to implement, and gives relatively good balance of effect size and prevalence. It was suitable for our setting where we need to screen many neurons across 10 languages and 4 models without incurring the additional optimization overhead of LASSO. \\
    \midrule

    \textbf{Interpreter Model} & \{\textsc{LLaMA-3.3 70b}, \textbf{\textsc{GPT-4o}}\} & We test two models for generating interpretations: one open-weight model (\textsc{LLaMA-3.3 70b}) that can be run within our compute budget, and one closed-source model (\textsc{GPT-4o}). We use \textsc{GPT-4o} since sampling from \textsc{LLaMA-3.3 70b} was substantially slower and its generated concept descriptions were generally less detailed (e.g., ``engages in multi-step reasoning'').\\

    \bottomrule
    \end{tabular}
}
\caption{\textbf{Summary of the configuration space in SAE analysis.} \textbf{Options}: Available settings for each configuration dimension; we \textbf{bold} the option used in our final setup. \textbf{Rationale}: Brief justification for our choice. Note that since our final outputs are natural-language descriptions of latent reasoning concepts, these rationales are primarily guided by our own qualitative observations.}  
\label{tab:sae_config}
\end{table}

\begin{table}
\centering
\resizebox{\linewidth}{!}{%
    \begin{tabular}{p{0.27\linewidth}p{0.3\linewidth}p{0.3\linewidth}p{0.34\linewidth}}
    \toprule
    \textbf{Query (English)} & \textbf{Reasoning trace (English)} & \textbf{Reasoning trace (Swahili)} & \textbf{Reasoning trace (Swahili-MT)} \\
    \toprule

    Billy sells DVDs. He has 8 customers on Tuesday. The first 3 buy 1 DVD each. The next 2 buy 2 DVDs each. The last 3 buy nothing. How many DVDs did Billy sell?
    &
    [\ldots] Let me make sure I have the customer count right. 3 + 2 + 3 = 8. Yes, that matches the 8 customers stated. Good, I'm not misreading the groups. [\ldots] I haven't mixed those up, right? [\ldots] \textbackslash boxed\{7\}
    &
    [\ldots] Subiri. Acha niangalie tena. Tatizo linasema wateja 3 wa kwanza hununua DVD moja kila mmoja. Hiyo ni 3. [\ldots] Hmm, lakini ninahitaji kuhakikisha. [\ldots] \textbackslash boxed\{10\}
    &
    [\ldots] Wait. Let me look at it again. The problem says the first 3 customers buy one DVD each. That's 3. [\ldots] Hmm, but I need to make sure. [\ldots] \textbackslash boxed\{10\} \\

    \bottomrule
    \end{tabular}
}
\caption{\textbf{Qualitative comparison of \textcolor{cognitive}{self-checking} behavior across languages.} English trace leads to $\Delta\mathrm{Acc}>$0 while Swahili trace leads to $\Delta\mathrm{Acc}<$0.}
\label{tab:qualitative_2}
\end{table}

\begin{table*}
\centering
  \scriptsize
  \begin{tabularx}{\linewidth}{%
  l
  *{3}{Y}
  *{3}{Y}
  Y p{0.04\linewidth}   
  *{8}{Y}
  l}                                   
  \toprule
  \multirow{2}{*}{\textbf{Lang.}}
    & \multicolumn{3}{c}{\raisebox{-0.2em}{\includegraphics[height=1.1em]{figures/logo/multilingual.png}} \textcolor{multilingual}{\textbf{Alignment}}}
    & \multicolumn{5}{c}{\raisebox{-0.2em}{\includegraphics[height=1.1em]{figures/logo/intermediate.png}} \textcolor{intermediate}{\textbf{Reasoning Step}}}
    & \multicolumn{8}{c}{\raisebox{-0.2em}{\includegraphics[height=1.1em]{figures/logo/cognitive.png}} \textcolor{cognitive}{\textbf{Reasoning Flow}}}
    & \multirow{2}{*}{\textbf{Acc.}} \\
      \cmidrule(lr){2-4}
      \cmidrule(lr){5-9}
      \cmidrule(lr){10-17}
      & \textbf{C} & \textbf{St} & \textbf{Se}
      & \textbf{\#} & \textbf{V} & \textbf{D} & \textbf{I} & \textbf{$\mathcal{V}I$} 
      & \textbf{S} & \textbf{A} & \textbf{P} & \textbf{Pl} & \textbf{F} & \textbf{Fa} & \textbf{R} & \textbf{U}
  & \\
    
    \toprule
    \rowcolor{gray!20}
    \multicolumn{18}{c}{\textbf{\textsc{Distill-Qwen 1.5b}}} \\
    \midrule

    \textbf{bn} & 0.874 & 0.450 & 0.744 & 41 & 0.192 & 0.518 & 0.440 & -9.49 & 0.037 & 0.234 & 0.237 & 0.099 & 0.044 & 0.083 & 0.071 & 0.162 & 0.240 \\
    \textbf{de} & 0.847 & 0.590 & 0.801 & 4 & 0.326 & 0.779 & 0.578 & 2.54 & 0.008 & 0.344 & 0.282 & 0.127 & 0.031 & 0.089 & 0.111 & 0.007 & 0.432 \\
    \textbf{en} & --- & --- & --- & 32 & 0.199 & 0.802 & 0.670 & -6.40 & 0.068 & 0.309 & 0.127 & 0.152 & 0.041 & 0.087 & 0.127 & 0.084 & 0.876 \\
    \textbf{es} & 0.865 & 0.571 & 0.832 & 5 & 0.300 & 0.874 & 0.674 & -2.10 & 0.000 & 0.357 & 0.273 & 0.094 & 0.038 & 0.140 & 0.094 & 0.004 & 0.592 \\
    \textbf{fr} & 0.864 & 0.600 & 0.831 & 5 & 0.332 & 0.770 & 0.579 & -2.34 & 0.020 & 0.309 & 0.244 & 0.185 & 0.041 & 0.093 & 0.095 & 0.014 & 0.496 \\
    \textbf{ru} & 0.855 & 0.543 & 0.796 & 13 & 0.292 & 0.734 & 0.593 & -8.84 & 0.019 & 0.303 & 0.225 & 0.140 & 0.064 & 0.118 & 0.092 & 0.036 & 0.524 \\
    \textbf{sw} & 0.825 & 0.424 & 0.653 & 24 & 0.210 & 0.216 & 0.178 & 7.29 & 0.005 & 0.136 & 0.398 & 0.076 & 0.042 & 0.059 & 0.051 & 0.137 & 0.024 \\
    \textbf{te} & 0.850 & 0.495 & 0.678 & 41 & 0.222 & 0.264 & 0.209 & -6.44 & 0.002 & 0.201 & 0.344 & 0.092 & 0.044 & 0.109 & 0.077 & 0.056 & 0.056 \\
    \textbf{th} & 0.833 & 0.467 & 0.578 & 27 & 0.203 & 0.236 & 0.175 & -1.19 & 0.007 & 0.170 & 0.286 & 0.073 & 0.074 & 0.061 & 0.054 & 0.016 & 0.124 \\
    \textbf{zh} & 0.848 & 0.557 & 0.824 & 27 & 0.236 & 0.821 & 0.658 & -12.6 & 0.039 & 0.344 & 0.197 & 0.082 & 0.044 & 0.109 & 0.119 & 0.062 & 0.680 \\

    \midrule
    \rowcolor{gray!20}
    \multicolumn{18}{c}{\textbf{\textsc{Distill-Qwen 7b}}} \\
    \midrule

    \textbf{bn} & 0.874 & 0.604 & 0.737 & 5 & 0.388 & 0.791 & 0.591 & -9.66 & 0.003 & 0.261 & 0.371 & 0.070 & 0.052 & 0.102 & 0.103 & 0.028 & 0.540 \\
    \textbf{de} & 0.847 & 0.559 & 0.794 & 5 & 0.355 & 0.878 & 0.654 & -6.59 & 0.002 & 0.329 & 0.248 & 0.133 & 0.026 & 0.131 & 0.129 & 0.002 & 0.764 \\
    \textbf{en} & --- & --- & --- & 29 & 0.089 & 0.760 & 0.702 & -5.31 & 0.150 & 0.239 & 0.087 & 0.106 & 0.054 & 0.092 & 0.139 & 0.130 & 0.956 \\
    \textbf{es} & 0.865 & 0.597 & 0.816 & 5 & 0.289 & 0.892 & 0.679 & -3.57 & 0.001 & 0.403 & 0.268 & 0.096 & 0.039 & 0.109 & 0.082 & 0.003 & 0.824 \\
    \textbf{fr} & 0.864 & 0.564 & 0.816 & 5 & 0.322 & 0.815 & 0.638 & -5.13 & 0.001 & 0.345 & 0.191 & 0.150 & 0.037 & 0.184 & 0.091 & 0.002 & 0.804 \\
    \textbf{ru} & 0.855 & 0.575 & 0.787 & 5 & 0.325 & 0.897 & 0.673 & -5.55 & 0.003 & 0.374 & 0.244 & 0.117 & 0.033 & 0.084 & 0.138 & 0.006 & 0.816 \\
    \textbf{sw} & 0.825 & 0.526 & 0.669 & 10 & 0.300 & 0.378 & 0.296 & -11.4 & 0.009 & 0.182 & 0.399 & 0.065 & 0.024 & 0.113 & 0.086 & 0.032 & 0.108 \\
    \textbf{te} & 0.850 & 0.515 & 0.759 & 10 & 0.314 & 0.589 & 0.451 & -9.20 & 0.007 & 0.258 & 0.332 & 0.092 & 0.034 & 0.133 & 0.084 & 0.041 & 0.288 \\
    \textbf{th} & 0.833 & 0.550 & 0.734 & 8 & 0.322 & 0.624 & 0.496 & -8.59 & 0.020 & 0.244 & 0.302 & 0.136 & 0.037 & 0.157 & 0.072 & 0.026 & 0.608 \\
    \textbf{zh} & 0.848 & 0.551 & 0.802 & 9 & 0.278 & 0.920 & 0.721 & -4.85 & 0.025 & 0.377 & 0.233 & 0.068 & 0.046 & 0.106 & 0.116 & 0.028 & 0.824 \\

    \midrule
    \rowcolor{gray!20}
    \multicolumn{18}{c}{\textbf{\textsc{Qwen-3 8b}}} \\
    \midrule

    \textbf{bn} & 0.874 & 0.425 & 0.914 & 25 & 0.106 & 0.760 & 0.686 & -6.47 & 0.128 & 0.201 & 0.149 & 0.111 & 0.043 & 0.090 & 0.060 & 0.218 & 0.900 \\
    \textbf{de} & 0.847 & 0.487 & 0.940 & 13 & 0.162 & 0.823 & 0.698 & -4.25 & 0.158 & 0.215 & 0.191 & 0.144 & 0.032 & 0.098 & 0.041 & 0.121 & 0.968 \\
    \textbf{en} & --- & --- & --- & 26 & 0.090 & 0.748 & 0.687 & -4.69 & 0.145 & 0.194 & 0.090 & 0.095 & 0.062 & 0.086 & 0.089 & 0.238 & 0.988 \\
    \textbf{es} & 0.865 & 0.492 & 0.940 & 14 & 0.149 & 0.837 & 0.722 & -4.52 & 0.144 & 0.224 & 0.175 & 0.135 & 0.035 & 0.088 & 0.045 & 0.155 & 0.968 \\
    \textbf{fr} & 0.864 & 0.487 & 0.935 & 15 & 0.135 & 0.824 & 0.716 & -5.58 & 0.156 & 0.219 & 0.179 & 0.141 & 0.034 & 0.081 & 0.050 & 0.140 & 0.956 \\
    \textbf{ru} & 0.855 & 0.449 & 0.865 & 20 & 0.110 & 0.794 & 0.709 & -7.26 & 0.166 & 0.238 & 0.130 & 0.118 & 0.059 & 0.081 & 0.060 & 0.148 & 0.956 \\
    \textbf{sw} & 0.825 & 0.340 & 0.837 & 41 & 0.098 & 0.744 & 0.691 & -9.22 & 0.048 & 0.138 & 0.203 & 0.076 & 0.032 & 0.082 & 0.028 & 0.391 & 0.372 \\
    \textbf{te} & 0.850 & 0.435 & 0.907 & 20 & 0.150 & 0.767 & 0.666 & -6.90 & 0.112 & 0.197 & 0.211 & 0.106 & 0.028 & 0.092 & 0.044 & 0.209 & 0.840 \\
    \textbf{th} & 0.833 & 0.453 & 0.924 & 18 & 0.133 & 0.834 & 0.736 & -11.1 & 0.127 & 0.198 & 0.164 & 0.130 & 0.040 & 0.106 & 0.050 & 0.186 & 0.900 \\
    \textbf{zh} & 0.848 & 0.394 & 0.851 & 26 & 0.074 & 0.695 & 0.636 & -2.76 & 0.140 & 0.185 & 0.084 & 0.092 & 0.062 & 0.085 & 0.069 & 0.283 & 0.964 \\

    \midrule
    \rowcolor{gray!20}
    \multicolumn{18}{c}{\textbf{\textsc{Qwen-3 14b}}} \\
    \midrule
    
    \textbf{bn} & 0.874 & 0.440 & 0.925 & 20 & 0.119 & 0.765 & 0.677 & -5.98 & 0.136 & 0.231 & 0.164 & 0.120 & 0.044 & 0.091 & 0.058 & 0.156 & 0.940 \\
    \textbf{de} & 0.847 & 0.455 & 0.941 & 17 & 0.126 & 0.783 & 0.693 & -3.97 & 0.162 & 0.196 & 0.139 & 0.118 & 0.053 & 0.113 & 0.055 & 0.163 & 0.968 \\
    \textbf{en} & --- & --- & --- & 33 & 0.089 & 0.697 & 0.648 & -0.51 & 0.149 & 0.198 & 0.085 & 0.091 & 0.056 & 0.091 & 0.108 & 0.222 & 0.984 \\
    \textbf{es} & 0.865 & 0.487 & 0.941 & 16 & 0.139 & 0.823 & 0.716 & -5.16 & 0.157 & 0.225 & 0.158 & 0.120 & 0.042 & 0.102 & 0.059 & 0.136 & 0.984 \\
    \textbf{fr} & 0.864 & 0.484 & 0.934 & 16 & 0.136 & 0.833 & 0.724 & -5.32 & 0.157 & 0.212 & 0.174 & 0.116 & 0.038 & 0.086 & 0.057 & 0.159 & 0.968 \\
    \textbf{ru} & 0.855 & 0.447 & 0.938 & 21 & 0.109 & 0.788 & 0.708 & -5.59 & 0.159 & 0.215 & 0.120 & 0.115 & 0.059 & 0.091 & 0.070 & 0.172 & 0.984 \\
    \textbf{sw} & 0.825 & 0.381 & 0.875 & 34 & 0.111 & 0.738 & 0.668 & -8.01 & 0.084 & 0.169 & 0.166 & 0.103 & 0.037 & 0.108 & 0.047 & 0.281 & 0.644 \\
    \textbf{te} & 0.850 & 0.416 & 0.918 & 25 & 0.113 & 0.725 & 0.652 & -5.68 & 0.120 & 0.206 & 0.147 & 0.111 & 0.051 & 0.105 & 0.066 & 0.194 & 0.908 \\
    \textbf{th} & 0.833 & 0.432 & 0.930 & 23 & 0.113 & 0.773 & 0.690 & -2.48 & 0.132 & 0.196 & 0.133 & 0.125 & 0.047 & 0.119 & 0.064 & 0.184 & 0.960 \\
    \textbf{zh} & 0.848 & 0.381 & 0.852 & 38 & 0.064 & 0.586 & 0.544 & -1.35 & 0.151 & 0.195 & 0.064 & 0.081 & 0.053 & 0.082 & 0.101 & 0.271 & 0.972 \\
    
  \bottomrule
  \end{tabularx}
\caption{\textbf{Raw feature values and accuracies for each model and language in MGSM-Rev2.} Each column maps back to the features in~\autoref{tab:features}. For \textcolor{multilingual}{\textbf{Multilingual Alignment}} features, \textbf{C}: COMET-QE; \textbf{St}: Structural similarity; \textbf{Se}: Semantic similarity. For \textcolor{intermediate}{\textbf{Reasoning Step}} features, \textbf{\#}: Num. Steps; \textbf{V}: Validity; \textbf{D}: Direct utility; \textbf{I}: Indirect utility; \textbf{$\mathcal{V}I$}: $\mathcal{V}$-Information. For \textcolor{cognitive}{\textbf{Reasoning Flow}} features, \textbf{S}: Self-Checking, \textbf{A}: Active Computation, \textbf{P}: Problem Setup, \textbf{Pl}: Plan Generation; \textbf{F}: Final Answer Emission; \textbf{Fa}: Fact Retrieval; \textbf{R}: Result Consolidation; \textbf{U}: Uncertainty Management.}
\label{tab:raw_mgsm}
\end{table*}

\begin{table*}
\centering
  \scriptsize
  \begin{tabularx}{\linewidth}{%
  l
  *{3}{Y}
  *{3}{Y}
  Y p{0.04\linewidth}   
  *{8}{Y}
  l}                                  
  \toprule
  \multirow{2}{*}{\textbf{Lang.}}
    & \multicolumn{3}{c}{\raisebox{-0.2em}{\includegraphics[height=1.1em]{figures/logo/multilingual.png}} \textcolor{multilingual}{\textbf{Alignment}}}
    & \multicolumn{5}{c}{\raisebox{-0.2em}{\includegraphics[height=1.1em]{figures/logo/intermediate.png}} \textcolor{intermediate}{\textbf{Reasoning Step}}}
    & \multicolumn{8}{c}{\raisebox{-0.2em}{\includegraphics[height=1.1em]{figures/logo/cognitive.png}} \textcolor{cognitive}{\textbf{Reasoning Flow}}}
    & \multirow{2}{*}{\textbf{Acc.}} \\
      \cmidrule(lr){2-4}
      \cmidrule(lr){5-9}
      \cmidrule(lr){10-17}
      & \textbf{C} & \textbf{St} & \textbf{Se}
      & \textbf{\#} & \textbf{V} & \textbf{D} & \textbf{I} & \textbf{$\mathcal{V}I$} 
      & \textbf{S} & \textbf{A} & \textbf{P} & \textbf{Pl} & \textbf{F} & \textbf{Fa} & \textbf{R} & \textbf{U}
  & \\
    
    \toprule
    \rowcolor{gray!20}
    \multicolumn{18}{c}{\textbf{\textsc{Distill-Qwen 1.5b}}} \\
    \midrule

    \textbf{bn} & 0.814 & 0.282 & 0.865 & 354 & 0.018 & 0.674 & 0.668 & -1.86 & 0.020 & 0.264 & 0.030 & 0.100 & 0.009 & 0.133 & 0.083 & 0.204 & 0.167 \\
    \textbf{de} & 0.815 & 0.491 & 0.825 & 181 & 0.088 & 0.624 & 0.575 & 20.9 & 0.023 & 0.317 & 0.117 & 0.171 & 0.024 & 0.152 & 0.073 & 0.064 & 0.117 \\
    \textbf{en} & --- & --- & --- & 424 & 0.008 & 0.565 & 0.562 & 1.81 & 0.049 & 0.227 & 0.019 & 0.096 & 0.006 & 0.172 & 0.080 & 0.086 & 0.233 \\
    \textbf{es} & 0.820 & 0.424 & 0.865 & 303 & 0.056 & 0.542 & 0.511 & 10.6 & 0.009 & 0.232 & 0.095 & 0.118 & 0.019 & 0.123 & 0.083 & 0.091 & 0.117 \\
    \textbf{fr} & 0.834 & 0.407 & 0.864 & 208 & 0.101 & 0.625 & 0.598 & 14.5 & 0.053 & 0.269 & 0.102 & 0.171 & 0.026 & 0.127 & 0.096 & 0.098 & 0.167 \\
    \textbf{ru} & 0.822 & 0.339 & 0.827 & 275 & 0.008 & 0.370 & 0.367 & 0.54 & 0.037 & 0.228 & 0.015 & 0.095 & 0.009 & 0.145 & 0.064 & 0.110 & 0.100 \\
    \textbf{sw} & 0.790 & 0.340 & 0.853 & 256 & 0.079 & 0.485 & 0.468 & 10.3 & 0.024 & 0.220 & 0.089 & 0.105 & 0.008 & 0.128 & 0.083 & 0.197 & 0.067 \\
    \textbf{te} & 0.788 & 0.310 & 0.851 & 262 & 0.043 & 0.473 & 0.466 & -4.51 & 0.016 & 0.237 & 0.085 & 0.089 & 0.011 & 0.091 & 0.074 & 0.235 & 0.050 \\
    \textbf{th} & 0.795 & 0.350 & 0.761 & 147 & 0.085 & 0.340 & 0.316 & 14.4 & 0.019 & 0.245 & 0.161 & 0.109 & 0.061 & 0.095 & 0.043 & 0.127 & 0.067 \\
    \textbf{zh} & 0.809 & 0.284 & 0.854 & 203 & 0.019 & 0.525 & 0.519 & 3.72 & 0.067 & 0.286 & 0.024 & 0.141 & 0.016 & 0.147 & 0.097 & 0.159 & 0.200 \\

    \midrule
    \rowcolor{gray!20}
    \multicolumn{18}{c}{\textbf{\textsc{Distill-Qwen 7b}}} \\
    \midrule

    \textbf{bn} & 0.814 & 0.303 & 0.903 & 258 & 0.018 & 0.683 & 0.672 & -10.1 & 0.043 & 0.307 & 0.041 & 0.130 & 0.014 & 0.130 & 0.086 & 0.138 & 0.300 \\
    \textbf{de} & 0.815 & 0.390 & 0.866 & 263 & 0.039 & 0.610 & 0.594 & -7.92 & 0.033 & 0.333 & 0.038 & 0.110 & 0.016 & 0.156 & 0.097 & 0.066 & 0.333 \\
    \textbf{en} & --- & --- & --- & 395 & 0.012 & 0.526 & 0.522 & -7.30 & 0.045 & 0.271 & 0.013 & 0.101 & 0.007 & 0.131 & 0.104 & 0.064 & 0.433 \\
    \textbf{es} & 0.820 & 0.334 & 0.903 & 350 & 0.015 & 0.657 & 0.650 & -7.98 & 0.032 & 0.273 & 0.039 & 0.123 & 0.010 & 0.155 & 0.095 & 0.073 & 0.367 \\
    \textbf{fr} & 0.834 & 0.362 & 0.895 & 331 & 0.030 & 0.588 & 0.581 & -10.2 & 0.037 & 0.300 & 0.039 & 0.105 & 0.009 & 0.110 & 0.093 & 0.091 & 0.433 \\
    \textbf{ru} & 0.822 & 0.377 & 0.837 & 276 & 0.017 & 0.472 & 0.464 & -6.18 & 0.026 & 0.322 & 0.041 & 0.118 & 0.011 & 0.152 & 0.072 & 0.084 & 0.283 \\
    \textbf{sw} & 0.790 & 0.309 & 0.846 & 276 & 0.054 & 0.653 & 0.639 & -12.5 & 0.026 & 0.252 & 0.064 & 0.095 & 0.011 & 0.154 & 0.066 & 0.149 & 0.117 \\
    \textbf{te} & 0.788 & 0.342 & 0.876 & 300 & 0.029 & 0.611 & 0.604 & -12.3 & 0.025 & 0.333 & 0.044 & 0.114 & 0.010 & 0.139 & 0.091 & 0.136 & 0.250 \\
    \textbf{th} & 0.795 & 0.375 & 0.856 & 171 & 0.031 & 0.534 & 0.520 & -11.5 & 0.061 & 0.319 & 0.037 & 0.163 & 0.017 & 0.138 & 0.083 & 0.139 & 0.300 \\
    \textbf{zh} & 0.809 & 0.337 & 0.859 & 202 & 0.012 & 0.580 & 0.574 & -9.80 & 0.057 & 0.328 & 0.023 & 0.145 & 0.012 & 0.152 & 0.102 & 0.121 & 0.333 \\

    \midrule
    \rowcolor{gray!20}
    \multicolumn{18}{c}{\textbf{\textsc{Qwen-3 8b}}} \\
    \midrule

    \textbf{bn} & 0.814 & 0.310 & 0.901 & 258 & 0.025 & 0.663 & 0.657 & -11.0 & 0.057 & 0.290 & 0.027 & 0.147 & 0.011 & 0.156 & 0.071 & 0.152 & 0.500 \\
    \textbf{de} & 0.815 & 0.320 & 0.931 & 284 & 0.021 & 0.721 & 0.716 & -8.85 & 0.038 & 0.309 & 0.016 & 0.123 & 0.009 & 0.232 & 0.073 & 0.147 & 0.617 \\
    \textbf{en} & --- & --- & --- & 257 & 0.023 & 0.712 & 0.705 & -8.40 & 0.073 & 0.312 & 0.019 & 0.164 & 0.012 & 0.164 & 0.080 & 0.105 & 0.700 \\
    \textbf{es} & 0.820 & 0.349 & 0.941 & 269 & 0.013 & 0.724 & 0.718 & -8.92 & 0.057 & 0.331 & 0.016 & 0.164 & 0.010 & 0.157 & 0.074 & 0.085 & 0.683 \\
    \textbf{fr} & 0.834 & 0.318 & 0.931 & 266 & 0.017 & 0.807 & 0.801 & -8.43 & 0.071 & 0.334 & 0.032 & 0.156 & 0.010 & 0.174 & 0.071 & 0.099 & 0.667 \\
    \textbf{ru} & 0.822 & 0.313 & 0.860 & 237 & 0.020 & 0.716 & 0.711 & -6.25 & 0.108 & 0.319 & 0.018 & 0.159 & 0.009 & 0.157 & 0.088 & 0.123 & 0.583 \\
    \textbf{sw} & 0.790 & 0.292 & 0.893 & 278 & 0.016 & 0.688 & 0.683 & -14.0 & 0.041 & 0.285 & 0.036 & 0.117 & 0.007 & 0.134 & 0.061 & 0.212 & 0.317 \\
    \textbf{te} & 0.788 & 0.308 & 0.903 & 255 & 0.041 & 0.665 & 0.657 & -11.3 & 0.056 & 0.294 & 0.035 & 0.113 & 0.012 & 0.152 & 0.066 & 0.186 & 0.383 \\
    \textbf{th} & 0.795 & 0.308 & 0.917 & 314 & 0.022 & 0.655 & 0.650 & -14.2 & 0.048 & 0.253 & 0.036 & 0.111 & 0.008 & 0.193 & 0.079 & 0.152 & 0.567 \\
    \textbf{zh} & 0.809 & 0.248 & 0.859 & 288 & 0.009 & 0.645 & 0.642 & -16.3 & 0.053 & 0.260 & 0.020 & 0.159 & 0.012 & 0.188 & 0.076 & 0.214 & 0.533 \\

    \midrule
    \rowcolor{gray!20}
    \multicolumn{18}{c}{\textbf{\textsc{Qwen-3 14b}}} \\
    \midrule
    
    \textbf{bn} & 0.814 & 0.329 & 0.904 & 298 & 0.019 & 0.744 & 0.738 & -5.17 & 0.047 & 0.360 & 0.024 & 0.131 & 0.010 & 0.162 & 0.089 & 0.092 & 0.567 \\
    \textbf{de} & 0.815 & 0.340 & 0.927 & 285 & 0.016 & 0.708 & 0.703 & -6.76 & 0.050 & 0.339 & 0.033 & 0.128 & 0.009 & 0.162 & 0.078 & 0.095 & 0.650 \\
    \textbf{en} & --- & --- & --- & 262 & 0.015 & 0.692 & 0.688 & -2.49 & 0.106 & 0.320 & 0.014 & 0.133 & 0.008 & 0.146 & 0.078 & 0.081 & 0.667 \\
    \textbf{es} & 0.820 & 0.357 & 0.937 & 277 & 0.015 & 0.819 & 0.814 & -5.16 & 0.062 & 0.357 & 0.036 & 0.117 & 0.009 & 0.179 & 0.065 & 0.107 & 0.717 \\
    \textbf{fr} & 0.834 & 0.358 & 0.937 & 285 & 0.024 & 0.750 & 0.744 & -5.38 & 0.063 & 0.339 & 0.024 & 0.159 & 0.008 & 0.137 & 0.094 & 0.106 & 0.733 \\
    \textbf{ru} & 0.822 & 0.334 & 0.900 & 247 & 0.024 & 0.713 & 0.707 & -7.13 & 0.060 & 0.400 & 0.025 & 0.168 & 0.010 & 0.119 & 0.080 & 0.102 & 0.617 \\
    \textbf{sw} & 0.790 & 0.284 & 0.897 & 302 & 0.032 & 0.710 & 0.705 & -8.36 & 0.041 & 0.319 & 0.023 & 0.131 & 0.008 & 0.172 & 0.066 & 0.152 & 0.417 \\
    \textbf{te} & 0.788 & 0.292 & 0.898 & 286 & 0.023 & 0.646 & 0.641 & -8.38 & 0.067 & 0.283 & 0.025 & 0.153 & 0.008 & 0.144 & 0.074 & 0.175 & 0.483 \\
    \textbf{th} & 0.795 & 0.310 & 0.914 & 321 & 0.015 & 0.716 & 0.711 & -1.92 & 0.049 & 0.298 & 0.016 & 0.146 & 0.008 & 0.178 & 0.076 & 0.125 & 0.583 \\
    \textbf{zh} & 0.809 & 0.287 & 0.858 & 339 & 0.008 & 0.649 & 0.646 & -8.57 & 0.041 & 0.255 & 0.022 & 0.152 & 0.011 & 0.157 & 0.096 & 0.167 & 0.533 \\

  \bottomrule
  \end{tabularx}
\caption{\textbf{Raw feature values and accuracies for each model and language in AIME.} Each column maps back to the features in~\autoref{tab:features}. For \textcolor{multilingual}{\textbf{Multilingual Alignment}} features, \textbf{C}: COMET-QE; \textbf{St}: Structural similarity; \textbf{Se}: Semantic similarity. For \textcolor{intermediate}{\textbf{Reasoning Step}} features, \textbf{\#}: Num. Steps; \textbf{V}: Validity; \textbf{D}: Direct utility; \textbf{I}: Indirect utility; \textbf{$\mathcal{V}I$}: $\mathcal{V}$-Information. For \textcolor{cognitive}{\textbf{Reasoning Flow}} features, \textbf{S}: Self-Checking, \textbf{A}: Active Computation, \textbf{P}: Problem Setup, \textbf{Pl}: Plan Generation; \textbf{F}: Final Answer Emission; \textbf{Fa}: Fact Retrieval; \textbf{R}: Result Consolidation; \textbf{U}: Uncertainty Management.}
\label{tab:raw_aime}
\end{table*}

\begin{table}
\centering
\resizebox{\linewidth}{!}{%
    \begin{tabular}{p{0.24\linewidth} p{0.33\linewidth} p{0.5\linewidth} p{0.5\linewidth}}
    \toprule
    \textbf{Model / Language} & \textbf{Query} & \textbf{Reasoning trace (English)} & \textbf{Reasoning trace (Non-English)} \\
    \toprule

    \rowcolor{gray!20}
    \multicolumn{4}{c}{\textbf{Correct answers, Different reasoning}} \\
    \midrule

    \textbf{\textsc{Distill-Qwen 7b} (Bengali)} & Lloyd has an egg farm. His chickens produce 252 eggs per day and he sells them for $\$$2 per dozen. How much does Lloyd make on eggs per week? &   
    First, determine how many eggs Lloyd sells each week. He has 252 eggs per day, and there are 7 days in a week, so multiplying 252 by 7 gives the total number of eggs per week. Next, since eggs are sold by the dozen, divide the total number of eggs by 12 to find out how many dozens of eggs are sold each week. Finally, multiply the number of dozens by the selling price per dozen, which is $\$$2, to calculate the total earnings per week. \cmark
    & 
    The number of eggs they produce per day is 252. Eggs are sold at 2 dollars per dozen. One dozen contains 12 eggs. Therefore, the selling price per egg is 2 dollars $\div$ 12 = 1/6 dollar. So, if 252 eggs are sold per day, the income is 252 $\times$ (1/6) = 42 dollars. There are 7 days in a week, so the total weekly income is 42 $\times$ 7 = 294 dollars. \cmark
    \\
    \midrule

    \textbf{\textsc{Qwen3 4b} (French)} & Janet buys a brooch for her daughter. She pays \$500 for the material to make it and then another \$800 for the jeweler to construct it. After that, she pays 10\% of that to get it insured. How much did she pay? &     Okay, let me try to figure out how much Janet paid in total for the brooch. So, the problem says she buys a brooch for her daughter. She pays \$500 for the material, then another \$800 for the jeweler to construct it. Then she pays 10\% of that to get it insured. The question is asking how much she paid altogether. First, I need to make sure I understand all the parts of the problem. Let me break it down step by step. First expense is the material, which is \$500. Then the construction cost is \$800. So, those are two separate costs. Then, after that, she pays 10\% of `that` to get it insured. Wait, `that` – does that refer to the total of the material and construction costs, or just one of them? The wording says `10\% of that to get it insured.` The `that` is probably referring to the total amount she paid for the material and the jeweler. Let me check the original sentence again: `She pays \$500 for the material to make it and then another \$800 for the jeweler to construct it. After that, she pays 10\% of that to get it insured.`So `that` is referring to the previous total. So the `that` is the total of the material and the construction. So, \$500 + \$800 = \$1300. Then 10\% of that is 10\% of \$1300. So, the insurance cost is 10\% of \$1300. Then, the total amount she paid would be the sum of the material, construction, and insurance.Alternatively, maybe `that` refers to the construction cost? But the wording is `after that,` which would refer to the previous two payments. So, after paying for the material and the jeweler, she pays 10\% of that total. So, that makes sense. So, the insurance is 10\% of the total of the material and construction. So, let me compute that. First, the material is \$500, construction is \$800. So total of those two is 500 + 800 = 1300. Then, 10\% of 1300 is 0.10 * 1300 = 130. So, the insurance is \$130. Therefore, total amount paid is 500 + 800 + 130 = 1430. Wait, but let me check again. [...] \textcolor{red}{Alternatively, maybe the problem is trying to trick me?} [...] Therefore, the answer is \textbackslash\textbackslash boxed\{1430\}. \cmark & 
    Okay, let me try to figure out how much Janet spent in total. So, the problem says she buys a brooch for her daughter. First, she spends \$500 on materials for the crafting. Then, she spends an additional \$800 for the jewelry maker to do the work. After that, she spends 10\% of that total amount on insurance. The question is asking for the total amount she spent. Hmm, let me break this down step by step. First, the initial costs: \$500 for materials and \$800 for the jewelry maker. So, adding those together, that would be 500 + 800. Let me calculate that: 500 + 800 is 1300. So, the first two parts total \$1300. Then, she spends 10\% of this sum on insurance. Wait, the problem says ``10\% de cette somme'', which translates to 10\% of this amount. The ``cette somme'' here refers to the total of the first two expenses, right? So, 10\% of \$1300. Let me compute that. 10\% of 1300 is 0.10 * 1300, which equals 130. So, she spends an additional \$130 on insurance. Now, to find the total amount spent, I need to add the initial \$1300 to the insurance cost of \$130. That would be 1300 + 130. Let me add those numbers: 1300 + 130 is 1430. So, the total amount Janet spent is \$1430. Wait, let me double-check to make sure I didn't make a mistake. The first two expenses are definitely 500 + 800 = 1300. Then 10\% of 1300 is 130. Adding them together gives 1300 + 130 = 1430. Yeah, that seems right. The insurance is 10\% of the total of the materials and the jewelry work, not 10\% of each separately. So, the answer should be \$1430. \cmark \\

    \midrule
    \rowcolor{gray!20}
    \multicolumn{4}{c}{\textbf{Different answers, Similar reasoning}} \\

    \midrule 
    \textbf{\textsc{Qwen3 8b} (Chinese)} & 
    Frankie watches TV after he finishes his homework every night. On Monday and Tuesday, he watched a 1-hour episode of his favorite show each night. On Wednesday, he watched a few episodes of a 30-minute show. On Thursday, he finished homework early and watched a 1-hour episode and a 30-minute show. On Friday, he got to stay up late for the weekend, so he watched two 1-hour episodes. If he watched 7 hours of TV in all, how many 30-minute episodes did he watch on Wednesday?
    & 
    First, I'll calculate the total TV time Frankie spent on each day. On Monday and Tuesday, he watched a 1-hour episode each night, which totals 2 hours. On Wednesday, he watched either a few 30-minute episodes or a combination of 1-hour and 30-minute shows. Since the total TV time is 7 hours, I'll subtract the known time from Wednesday to find out how much time was spent on Wednesday. Subtracting the known time (1 hour for the 1-hour episode) from the total, we have 6 hours remaining for Wednesday. Let ($x$) represent the number of 30-minute episodes Frankie watched on Wednesday. Each 30-minute episode is 0.5 hours, so the time spent on 30-minute episodes is $0.5x$ hours. The equation becomes: $0.5x + 1 = 6$. Solving for $x$: $0.5x = 6 - 1$, $0.5x = 5$, $x = 10$. Therefore, Frankie watched 10 episodes of the 30-minute show on Wednesday. \xmark
    &
    First, let's assume he watched x episodes of 30-minute programs on Wednesday. On Monday and Tuesday, he watched 1 episode each day, for a total of 1 hour each day, or 2 hours in total. On Wednesday, he watched x episodes, each 30 minutes long, for a total of $0.5x$ hours. On Thursday, he watched one 1-hour episode and one 30-minute episode, for a total of 1.5 hours. On Friday, he watched 2 episodes, each 1 hour long, for a total of 2 hours.  Adding all the times together gives a total of 7 hours: $2 + 0.5x + 1.5 + 2 = 7$. Solving the equation: $0.5x = 7 - 2 - 1.5 - 2$, $0.5x = 1.5$, $x = 3$. Therefore, on Wednesday he watched 3 episodes, each 30 minutes long. \cmark
    \\

    \bottomrule
    \end{tabular}
}
\caption{\textbf{Qualitative examples of reasoning traces from English versus non-English queries.} For ease of interpretation, we translate reasoning traces from non-English queries into English using the Google Translate API.}  
\label{tab:qualitative}
\end{table}

\begin{figure}[t]
    \centering
      \begin{mdframed}[
        linecolor=lightgrayframe,
        linewidth=0.8pt,
        roundcorner=2pt,
        backgroundcolor=white
      ]
        \includegraphics[width=0.8\textwidth]{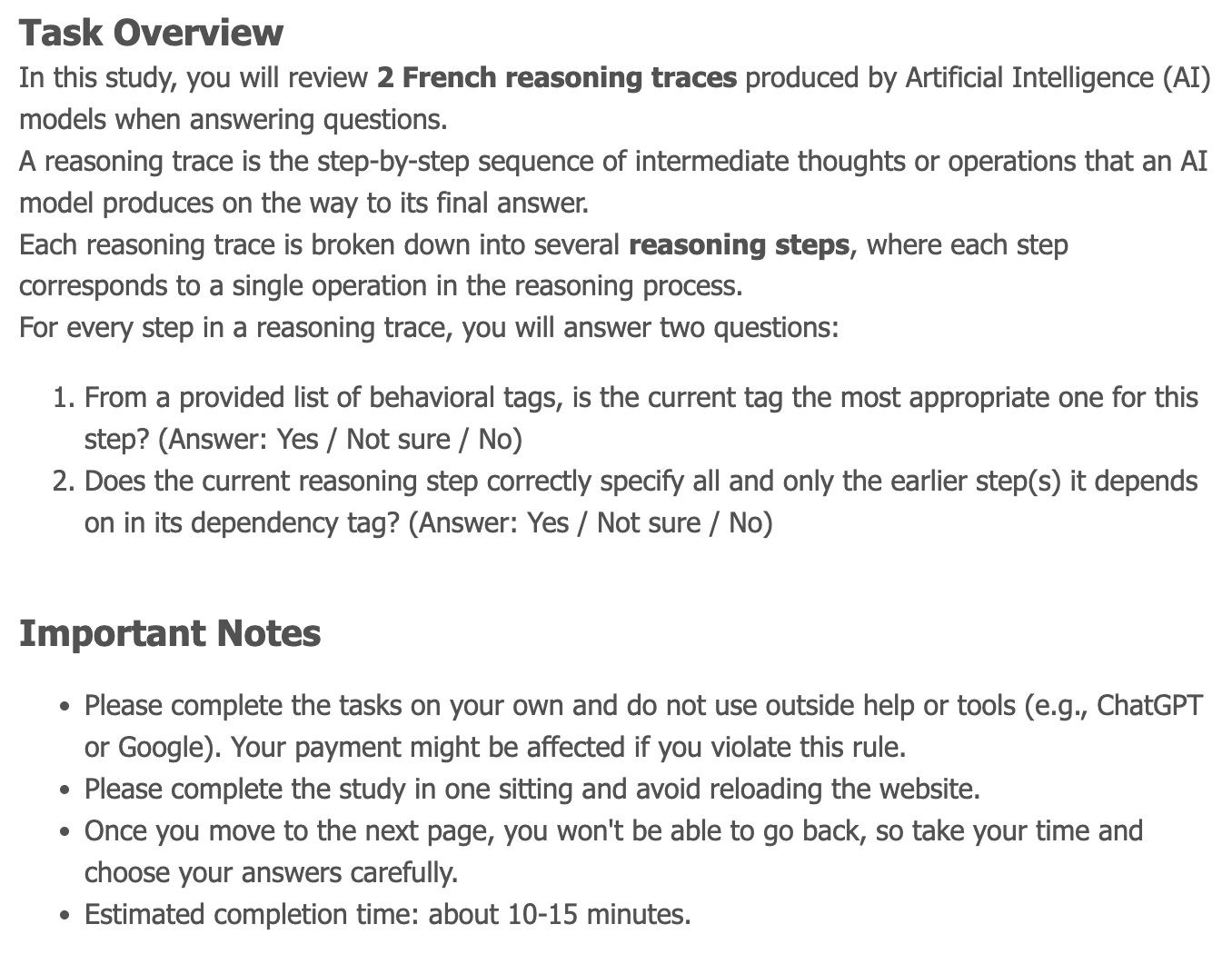}
      \end{mdframed}

      \begin{mdframed}[
        linecolor=lightgrayframe,
        linewidth=0.8pt,
        roundcorner=2pt,
        backgroundcolor=white
      ]
        \includegraphics[width=0.8\textwidth]{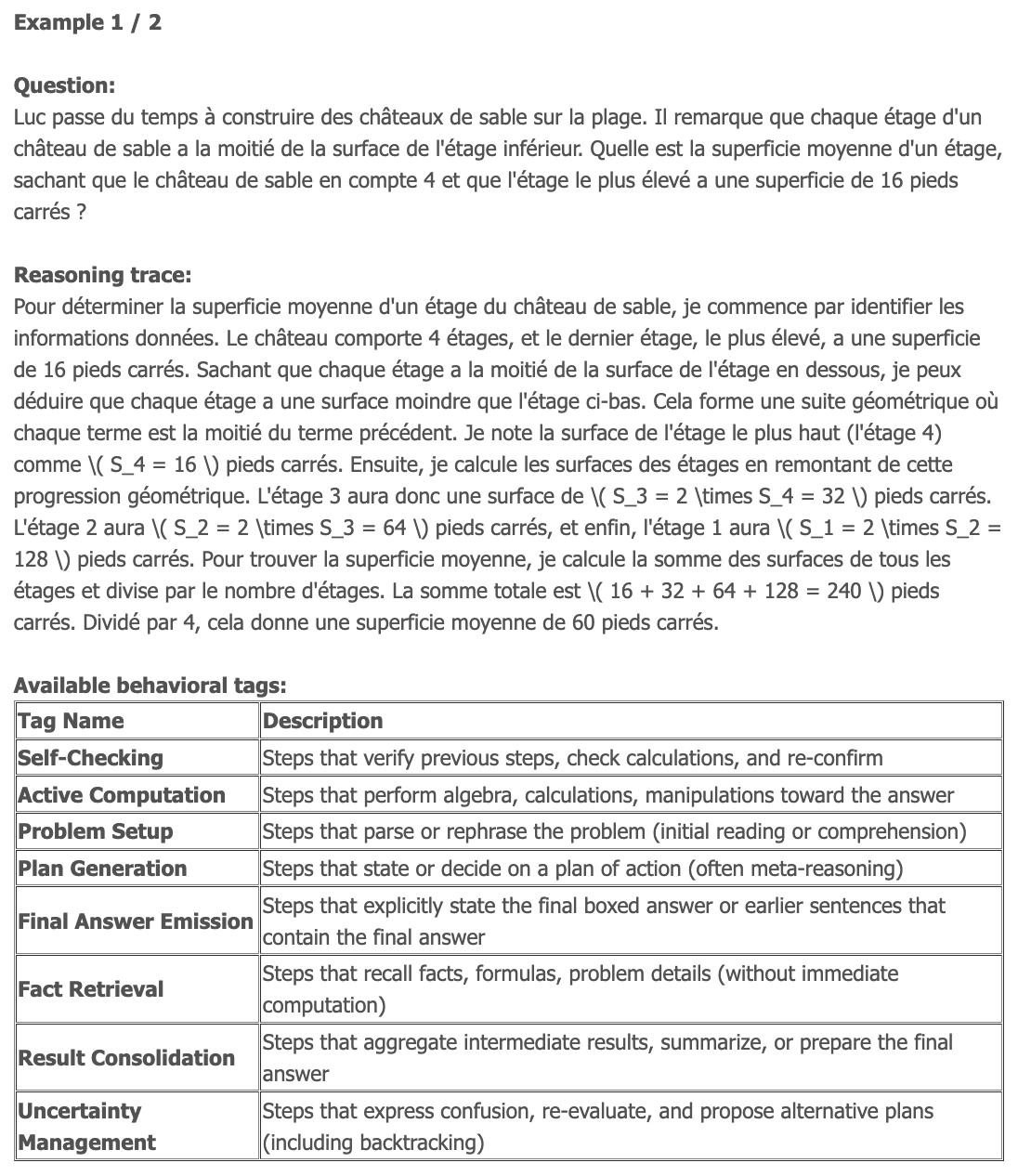}
      \end{mdframed}
\end{figure}

\begin{figure*}[!htbp]
\begin{prompt}[title={Prompt C.1. Reasoning Flow Annotation Prompt}]
\textbf{Instruction:} You are an expert in interpreting how Large Language Models solve \texttt{\{language\}} math problems using multi-step reasoning. Your task is to analyze a chain-of-thought reasoning trace, broken into discrete text sentences, and label each sentence with: \\
1. \textbf{function tags}: One or more labels that describe what this sentence is \textit{doing} functionally in the reasoning process. \\
2. \textbf{depends on}: A list of earlier sentence indices that this sentence directly depends on, e.g., uses information, results, or logic introduced in earlier sentences. \\

This annotation will be used to build a dependency graph and perform causal analysis, so please be precise and conservative: only mark a sentence as dependent on another if its reasoning clearly uses a previous sentence’s result or idea. \\

\textbf{Function Tags:} \\
1. problem setup: Parsing or rephrasing the problem (initial reading or comprehension). \\
2. plan generation: Stating or deciding on a plan of action (often meta-reasoning). \\
3. fact retrieval: Recalling facts, formulas, problem details (without immediate computation). \\
4. active computation: Performing algebra, calculations, manipulations toward the answer. \\
5. result consolidation: Aggregating intermediate results, summarizing, or preparing final answer. \\
6. uncertainty management: Expressing confusion, re-evaluating, proposing alternative plans (includes backtracking). \\
7. final answer emission: Explicit statement of the final boxed answer or earlier sentences that contain the final answer. \\
8. self checking: Verifying previous steps, checking calculations, and re-confirmations. \\
9. unknown: Use only if the sentence does not fit any of the above tags or is purely stylistic or semantic. \\

\textbf{Dependencies:} \\
For each sentence, include a list of earlier sentence indices that the reasoning in this sentence \textit{uses}. For example: \\
- If sentence 9 performs a computation based on a plan in sentence 4 and a recalled rule in sentence 5, then depends on: [4, 5] \\
- If sentence 24 uses final answer to verify correctness from sentence 23, then depends on: [23] \\
- If there's no clear dependency use an empty list: [] \\
- If sentence 13 performs a computation based on information in sentence 11, which in turn uses information from sentence 7, then depends on: [11, 7] \\

\textbf{Important Notes:} \\
- Make sure to include all dependencies for each sentence. \\
- Include both long-range and short-range dependencies. \\
- Do NOT forget about long-range dependencies. \\
- Try to be as comprehensive as possible. \\
- Make sure there is a path from earlier sentences to the final answer. \\
- ONLY label for the chain-of-thought sentence indices provided in brackets (e.g., [2]). \\

\textbf{Output Format:} \\
Return a dictionary with one entry per sentence, where each entry has: \\
- the sentence index (as the key, converted to a string), \\
- a dictionary with: \\
    - ``function tags'': list of tag strings \\
    - ``depends on'': list of sentence indices, converted to strings \\

Here is the expected format: \{
    ``1'': \{
        ``function tags'': [``problem setup''],
        ``depends on'': []
    \},
    ``4'': \{
        ``function tags'': [``plan generation''],
        ``depends on'': [``3'']
    \},
    ``5'': \{
        ``function tags'': [``fact retrieval''],
        ``depends on'': []
    \},
    ``9'': \{
        ``function tags'': [``active computation''],
        ``depends on'': [``4'', ``5'']
    \},
    ``24'': \{
        ``function tags'': [``uncertainty management''],
        ``depends on'': [``23'']
    \},
    ``32'': \{
        ``function tags'': [``final answer emission''],
        ``depends on'': [``9'', ``30'', ``32'']
    \}, ...
\}
\\ \\
Here is the math problem in English: \texttt{\{English query\}} \\
Here is the math problem in \texttt{\{language\}}: \texttt{\{Non-English query\}} \\
Here is the full chain-of-thought, broken into sentences: \texttt{\{reasoning steps\}} \\

Now label each sentence with function tags and dependencies.
\end{prompt}

\end{figure*}

\end{document}